\pdfoutput=1

\documentclass[11pt]{article}

\usepackage[final]{acl}

\usepackage{times}
\usepackage{latexsym}

\usepackage[T1]{fontenc}

\usepackage[utf8]{inputenc}
\usepackage{textgreek}

\usepackage{microtype}

\usepackage{inconsolata}

\usepackage{tcolorbox}
\usepackage{graphicx}
\usepackage{float}
\usepackage{booktabs}
\usepackage{makecell}
\usepackage{multirow}
\usepackage{enumitem}
\usepackage{calc}
\usepackage{multicol}
\usepackage{colortbl}
\usepackage{subcaption}
\usepackage{fancyvrb}
\usepackage{url}
\usepackage{pifont}
\usepackage{soul}
\usepackage{adjustbox}

\usepackage{listings}
\usepackage{listingsutf8}
\title{FinCoT: Grounding Chain-of-Thought in Expert Financial Reasoning}

\author{
 \textbf{Natapong Nitarach}, 
 \textbf{Warit Sirichotedumrong},
 \textbf{Panop Pitchayarthorn},
\\
\textbf{Pittawat Taveekitworachai},
 \textbf{Potsawee Manakul},
 \textbf{Kunat Pipatanakul}
\\
\\
 SCB 10X, SCBX Group
\\
}
\usepackage{amsfonts}  
\usepackage{amsmath}
\usepackage{cleveref}
\crefname{figure}{Fig.}{Figs.} 
\Crefname{figure}{Figure}{Figures} 
\usepackage{caption}

\usepackage{color}
\usepackage{listings}
\definecolor{deepblue}{rgb}{0,0,0.5}
\definecolor{deepred}{rgb}{0.6,0,0}
\definecolor{deepgreen}{rgb}{0,0.5,0}
\definecolor{deeppink}{rgb}{1.0,0.08,0.58}
\definecolor{judgegray}{gray}{0.95}
\definecolor{lightyellow}{RGB}{255, 255, 224}
\usepackage{needspace}

\usepackage{xcolor}
\newcommand{\goodtxt}[1]{\textcolor{green!60!black}{#1}}
\newcommand{\badtxt}[1]{\textcolor{red}{#1}}
\newcommand{\up}{\(\uparrow\)}
\newcommand{\down}{\(\downarrow\)}
\begin{document}
\maketitle
\begin{abstract}
This paper presents FinCoT, a structured chain-of-thought (CoT) prompting framework that embeds domain-specific expert financial reasoning blueprints to guide large language models' behaviors. We identify three main prompting styles in financial NLP (FinNLP): (1) standard prompting (zero-shot), (2) unstructured CoT (free-form reasoning), and (3) structured CoT (with explicitly structured reasoning steps). Prior work has mainly focused on the first two, while structured CoT remains underexplored and lacks domain expertise incorporation. Therefore, we evaluate all three prompting approaches across ten CFA-style financial domains and introduce FinCoT as the first structured finance-specific prompting approach incorporating blueprints from domain experts. FinCoT improves the accuracy of a general-purpose model, Qwen3-8B-Base, from 63.2\% to 80.5\%, and boosts Fin-R1 (7B), a finance-specific model, from 65.7\% to 75.7\%, while reducing output length by up to 8.9$\times$ and 1.16$\times$ compared to structured CoT methods, respectively. We find that FinCoT proves most effective for models lacking financial post-training. Our findings show that FinCoT does not only improve performance and reduce inference costs but also yields more interpretable and expert-aligned reasoning traces.
\end{abstract}

\section{Introduction}
Financial decision–making, such as stochastic modeling, risk assessment, portfolio optimization, and algorithmic trading~\citep{markowitz1952portfolio, black1973pricing, rockafellar2000cvar, avellaneda2008hft}, demands precise mathematics and domain-specific reasoning~\cite{lewkowycz2022solvingquantitativereasoningproblems,wen2025enhancingreasoningadaptlarge}. Recent advances in large foundation models for finance, such as the multimodal \textsc{FinTral}~\citep{bhatia2024fintralfamilygpt4level} and language-centric \textsc{Fin-R1}~\citep{liu2025finr1largelanguagemodel}, demonstrate progress but still face challenges in interpretability and domain alignment~\citep{nie2024survey,arya2024llms,Lee_2025}. Accordingly, these shortcomings motivate stricter control over a model's intermediate reasoning path, which we explore via prompt design.

Prompting guides LLM reasoning without extra training. Methods such as Chain-of-Thought~\citep{wei2023chainofthoughtpromptingelicitsreasoning}, Code Prompting~\citep{hu2023codepromptingneuralsymbolic}, Plan-and-Solve~\citep{wang-etal-2023-plan}, and Self-Reflection~\citep{Renze_2024} encourage stepwise thinking but remain domain-agnostic. In finance, this can lead to omissions in valuation checks or confusion between basis points and percentages. Yet real-world analysis follows well-defined workflows—valuation, discounting, portfolio attribution—that depend on explicit intermediate structure. Embedding such structure in the prompt helps the model verify units, formulas, and boundary conditions, improving interpretability and alignment with expert practice.

We design \textbf{FinCoT}, a zero-shot prompt that injects expert financial workflows-encoded as Mermaid blueprints-into a structured CoT template, yielding auditable reasoning without fine-tuning.  
Across ten CFA domains, FinCoT significantly boosts accuracy (most in quantitative areas) and produces shorter, clearer outputs than standard or structured CoT prompts.
This paper offers three main contributions:
\begin{itemize}[leftmargin=*]
\setlength\itemsep{-0.1em}
    \item We provide a comprehensive investigation and the first taxonomy of financial prompting--covering standard prompting, unstructured CoT, and structured CoT/FinCoT--clarifying how each paradigm addresses domain-specific reasoning requirements.
    \item We release nine blueprint templates—conceptual diagrams modeled after the Unified Modeling Language (UML)~\citep{inproceedings} and rendered in Mermaid syntax--that LLMs can parse as plain--text hints to drive structured reasoning both within FinCoT and in broader domain-specific prompting scenarios.
    \item On 1.032k CFA--style questions across ten financial domains and four open-source LLMs, FinCoT shows notable gains--up to +17.3\,pp in accuracy ($p<0.001$)--particularly on pretrained models and quantitatively structured tasks. While less effective on instruction-tuned or niche domains, FinCoT consistently reduces verbosity ($\sim$8$\times{}$ fewer tokens) and improves reasoning trace clarity under a three--point interpretability rubric.
\end{itemize}

\begin{figure*}[htbp]
    \centering
    \includegraphics[width=\textwidth]{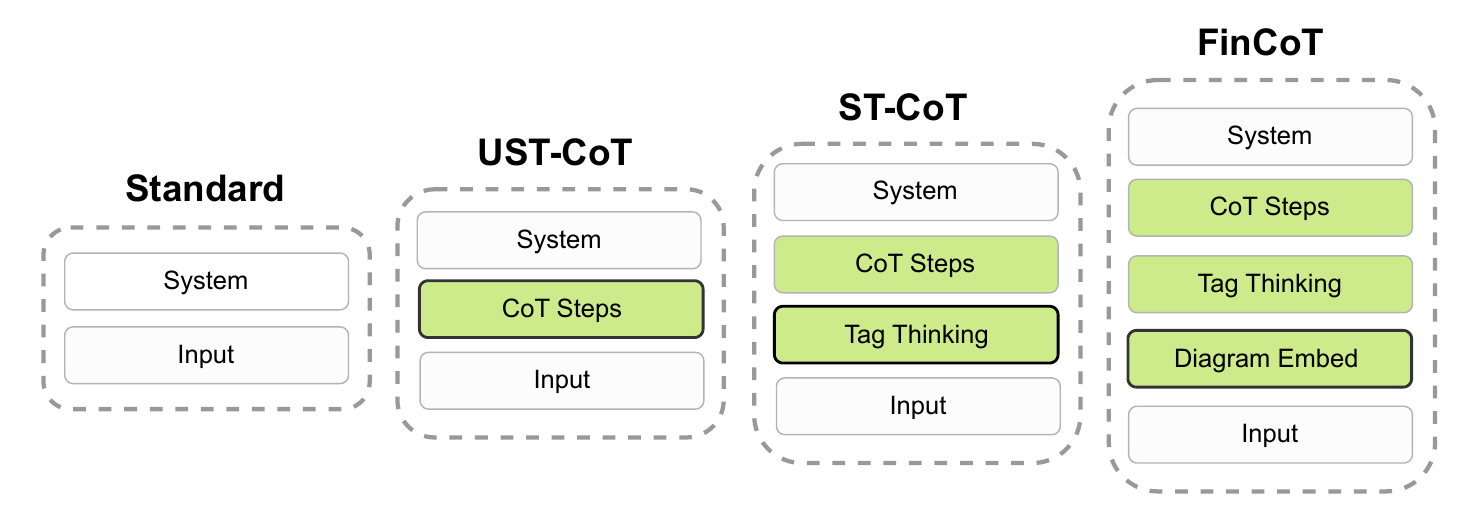}
    \caption{Taxonomy of prompting strategies by reasoning structure: SP, UST-CoT, ST-CoT, and FinCoT. Green blocks indicate added reasoning control--CoT steps, tagged thoughts, and expert diagrams (Diagram Embed)--showing the evolution toward more interpretable, domain-aligned prompts.}
    \label{fig:sp_ust_cot_fincot}
\end{figure*}
\section{Background and Related Work}
\label{sec:related}
\subsection{Prompt Engineering}
\paragraph{Standard Prompting (SP):}\label{mtd:standard_prompt}
Refers to the baseline technique of simply providing a natural language instruction to an LLM, without providing any intermediate \texttt{`thinking'} steps, demonstrations, or explicit reasoning cues-i.e., a zero-shot setup. While the GPT-3 paper~\citep{brown2020language} popularized few-shot prompting via exemplars, more recent work formalizes and benchmarks zero-shot prompting as a distinct paradigm~\citep{wei2022finetunedlanguagemodelszeroshot}. Our implementation follows the formulation shown in Appendix Listing 1 (ZS) in~\cite{callanan-etal-2024-gpt}, and is used to represent the standard prompting baseline.

\paragraph{Unstructured Chain-of-Thought (UST-CoT):}\label{mtd:ust_cot}
A general-purpose reasoning strategy using free-form CoT to establish a baseline for unconstrained prompting. These include:

\noindent\textbf{{\small\textbullet} Chain-of-Thought (CoT)}~\citep{wei2023chainofthoughtpromptingelicitsreasoning}: Decompose reasoning into intermediate steps, encouraging the model to deliberately and systematically \texttt{`think'} before finalizing an answer.

\noindent\textbf{{\small\textbullet} Code Prompting}~\citep{hu2023codepromptingneuralsymbolic}: Translates problems into executable code, allowing the model to simulate logic or perform precise computations. In other words, LLMs are elicited to reason explicitly and transparently through code.

\noindent\textbf{{\small\textbullet} Plan-and-Solve}~\citep{wang-etal-2023-plan}: Separates planning from solving, where the model first outlines a high-level plan, then executes the reasoning based on that plan.

In addition to these, other prompting techniques have emerged, such as Tree of Thoughts (ToT)~\citep{yao2023treethoughtsdeliberateproblem}, which explores multiple reasoning paths via tree-structured search; Graph of Thoughts (GoT)~\citep{Besta_2024}, which frames reasoning as a graph with LLM-generated nodes and edges for flexible information flow. These methods enhance expressiveness for general tasks; they are not tailored for finance, which requires mathematical precision and domain-specific constraints. We adopt the template from Appendix Listing 2 CoT~\cite{callanan-etal-2024-gpt} as the default prompt for this baseline.

\paragraph{Structured Chain-of-Thought (ST-CoT):}\label{mtd:st_cot}
ST-CoT augments a prompt with tags such as \texttt{\allowbreak<\allowbreak thinking\allowbreak>} and \texttt{\allowbreak<\allowbreak output\allowbreak>} that break the model’s response into explicit, modular stages, promoting incremental, easily replaceable reasoning blocks. This tag-driven format has already appeared in open-source trials.\footnote{\url{https://gist.github.com/davidmezzetti/1fac6bd406857431f1cdc74545bdfba9}}  \Cref{fig:sp_ust_cot_fincot} visually contrasts ST-CoT with SP, UST-CoT, and FinCoT.

FinCoT (§\ref{mtd:structured_expertise_step}) inherits ST-CoT's structure but injects domain-specific Mermaid blueprints to ground each step in expert workflows. Unlike Universal Self-Adaptive Prompting~\citep{wan-etal-2023-universal}, which derives few-shot exemplars from LLM memory, FinCoT encodes human-crafted financial reasoning, favoring interpretability and control. The three categories Standard Prompting (direct instruction), Unstructured CoT (free-form reasoning), and Structured CoT (tag-driven with optional expert hints) offer a unified lens for classifying financial prompting.

\subsection{LLMs in Domain-Specific Financial Reasoning}

Existing approaches to adapting large foundation models for financial reasoning currently fall into three broad paradigms:

\paragraph{Prompting-based:} Methods use few-shot prompts with CFA-style queries, as in ``Can GPT Models be Financial Analysts?''~\cite{callanan-etal-2024-gpt}, which evaluate ChatGPT and GPT-4 on mock CFA exams.

\paragraph{Fine-tuning-based:} Methods adapt models via supervised fine-tuning with curated QA/classification datasets~\citep{chen2022finqadatasetnumericalreasoning, harsha-etal-2025-synthetic} or continued pretraining on domain-specific corpora~\citep{yang2020finbertpretrainedlanguagemodel, lee-etal-2024-finale, bhatia2024fintralfamilygpt4level, ke2025demystifyingdomainadaptiveposttrainingfinancial, liu2025finr1largelanguagemodel}. While effective, these approaches rely on labeled data and general language modeling objectives, lacking structuring of financial reasoning, thus limiting interpretability and alignment with expert logic.

Despite advances, prior work has largely overlooked structured reasoning grounded in real-world workflows. We introduce a domain-based prompting framework designed to reflect step-by-step expert logic and evaluate it on CFA-style exam tasks.

\section{FinCoT: Augmenting CoT with Structured Financial Expertise}\label{mtd:structured_expertise_step}
\begin{figure}[!ht]
    \vspace{-1em}
    \centerline{
    \includegraphics[width=\linewidth,keepaspectratio]{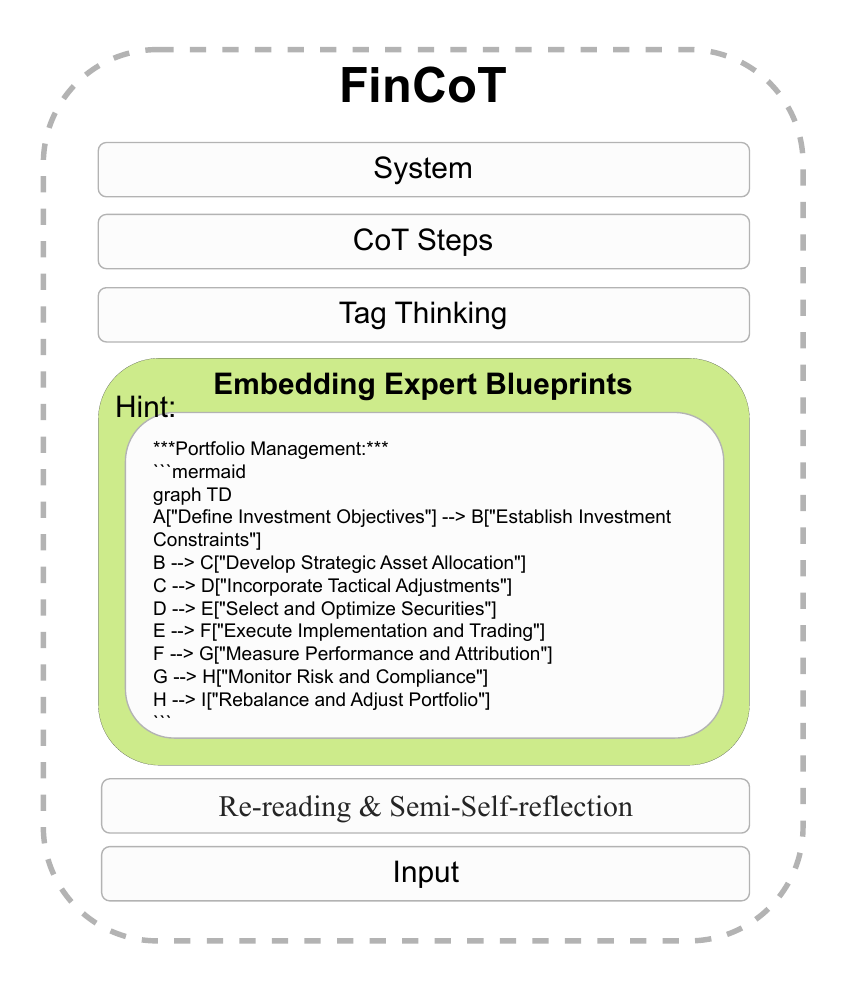}}
    \caption{Overview of FinCoT, integrates structured, expert-guided reasoning layers Mermaid diagrams, plan-and-solve scaffolding, and reflective validation to improve performance in financial tasks.}
    \label{fig:fincoT}
\end{figure}

We introduce FinCoT (Financial Chain-of-Thought), a structured prompting framework that enhances LLM reasoning in specialized financial domains. Building upon ST-CoT approaches, FinCoT explicitly embeds expert-derived problem-solving methodologies directly into prompts, guiding LLMs to follow domain-specific reasoning pathways without requiring model fine-tuning. \Cref{fig:fincoT} illustrates the FinCoT architecture, which integrates expert-guided reasoning layers and reflective validation to improve performance in financial tasks.

\subsection{The FinCoT Prompt Framework}
The FinCoT prompting framework integrates the following key components and logical steps:

\begin{enumerate}[leftmargin=*]
\setlength\itemsep{-0.1em}
\item \textbf{System:} A single, top-level message that frames the task (e.g., ``You are a CFA candidate; treat the following as a finance question'').

\item \textbf{Guided Step-by-Step Execution:} The prompt reserves two tag blocks \texttt{\allowbreak<\allowbreak thinking\allowbreak>} for intermediate reasoning and \texttt{\allowbreak<\allowbreak output\allowbreak>} for the final answer-thereby enforcing a structured chain-of-thought (ST-CoT) in one turn.

\item \textbf{Expert Reasoning Blueprint (via Mermaid Diagram~\citep{mermaid}:} A domain-specific, embedded expert blueprint with Mermaid diagram concerning the generation of diagrams (see §\ref{step:mermaid_gen}), serve as a "Hint" within the context of the prompt. This blueprint explicitly outlines the recommended step-by-step problem-solving strategy for the given financial domain and is curated through a systematic process detailed in~\cref{mtd:finexp_pipeline} to ensure consistency and domain alignment.

\item \textbf{Re-Reading \& Semi Self-Reflection:} Inside the \texttt{\allowbreak<\allowbreak output\allowbreak>} tags, the model briefly checks its reasoning for consistency before committing the final answer. We call this “semi-reflection” because we drop the separate \texttt{<reflection>} block-avoiding per-step scoring and self-bias noted by \citet{xu-etal-2024-pride} yet still include a short self-check within \texttt{\allowbreak<\allowbreak output\allowbreak>}.
  
\end{enumerate}

\subsection{Embedding Expert Blueprints}\label{mtd:finexp_pipeline}
The creation of effective expert reasoning blueprints involves a systematic multistage process designed to transform a wide range of financial knowledge into structured and actionable LLM diagrams. 
\begin{figure}[!ht]
    \centerline{
    \includegraphics[width=\linewidth,keepaspectratio]{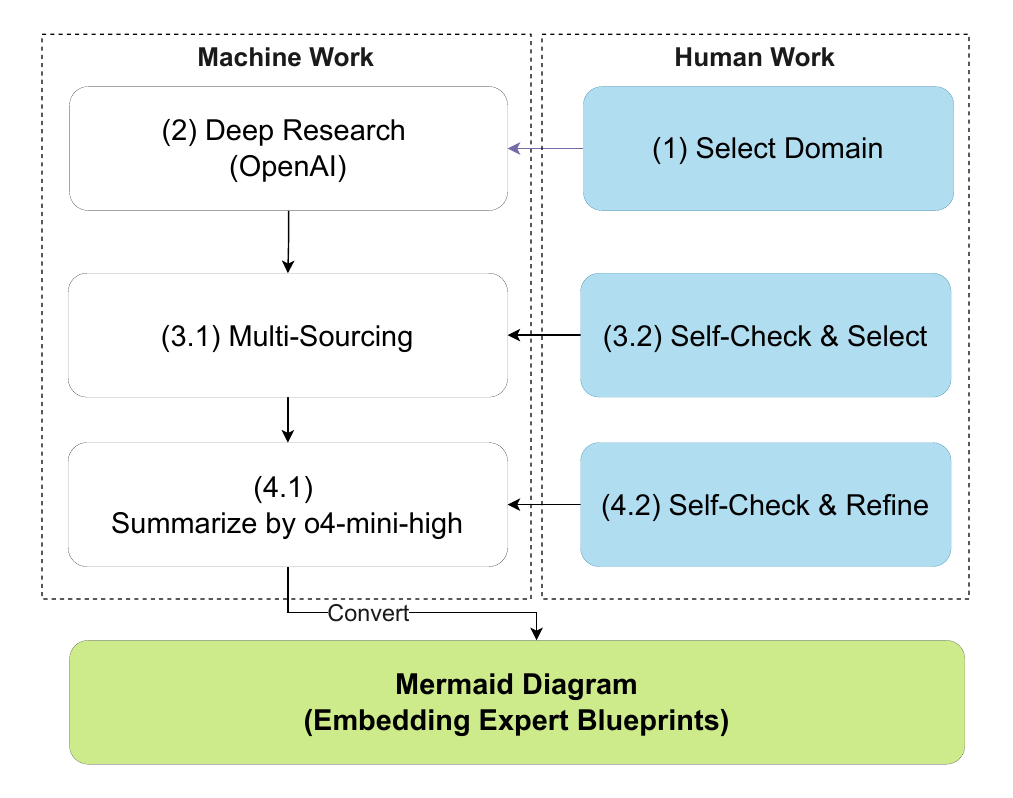}}
    \caption{Pipeline for curating financial expert reasoning. Each stage systematically transforms raw financial knowledge into structured Mermaid diagrams for FinCoT prompting.}
    \label{fig:curation_pipeline}
\end{figure}

\paragraph{Curation Pipeline for Expert Reasoning:}To construct expert blueprints for each financial domain, we implement a hybrid pipeline combining machine assistance and human judgment, as illustrated in \Cref{fig:curation_pipeline}. The process includes the following stages:

\begin{enumerate}[leftmargin=*]
\setlength\itemsep{-0.1em}
\item \textbf{Scope Definition and Knowledge Aggregation:}\label{step:scope_agg}
    The target CFA domain is scoped (e.g., Economics focusing on supply and demand), and relevant expert strategies are aggregated, using Deep Research\footnote{An AI agent for retrieving/synthesizing knowledge from public sources such as CFA notes, academic texts, and industry guides.}, from diverse authoritative sources. Outputs are validated by human-in-the-loop reviewers with financial knowledge (e.g., finance graduates or CFA charterholders) to ensure conceptual accuracy and domain alignment.
    
    \item \textbf{Validation and Synthesis:}\label{step:validation_synth}
    We cross-reference and synthesize the aggregated knowledge to ensure accuracy, identify core principles and filter redundancies. 
    \item \textbf{Iterative Refinement into Structured Workflows:}\label{step:refine_workflow}
    The synthesized expert knowledge is iteratively transformed into logical step-by-step reasoning workflows. This refinement process focuses on ensuring the coherence, correctness, and clarity of the resulting problem-solving strategies for each financial domain.
    
    \item \textbf{Mermaid Diagram Generation:}\label{step:mermaid_gen}
    This refined workflow is translated into a Mermaid diagram~\citep{DeBari2024UML} using its text-based syntax. We selected Mermaid due to its LLM prompt compatibility and clear visual guidance aligning with the FinCoT prompt. The diagrams are constructed based on the source content validated and synthesized first in~\ref{step:validation_synth}, and then applied to each financial domain, with the entire collection organized and described in~\Cref{appendix:blueprint_sources} as reference blueprints\footnote{The resulting expert blueprints are reviewed for conceptual consistency and practical correctness (but not guaranteed precision) by CFA Level III charterholders.}.

    \item \textbf{Prompt Integration:}\label{step:prompt_int}
    The text-based Mermaid blueprint is embedded as ``Hint" within the FinCoT prompt template (Appendix~\ref{appendix:fincot_prompt_template}), directly guiding the LLM's reasoning.
\end{enumerate}
\section{Experimental Setup}
\paragraph{Model Configurations and Inference Parameters:} 
We selected the Qwen family of models due to their strong baseline performance in zero-shot financial reasoning. In preliminary evaluations, \textbf{Qwen2.5-7B-Instruct} achieved 69.7\% accuracy on standard task prompts, substantially outperforming \textbf{Llama3.1-8B-Instruct} (46.3\%), motivating its use as our primary model family. To evaluate the impact of both instruction tuning and domain-specific adaptation, we compare two model groups.
\noindent\textbf{(A) General-purpose foundation models}
\vspace{-0.3em}
\begin{itemize}[leftmargin=1.25em]
\setlength\itemsep{0pt}
  \item \textbf{Qwen2.5-7B (pretrained model)} vs.\ \textbf{Qwen2.5-7B-Instruct}: to assess the effect of instruction tuning on a strong general-purpose foundation model.
  \item \textbf{Qwen3-8B-Base (pretrained model)} vs. \textbf{Qwen3-8B}, \textbf{Qwen3-8B (Thinker)}: to isolate the impact of ST-CoT prompting within the same architecture.
  \item \textbf{Gemma-3-12B-IT (Text-only)}: an instruction-tuned model recently released, comparable in size to Qwen3-8B. It achieved 52.81\% accuracy on the Flare-CFA benchmark, outperforming Llama3.1-8B-Instruct, and serves as a competitive baseline.
\end{itemize}
\vspace{-0.3em}
\noindent\textbf{(B) Financial-specific reasoning models}
\vspace{-0.3em}
\begin{itemize}[leftmargin=1.25em]
\setlength\itemsep{0pt}
  \item \textbf{Fin-R1 (7B)}: adapted from Qwen2.5-7B-Instruct using supervised and reinforcement learning on a financial dataset distilled from DeepSeek-R1~\citep{liu2025finr1largelanguagemodel}.
  \item \textbf{DianJin-R1-7B}: fine-tuned from Qwen2.5-7B-Instruct using CFLUE, FinQA, and CCC, with GRPO to improve domain-specific reasoning~\citep{zhu2025dianjinr1evaluatingenhancingfinancial}.
  \item \textbf{Fin-o1-8B}: built on Qwen3-8B and trained on the FinCoT\footnote{The FinCoT dataset is constructed by TheFinAI and publicly available at \url{https://huggingface.co/datasets/TheFinAI/FinCoT}. It combines financial QA datasets such as FinQA, ConvFinQA, TATQA, DocMath-Eval, Econ-Logic, BizBench-QA, and DocFinQA, with GPT-4o-generated reasoning traces to enhance structured financial question answering. Note that this dataset is not derived from our prompting approach.} corpus using SFT and GRPO, setting a strong benchmark in financial reasoning~\citep{qian2025fino1transferabilityreasoningenhancedllms}.
\end{itemize}

All experiments used a maximum sequence length of 16.384k tokens. Following best practices for decoding stability~\citep{du2025optimizingtemperaturelanguagemodels}, we set the generation temperature to 0.2 to encourage focused and consistent outputs under evaluation conditions.

\paragraph{Prompting Strategies Compared:}  
Our study evaluates the effectiveness of FinCoT against three baseline prompting strategies: SP, UST-CoT, and ST-CoT, which were detailed in \Cref{sec:related}. For clarity in this section:
\begin{itemize}[leftmargin=*]
\setlength\itemsep{-0.1em}
\item \textbf{SP:} The model receives only the target question.
\item \textbf{UST-CoT:} In addition to the question, the model is given a generic cue to reason step-by-step.
\item \textbf{ST-CoT:} This strategy employs structural tags (e.g., \texttt{\allowbreak<\allowbreak thinking\allowbreak>}, \texttt{\allowbreak<\allowbreak output\allowbreak>}) to guide the model in generating an organized step-by-step reasoning trace for the target question.
\item \textbf{FinCoT:} A zero-shot prompting method that integrates expert domain templates (excluding the Ethics domain). Each prompt includes a Mermaid diagram as a ``Hint'' to guide structured financial reasoning via relevant domain insights.
\end{itemize}

While all methods operate in a zero-shot setting, FinCoT uniquely injects expert-guided structure through diagrams. Recent studies suggest that even large reasoning models may struggle with instruction-following when overloaded with reasoning cues~\citep{li2025thinkingfailspitfallsreasoning, fu2025scalingreasoninglosingcontrol, jang2025reasoningmodelstubborndiagnosing, yao2025reasoningmodelspronehallucination}, though this remains underexplored in financial contexts. We thus evaluate whether CoT-style prompts (UST-CoT, ST-CoT, FinCoT) enhance instruction-following compared to SP.

\paragraph{Evaluation Benchmark:}\label{cfa_eval}
To assess financial reasoning, we use 1.032k multiple-choice questions from the CFA-Easy subset of FinEval (also referred to as Flare-CFA), originally introduced by~\citet{ke2025demystifyingdomainadaptiveposttrainingfinancial}. This curated set reflects the rigor of CFA exams and enables evaluation across both theoretical and practical domains. Each question is categorized into one of ten CFA domains using GPT-4o with a dedicated classification prompt (see Appendix~\ref{appendix:classify_domain_prompt}), and Figure~\ref{appendix:fig_domain_distribution} shows the resulting domain distribution.

\paragraph{Evaluation Metrics:}
We report \textbf{accuracy} as the metric, defined as the percentage of questions where the model's prediction matches the ground truth. To assess response efficiency, we also report the \textbf{average output length} (in tokens) across questions. For statistical significance, we use a paired bootstrap test~\citep{efron1994introduction} with $B{=}10k$ resamples over binary correctness scores, reporting the mean difference ($\Delta$), 95\% confidence interval, and $p$-value. Additionally, we measure the proportion of financial domains where a method improves accuracy by at least 1\% over SP and compute the average domain-wise accuracy gain.

\section{Results and Discussions}
\subsection{Financial Reasoning Performance}
\begin{table*}[htbp]
  \centering
  \scriptsize
  \setlength\tabcolsep{1.2pt}
  \renewcommand\arraystretch{1.4}
  \begin{tabular}{lccccccccc}
    \toprule
    \multirow{3}{*}{\textbf{Prompt}} &
    \multicolumn{9}{c}{\textbf{Accuracy (\%)}} \\
    \cmidrule(lr){2-10}
    & \multicolumn{6}{c}{\textbf{General models}} 
    & \multicolumn{3}{c}{\textbf{Financial models}} \\
    \cmidrule(lr){2-7} \cmidrule(lr){8-10}
    & \makecell{Qwen2.5-7B}
    & \makecell{Qwen2.5-7B\\Instruct}
    & \makecell{Qwen3-8B\\Base}
    & Qwen3-8B
    & \makecell{Qwen3-8B\\(Thinker)}
    & \makecell{Gemma-3-12B\\IT}
    & \makecell{Fin-R1\\7B}
    & \makecell{DianJin-R1\\7B}
    & \makecell{Fin-o1\\8B} \\
    \midrule
    SP &
      54.07 &
      69.67 &
      63.18 &
      74.42 &
      88.18 &
      52.81 &
      65.70 &
      78.39 &
      \textbf{79.65} \\
    UST-CoT &
      67.83 \goodtxt{(\up13.76)} &
      \textbf{75.68}* \goodtxt{(\up6.01)} &
      72.58 \goodtxt{(\up9.40)} &
      \textbf{82.36}* \goodtxt{(\up7.94)} &
      \textbf{89.05}* \goodtxt{(\up0.87)} &
      \textbf{77.81}* \goodtxt{(\up25.00)} &
      75.19 \goodtxt{(\up9.49)} &
      67.73 \badtxt{(\down10.66)} &
      79.36 \badtxt{(\down0.29)} \\
    ST-CoT &
      \textbf{70.35}* \goodtxt{(\up16.28)} &
      74.52 \goodtxt{(\up4.85)} &
      78.49 \goodtxt{(\up15.31)} &
      81.01 \goodtxt{(\up6.59)} &
      88.18 &
      76.74 \goodtxt{(\up23.93)} &
      74.32 \goodtxt{(\up8.62)} &
      68.80 \badtxt{(\down9.59)} &
      78.39 \badtxt{(\down1.26)} \\
      FinCoT &
      62.02 \goodtxt{(\up7.95)} &
      74.22 \goodtxt{(\up4.55)} &
      \textbf{80.52}* \goodtxt{(\up17.34)} &
      81.10 \goodtxt{(\up6.68)} &
      87.21 \badtxt{(\down0.97)} &
      75.58 \goodtxt{(\up22.77)} &
      \textbf{75.78}* \goodtxt{(\up10.08)} &
      \textbf{79.75}* \goodtxt{(\up1.36)} &
      77.23 \badtxt{(\down2.42)} \\
    \midrule
    \rowcolor{gray!16}
    \multicolumn{10}{c}{\textbf{Domain-wise performance of FinCoT}} \\
    \midrule
    Economics    & 
      69.09 \goodtxt{(\up15.02)} & 
      73.26 \goodtxt{(\up3.59)}  & 
      79.26 \goodtxt{(\up16.08)} & 
      79.55 \goodtxt{(\up5.13)}  & 
      86.92 \badtxt{(\down1.26)} & 
      74.61 \goodtxt{(\up21.80)} & 
      73.45 \goodtxt{(\up7.75)}  & 
      55.52 \badtxt{(\down22.87)} & 
      78.00 \badtxt{(\down1.65)} \\
    
    FixedIncome  & 
      68.12 \goodtxt{(\up14.05)} & 
      73.35 \goodtxt{(\up3.68)}  & 
      78.88 \goodtxt{(\up15.70)} & 
      80.81 \goodtxt{(\up6.39)}  & 
      87.21 \badtxt{(\down0.97)} & 
      76.45 \goodtxt{(\up23.64)} & 
      74.22 \goodtxt{(\up8.52)}  & 
      66.86 \badtxt{(\down11.53)} & 
      76.74 \badtxt{(\down2.91)} \\
    
    Quant.Meth.   & 
      68.02 \goodtxt{(\up13.95)} & 
      \textbf{75.19} \goodtxt{(\up5.52)} & 
      80.14 \goodtxt{(\up16.96)} & 
      80.91 \goodtxt{(\up6.49)}  & 
      \textbf{87.79} \badtxt{(\down0.39)} & 
      75.68 \goodtxt{(\up22.87)} & 
      74.90 \goodtxt{(\up9.20)}  & 
      65.79 \badtxt{(\down12.60)} & 
      77.42 \badtxt{(\down2.23)} \\
    
    EquityInvest. & 
      69.09 \goodtxt{(\up15.02)} & 
      74.22 \goodtxt{(\up4.55)}  & 
      79.26 \goodtxt{(\up16.08)} & 
      80.52 \goodtxt{(\up6.10)}  & 
      86.72 \badtxt{(\down1.46)} & 
      76.45 \goodtxt{(\up23.64)} & 
      74.42 \goodtxt{(\up8.72)}  & 
      62.31 \badtxt{(\down16.08)} & 
      78.68 \badtxt{(\down0.97)} \\
    
    Port.Manage.   & 
      67.54 \goodtxt{(\up13.47)} & 
      74.13 \goodtxt{(\up4.46)}  & 
      \textbf{80.72} \goodtxt{(\up17.54)} & 
      80.91 \goodtxt{(\up6.49)}  & 
      86.92 \badtxt{(\down1.26)} & 
      77.03 \goodtxt{(\up24.22)} & 
      75.00 \goodtxt{(\up9.30)}  & 
      62.02 \badtxt{(\down16.37)} & 
      76.55 \badtxt{(\down3.10)} \\
    
    Derivatives    & 
      68.90 \goodtxt{(\up14.83)} & 
      73.64 \goodtxt{(\up3.97)}  & 
      79.84 \goodtxt{(\up16.66)} & 
      80.81 \goodtxt{(\up6.39)}  & 
      87.21 \badtxt{(\down0.97)} & 
      \textbf{77.23} \goodtxt{(\up24.42)} & 
      \textbf{76.16} \goodtxt{(\up10.46)} & 
      \textbf{71.80} \badtxt{(\down6.59)}  & 
      \textbf{79.94} \goodtxt{(\up0.29)}  \\
    
    Fin. Reporting & 
      \textbf{69.28} \goodtxt{(\up15.21)} & 
      73.84 \goodtxt{(\up4.17)}  & 
      79.07 \goodtxt{(\up15.89)} & 
      \textbf{81.10} \goodtxt{(\up6.68)} & 
      87.02 \badtxt{(\down1.16)} & 
      75.87 \goodtxt{(\up23.06)} & 
      72.87 \goodtxt{(\up7.17)}  & 
      62.69 \badtxt{(\down15.70)} & 
      77.23 \badtxt{(\down2.42)} \\
    
    Alter.Invest.  & 
      68.99 \goodtxt{(\up14.92)} & 
      74.90 \goodtxt{(\up5.23)}  & 
      78.97 \goodtxt{(\up15.79)} & 
      79.94 \goodtxt{(\up5.52)}  & 
      87.50 \badtxt{(\down0.68)} & 
      76.94 \goodtxt{(\up24.13)} & 
      74.90 \goodtxt{(\up9.20)}  & 
      56.98 \badtxt{(\down21.41)} & 
      78.88 \badtxt{(\down0.77)} \\
    
    Corp.Issuers   & 
      68.31 \goodtxt{(\up14.24)} & 
      74.32 \goodtxt{(\up4.65)}  & 
      79.26 \goodtxt{(\up16.08)} & 
      79.36 \goodtxt{(\up4.94)}  & 
      87.02 \badtxt{(\down1.16)} & 
      \textbf{77.23} \goodtxt{(\up24.42)} & 
      75.58 \goodtxt{(\up9.88)}  & 
      60.08 \badtxt{(\down18.31)} & 
      79.07 \badtxt{(\down0.58)} \\
    
    \bottomrule
  \end{tabular}
  \caption{Comparison of accuracy (\%) of prompting techniques. `FinCoT' simultaneously applies expert reasoning blueprints from all CFA domains, while each `(DomainName)' (e.g., `Economics') row applies domain-specific blueprints individually. (\up/\down) Denote accuracy improvement or decline relative to the SP baseline, colored \goodtxt{green} for (\up) and \badtxt{red} for (\down). \textbf{Bold} values highlight the best-performing prompt variant for each model. (*) Indicates that the accuracy improvement among the model-level prompt variants is statistically significant ($p<0.05$) based on paired bootstrap testing; domain-specific rows are not tested for significance.}
  \label{tab:accuracy}
\end{table*}
\paragraph{Baseline Performance:}~\Cref{tab:accuracy} reports zero-shot accuracies for four prompting strategies (SP, UST-CoT, ST-CoT, FinCoT) across our model suite. Under the basic SP prompt, the instruction-tuned \textbf{Qwen3-8B (Thinker)} attains the highest accuracy among general-purpose models (88.18\%), while the financial model \textbf{Fin-o1-8B} leads its group at 79.65\%. These strong baselines clearly highlight the effectiveness of instruction tuning and domain specificity.
\paragraph{Pretrained Models:} On \textbf{Qwen2.5-7B}, FinCoT (All Blueprints) yields a +7.95~pp improvement over SP (95\% CI [6.30, 9.59], $p<0.001$), while UST-CoT and ST-CoT also exceed the baseline by +13.76~pp and +16.28~pp. When FinCoT is applied using a single-domain blueprint (e.g., Financial Reporting), the gains increase substantially to +15.21~pp. Similarly, on \textbf{Qwen3-8B-Base}, FinCoT delivers the strongest overall boost (+17.35~pp, 95\% CI [15.02, 19.77], $p<0.001$). These findings further underscore the importance of structured domain knowledge, particularly for models lacking instruction or domain alignment.
\paragraph{Instruction Models:} We evaluate three prominent instruction-tuned variants. On \textbf{Qwen2.5-7B-Instruct}, FinCoT improves accuracy by +4.55~pp over SP, compared to -1.46~pp with UST-CoT and -0.3~pp with ST-CoT. On \textbf{Qwen3-8B (Thinker)}, FinCoT yields a slight drop (–0.96~pp), while ST-CoT shows no change and UST-CoT yields a modest +0.87~pp. \textbf{Gemma-3-12B-IT}, a strong instruction-tuned baseline (52.81\% SP), benefits substantially from all strategies: +25.00~pp (UST-CoT), +23.93~pp (ST-CoT), and +22.77~pp (FinCoT). Notably, domain-specific FinCoT prompts (e.g., Derivatives, Corporate Issuers) provide even larger boosts (+24.42~pp), indicating that blueprint reasoning complements instruction tuning by addressing specialized financial gaps.
\paragraph{Financial-Specific Models:} FinCoT also helps specialized models like \textbf{Fin-R1}, confirming that blueprint prompting provides complementary gains beyond fine-tuning. However, for models with strong built-in reasoning, such as \textbf{DianJin‑R1‑7B} and \textbf{Fin‑o1‑8B}, FinCoT offers limited improvement or slight degradation—likely due to conflicts between external scaffolds and internal reasoning routines. These outcomes suggest diminishing returns for CoT prompting when domain alignment and reasoning are already deeply encoded.

Overall, FinCoT is most impactful for models lacking prior task-specific adaptation. By grounding reasoning in structured financial workflows, it bridges key gaps in zero-shot settings without requiring additional tuning. This pattern highlights a trade-off between model internalization and prompt-time controllability. \textit{Future work could explore hybrid strategies that adapt prompting depth based on model alignment.}

\begin{figure}[!ht]
  \centering
  \begin{subfigure}[b]{0.48\textwidth}
    \centering
    \includegraphics[width=\linewidth,keepaspectratio]{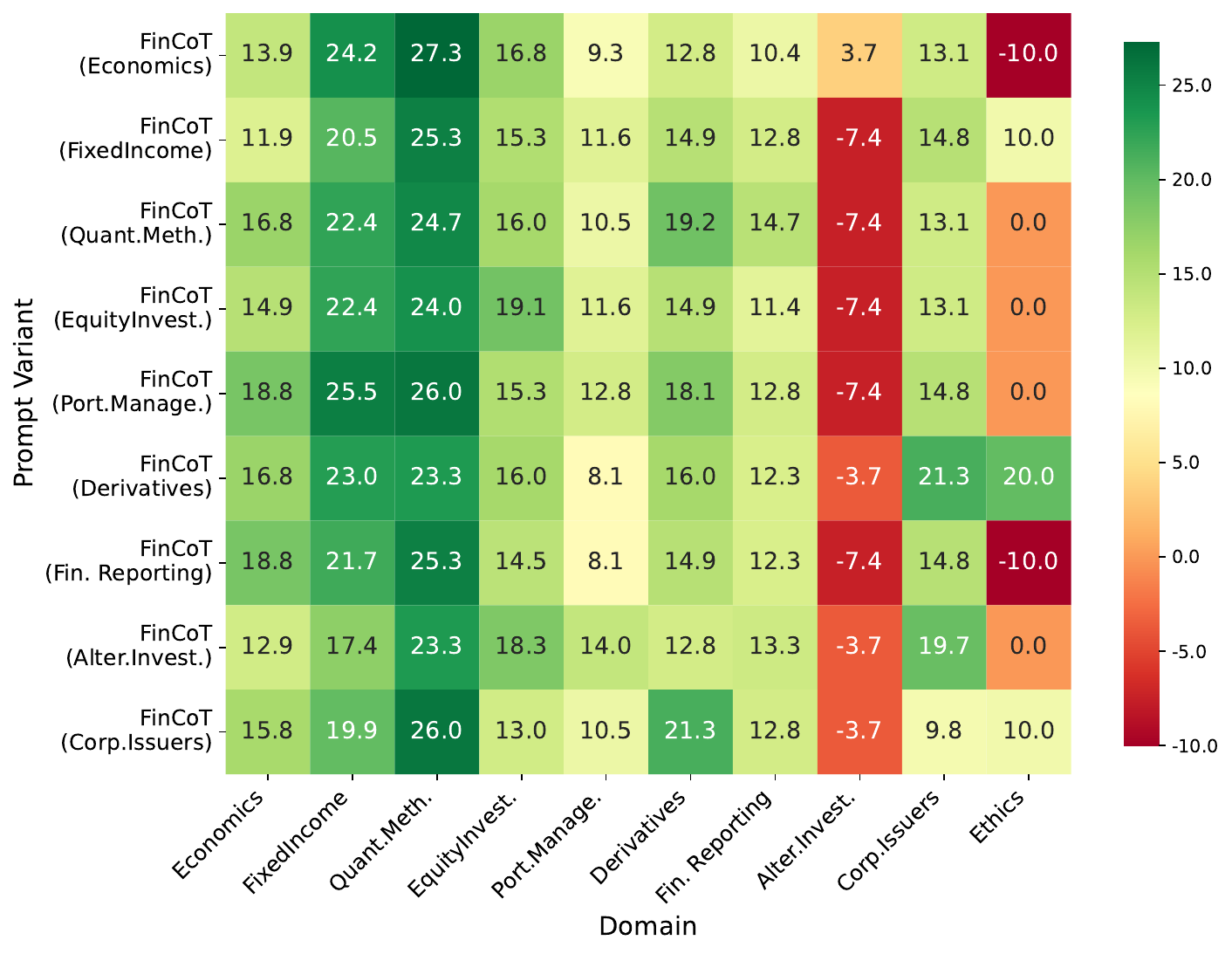}
    \caption{}
    \label{fig:a_across_domain_a}
  \end{subfigure}%
  \hfill
  \begin{subfigure}[b]{0.48\textwidth}
    \centering
    \includegraphics[width=\linewidth,keepaspectratio]{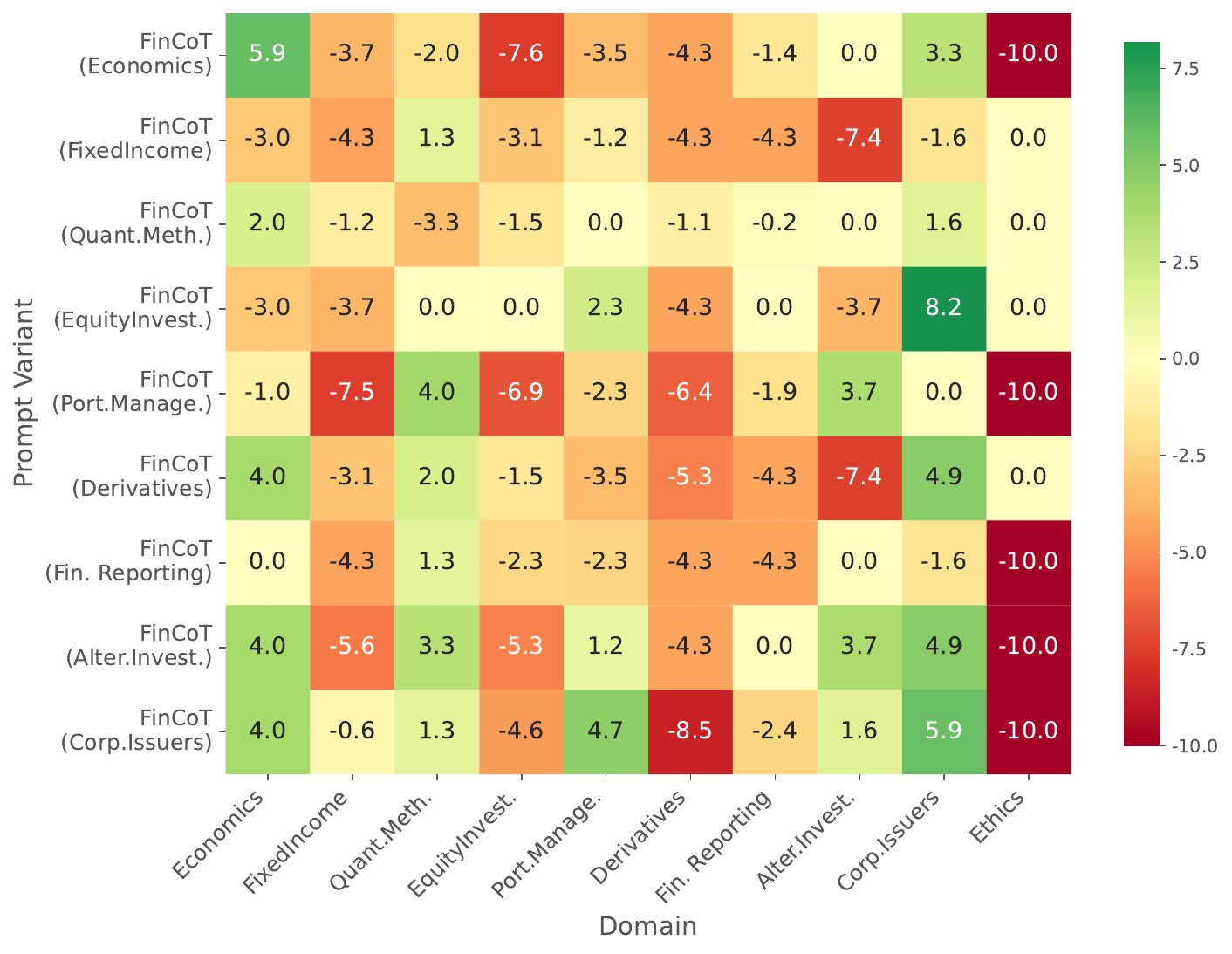}
    \caption{}
    \label{fig:b_across_domain_b}
  \end{subfigure}
  \caption{Accuracy improvements (\%) of each FinCoT domain-specific prompt compared to Standard Prompting (SP). Subfigure (a) shows results on Qwen3-8B-Base (pretrained), while (b) shows results on Fin-o1-8B (finance-specific).}
\end{figure}
\paragraph{Cross-domain behavior of FinCoT:}\label{eval:cross_domain_characteristics}
This section examines how pretrained and finance-specific models respond to structured prompting with FinCoT. Each domain-specific blueprint is applied across all CFA domains to evaluate its transferability, and accuracy differences relative to SP are measured. Figure~\ref{fig:a_across_domain_a} and~\ref{fig:b_across_domain_b} visualize results for Qwen3-8B-Base (pretrained) and Fin-o1-8B (finance-specific), while Table~\ref{tab:accuracy} provides a comprehensive summary of overall model-level accuracy.
On Qwen3-8B-Base, FinCoT generally improves performance, though gains are not universal. Prompts from quantitative domains such as \emph{Derivatives}, \emph{Portfolio Management}, and \emph{Corporate Issuers} yield average gains exceeding +13~pp. The blueprint structure provides inductive guidance that enhances decomposition, formula selection, and financial term alignment. In several domains, Qwen3-8B-Base with FinCoT matches or surpasses the SP baseline of Fin-o1-8B, despite the absence of any task-specific training.

On Fin-o1-8B, gains from FinCoT are more modest, typically within the +1–4~pp range. Minor declines appear in some domains (e.g., Fixed Income, Equity Investments), suggesting that additional scaffolding may interfere with optimized internal reasoning acquired during fine-tuning. Structured prompts may over-specify solutions or reduce instruction-following flexibility.

These findings highlight FinCoT's complementary role. For pretrained models, FinCoT acts as a lightweight yet effective augmentation layer at inference time, reducing the performance gap with fine-tuned models. For already fine-tuned models, careful prompt selection or adaptation may be necessary to preserve existing reasoning strengths without introducing conflict. A broader breakdown of FinCoT performance across additional models is provided in Appendix~\ref{appendix:radar_behavior_accuracy}, where radar plots illustrate domain-wise patterns and complement the main analysis.

\begin{figure*}[htbp] 
  \centering
  \includegraphics[width=1.0\textwidth]{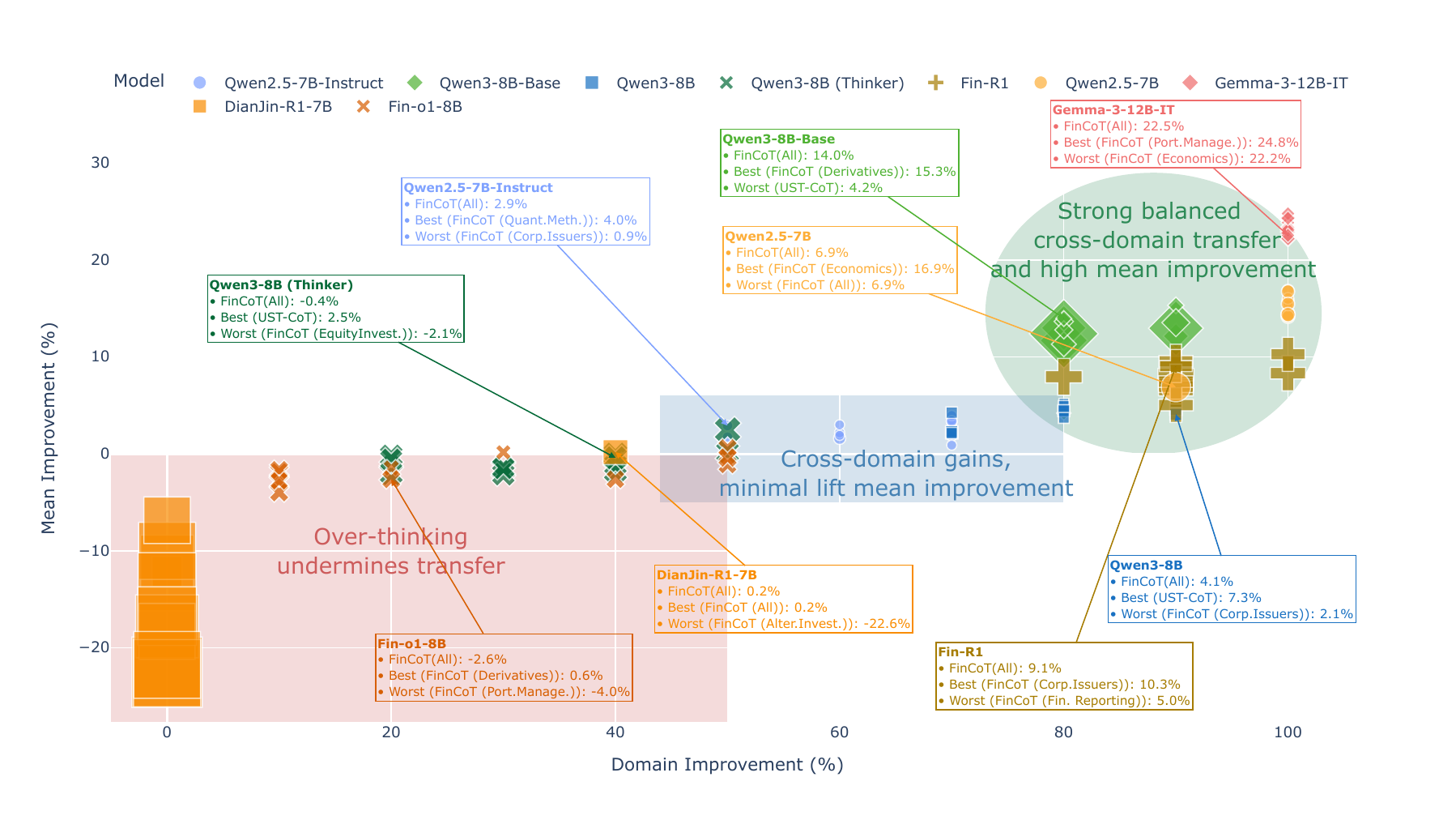}
  \caption{Comparison of prompt strategies across financial domains. Each point represents a domain-method pair, with position indicating accuracy improvement and domain coverage. Circle size encodes generated tokens.}
  \label{fig:across_domain_improvement}
\end{figure*}

\subsection{Efficiency Analysis}\label{eval:efficiency_analysis}
Effective deployment of foundation models in financial settings requires balancing verbosity and accuracy while emphasizing efficiency. FinCoT offers a prompt-based alternative to fine-tuned models (Fin-o1, DianJin-R1, Fin-R1), demonstrating similar accuracy as seen in Fig.~\ref{fig:across_domain_improvement}, with token lengths detailed in Appendix~\ref{appendix:tab_average_output_tokens} (Tab.~\ref{tab:output_tokens}). Our analysis concentrates on output tokens since prompt encoding occurs once with parallel self-attention $O(n_{\text{in}}^{2})$ (rapid)~\citep{vaswani2023attentionneed}, while decoding involves $O(n_{\text{out}})$ sequential steps with per-step KV-cache updates (high memory usage)~\citep{gu2018nonautoregressiveneuralmachinetranslation,shazeer2019fasttransformerdecodingwritehead}. Comprehensive input and output token data are provided in Appendix~\ref{appendix:tab_average_input_tokens}(Tab.~\ref{tab:input_tokens}), and recent decoding accelerations such as speculative sampling~\citep{chen2023acceleratinglargelanguagemodel} and FlashAttention~\citep{dao2022flashattentionfastmemoryefficientexact} further underscore output as the primary latency driver.

\paragraph{Output Token Length vs. Accuracy:} In deployments, prompting strategies that achieve high accuracy with minimal output length are desirable.

For general-purpose models such as Qwen3-8B-Base and Qwen3-8B (Thinker), FinCoT reduces output length while preserving or improving accuracy. On Qwen3-8B-Base, FinCoT improves accuracy from 78.49\% (ST-CoT) to 80.52\% (+2.03\,pp) while reducing average output from 3.42k to 0.38k tokens, an 8.9$\times$ compression. On Qwen3-8B (Thinker), FinCoT maintains 88.18\% accuracy with tokens dropping from 1.35k to 1.23k ($\sim$1.1$\times$). Results show that FinCoT’s structured blueprints enable more concise, focused reasoning in general-purpose models. Among financial-specific models such as DianJin-R1-7B and Fin-o1-8B, FinCoT shows minimal or negative improvement and little benefit in output compression, consistent with fine-tuned models internalizing domain-specific reasoning. Fin-o1-8B suggests that excessive prompt scaffolding may interfere with latent reasoning, reducing effectiveness and leading to overthinking that undermines transfer.

Three behavioral zones emerge from these results. First, models like Gemma-3-12B-IT and Qwen3-8B-Base show high mean improvement and strong cross-domain transfer, reflecting effective generalization. Second, models such as Qwen2.5-7B-Instruct and Qwen3-8B display noticeable cross-domain transfer with lower mean improvement, suggesting limited benefit. Third, models like Fin-o1-8B and Qwen3-8B (Thinker) exhibit low cross-domain adaptability and minimal performance lift, indicating that overly detailed prompting may conflict with internal reasoning.

These findings underscore the importance of aligning the prompting strategy with a model’s pretraining or fine-tuning to optimize performance and efficiency. \textit{With limited supervision, FinCoT provides a transparent, cost-effective alternative for enhancing financial reasoning.} For a price–sensitivity analysis of token cost, see Appendix~\ref{sec:efficiency-simulation} and Fig.~\ref{fig:efficiency-input-output}.

\section{Conclusion}
We presented \textbf{FinCoT}, a zero-shot prompting framework that embeds expert-curated Mermaid diagrams within structured chain-of-thought scaffolds. By grounding reasoning in domain logic, FinCoT bridges human financial workflows with LLM outputs, without model fine-tuning. While broadly applicable, FinCoT yields strong gains for general models (e.g., Qwen3-8B-Base), but more modest or negative effects for instruction or finance-specific models (e.g., Qwen-Thinker, Fin-o1, DianJin-R1), where added structure may interfere with learned reasoning.

Relative to SP, FinCoT improves Qwen3-8B-Base by +17.33 pp and Fin-R1 by +10.08 pp ($p<0.001$), outperforming some fine-tuned models. In contrast, instruction-tuned models like Qwen3-8B (Thinker) sometimes favor UST-CoT. Cross-domain results show blueprints from quantitative fields transfer best (+27.3 pp on Qwen3-8B-Base). FinCoT also reduces token output by up to $8\times$, offering interpretable, efficient prompting for regulated financial applications. Ultimately, FinCoT suggests that with meticulous prompt design, even general-purpose LLMs can approach the reasoning quality of fine-tuned financial experts in complex decision-making tasks.

\section*{Limitations}
Our evaluation highlights several limitations.  
(i) Efficiency gains come mainly from reduced output tokens, but larger inputs from structured templates add cost; overall, FinCoT is still competitive, especially in long-form reasoning.  
(ii) Domain routing via pre-classification risks template mismatch despite safeguards; adaptive selection methods are needed.  
(iii) Improvements are uneven, with domains like Alternative Investments and Ethics limited by small samples ($\sim$10–30); larger, balanced benchmarks are required.  
(iv) Blueprint creation requires expert effort ($\sim$2 hours/domain), and current evaluations are multiple-choice with rubric-based interpretability.

Overall, FinCoT offers structured, auditable reasoning rather than replacing fine-tuned models. Its blueprint methodology and plug-and-play usability demonstrate prompt-level supervision as lightweight knowledge distillation with potential for law, medicine, and engineering.

\section*{Acknowledgments}
We extend our gratitude to the Hatari Team for generously providing additional GPU resources, which significantly enhanced our computational capacity and enabled the successful execution of large-scale experiments. We also thank Tuntai Boriboonthana (CFA Charterholder) from InnovestX for his expert contributions in verifying blueprint fidelity and conducting random audits to ensure domain classification consistency in GPT-4o outputs. Their support and insights were invaluable to the development and evaluation of this work.

\bibliography{custom}

@misc{vaswani2023attentionneed,
      title={Attention Is All You Need}, 
      author={Ashish Vaswani and Noam Shazeer and Niki Parmar and Jakob Uszkoreit and Llion Jones and Aidan N. Gomez and Lukasz Kaiser and Illia Polosukhin},
      year={2023},
      eprint={1706.03762},
      archivePrefix={arXiv},
      primaryClass={cs.CL},
      url={https://arxiv.org/abs/1706.03762}, 
}

@misc{gu2018nonautoregressiveneuralmachinetranslation,
      title={Non-Autoregressive Neural Machine Translation}, 
      author={Jiatao Gu and James Bradbury and Caiming Xiong and Victor O. K. Li and Richard Socher},
      year={2018},
      eprint={1711.02281},
      archivePrefix={arXiv},
      primaryClass={cs.CL},
      url={https://arxiv.org/abs/1711.02281}, 
}

@misc{shazeer2019fasttransformerdecodingwritehead,
      title={Fast Transformer Decoding: One Write-Head is All You Need}, 
      author={Noam Shazeer},
      year={2019},
      eprint={1911.02150},
      archivePrefix={arXiv},
      primaryClass={cs.NE},
      url={https://arxiv.org/abs/1911.02150}, 
}

@misc{chen2023acceleratinglargelanguagemodel,
      title={Accelerating Large Language Model Decoding with Speculative Sampling}, 
      author={Charlie Chen and Sebastian Borgeaud and Geoffrey Irving and Jean-Baptiste Lespiau and Laurent Sifre and John Jumper},
      year={2023},
      eprint={2302.01318},
      archivePrefix={arXiv},
      primaryClass={cs.CL},
      url={https://arxiv.org/abs/2302.01318}, 
}

@misc{dao2022flashattentionfastmemoryefficientexact,
      title={FlashAttention: Fast and Memory-Efficient Exact Attention with IO-Awareness}, 
      author={Tri Dao and Daniel Y. Fu and Stefano Ermon and Atri Rudra and Christopher Ré},
      year={2022},
      eprint={2205.14135},
      archivePrefix={arXiv},
      primaryClass={cs.LG},
      url={https://arxiv.org/abs/2205.14135}, 
}

@misc{wen2025enhancingreasoningadaptlarge,
      title={Enhancing Reasoning to Adapt Large Language Models for Domain-Specific Applications}, 
      author={Bo Wen and Xin Zhang},
      year={2025},
      eprint={2502.04384},
      archivePrefix={arXiv},
      primaryClass={cs.CL},
      url={https://arxiv.org/abs/2502.04384}, 
}

@misc{lewkowycz2022solvingquantitativereasoningproblems,
      title={Solving Quantitative Reasoning Problems with Language Models}, 
      author={Aitor Lewkowycz and Anders Andreassen and David Dohan and Ethan Dyer and Henryk Michalewski and Vinay Ramasesh and Ambrose Slone and Cem Anil and Imanol Schlag and Theo Gutman-Solo and Yuhuai Wu and Behnam Neyshabur and Guy Gur-Ari and Vedant Misra},
      year={2022},
      eprint={2206.14858},
      archivePrefix={arXiv},
      primaryClass={cs.CL},
      url={https://arxiv.org/abs/2206.14858}, 
}

@misc{li2025thinkingfailspitfallsreasoning,
  title={When Thinking Fails: The Pitfalls of Reasoning for Instruction-Following in LLMs},
  author={Xiaomin Li and Zhou Yu and Zhiwei Zhang and Xupeng Chen and Ziji Zhang and Yingying Zhuang and Narayanan Sadagopan and Anurag Beniwal},
  year={2025},
  eprint={2505.11423},
  archivePrefix={arXiv},
  primaryClass={cs.CL}
}

@misc{fu2025scalingreasoninglosingcontrol,
  title={Scaling Reasoning, Losing Control: Evaluating Instruction Following in Large Reasoning Models},
  author={Tingchen Fu and Jiawei Gu and Yafu Li and Xiaoye Qu and Yu Cheng},
  year={2025},
  eprint={2505.14810},
  archivePrefix={arXiv},
  primaryClass={cs.CL}
}

@misc{jang2025reasoningmodelstubborndiagnosing,
  title={Reasoning Model is Stubborn: Diagnosing Instruction Overriding in Reasoning Models},
  author={Doohyuk Jang and Yoonjeon Kim and Chanjae Park and Hyun Ryu and Eunho Yang},
  year={2025},
  eprint={2505.17225},
  archivePrefix={arXiv},
  primaryClass={cs.AI}
}

@misc{yao2025reasoningmodelspronehallucination,
  title={Are Reasoning Models More Prone to Hallucination?},
  author={Zijun Yao and Yantao Liu and Yanxu Chen and Jianhui Chen and Junfeng Fang and Lei Hou and Juanzi Li and Tat-Seng Chua},
  year={2025},
  eprint={2505.23646},
  archivePrefix={arXiv},
  primaryClass={cs.CL}
}

@misc{qian2025fino1transferabilityreasoningenhancedllms,
      title={Fino1: On the Transferability of Reasoning-Enhanced LLMs and Reinforcement Learning to Finance}, 
      author={Lingfei Qian and Weipeng Zhou and Yan Wang and Xueqing Peng and Han Yi and Yilun Zhao and Jimin Huang and Qianqian Xie and Jian-yun Nie},
      year={2025},
      eprint={2502.08127},
      archivePrefix={arXiv},
      primaryClass={cs.CL},
      url={https://arxiv.org/abs/2502.08127}, 
}

@misc{zhu2025dianjinr1evaluatingenhancingfinancial,
      title={DianJin-R1: Evaluating and Enhancing Financial Reasoning in Large Language Models}, 
      author={Jie Zhu and Qian Chen and Huaixia Dou and Junhui Li and Lifan Guo and Feng Chen and Chi Zhang},
      year={2025},
      eprint={2504.15716},
      archivePrefix={arXiv},
      primaryClass={cs.AI},
      url={https://arxiv.org/abs/2504.15716}, 
}

@inproceedings{inproceedings,
  author    = {Engels, Gregor and Heckel, Reiko and Sauer, Stefan},
  title     = {{UML} -- A Universal Modeling Language?},
  booktitle = {Proceedings of the 21st International Conference on Application and Theory of Petri Nets (ICATPN 2000)},
  editor    = {Nielsen, Michael and Simpson, David},
  series    = {Lecture Notes in Computer Science},
  volume    = {1825},
  pages     = {24--38},
  year      = {2000},
  month     = oct,
  publisher = {Springer, Heidelberg},
  isbn      = {978-3-540-67693-5},
  doi       = {10.1007/3-540-44988-4_3},
}

@inproceedings{wan-etal-2023-universal,
    title = "Universal Self-Adaptive Prompting",
    author = "Wan, Xingchen  and
      Sun, Ruoxi  and
      Nakhost, Hootan  and
      Dai, Hanjun  and
      Eisenschlos, Julian  and
      Arik, Sercan  and
      Pfister, Tomas",
    editor = "Bouamor, Houda  and
      Pino, Juan  and
      Bali, Kalika",
    booktitle = "Proceedings of the 2023 Conference on Empirical Methods in Natural Language Processing",
    month = dec,
    year = "2023",
    address = "Singapore",
    publisher = "Association for Computational Linguistics",
    url = "https://aclanthology.org/2023.emnlp-main.461/",
    doi = "10.18653/v1/2023.emnlp-main.461",
    pages = "7437--7462"
}

@misc{mermaid,
  title        = {Mermaid: A JavaScript-based diagramming and charting tool},
  url          = {https://github.com/mermaid-js/mermaid},
  author       = {Knut Sveidqvist and contributors},
  year         = {2025},
  howpublished = {GitHub repository},
}

@inproceedings{xu-etal-2024-pride,
    title = "Pride and Prejudice: {LLM} Amplifies Self-Bias in Self-Refinement",
    author = "Xu, Wenda  and
      Zhu, Guanglei  and
      Zhao, Xuandong  and
      Pan, Liangming  and
      Li, Lei  and
      Wang, William",
    editor = "Ku, Lun-Wei  and
      Martins, Andre  and
      Srikumar, Vivek",
    booktitle = "Proceedings of the 62nd Annual Meeting of the Association for Computational Linguistics (Volume 1: Long Papers)",
    month = aug,
    year = "2024",
    address = "Bangkok, Thailand",
    publisher = "Association for Computational Linguistics",
    url = "https://aclanthology.org/2024.acl-long.826/",
    doi = "10.18653/v1/2024.acl-long.826",
    pages = "15474--15492"
}

@article{Besta_2024,
   title={Graph of Thoughts: Solving Elaborate Problems with Large Language Models},
   volume={38},
   ISSN={2159-5399},
   url={http://dx.doi.org/10.1609/aaai.v38i16.29720},
   DOI={10.1609/aaai.v38i16.29720},
   number={16},
   journal={Proceedings of the AAAI Conference on Artificial Intelligence},
   publisher={Association for the Advancement of Artificial Intelligence (AAAI)},
   author={Besta, Maciej and Blach, Nils and Kubicek, Ales and Gerstenberger, Robert and Podstawski, Michal and Gianinazzi, Lukas and Gajda, Joanna and Lehmann, Tomasz and Niewiadomski, Hubert and Nyczyk, Piotr and Hoefler, Torsten},
   year={2024},
   month=mar, pages={17682–17690} }

@article{Lee_2025,
   title={Large Language Models in Finance (FinLLMs)},
   ISSN={1433-3058},
   url={http://dx.doi.org/10.1007/s00521-024-10495-6},
   DOI={10.1007/s00521-024-10495-6},
   journal={Neural Computing and Applications},
   publisher={Springer Science and Business Media LLC},
   author={Lee, Jean and Stevens, Nicholas and Han, Soyeon Caren},
   year={2025},
   month=jan }

@misc{bhatia2024fintralfamilygpt4level,
      title={FinTral: A Family of GPT-4 Level Multimodal Financial Large Language Models}, 
      author={Gagan Bhatia and El Moatez Billah Nagoudi and Hasan Cavusoglu and Muhammad Abdul-Mageed},
      year={2024},
      eprint={2402.10986},
      archivePrefix={arXiv},
      primaryClass={cs.CL},
      url={https://arxiv.org/abs/2402.10986}, 
}

@inproceedings{harsha-etal-2025-synthetic,
    title = "Synthetic Data Generation Using Large Language Models for Financial Question Answering",
    author = "Harsha, Chetan  and
      Phogat, Karmvir Singh  and
      Dasaratha, Sridhar  and
      Puranam, Sai Akhil  and
      Ramakrishna, Shashishekar",
    editor = "Chen, Chung-Chi  and
      Moreno-Sandoval, Antonio  and
      Huang, Jimin  and
      Xie, Qianqian  and
      Ananiadou, Sophia  and
      Chen, Hsin-Hsi",
    booktitle = "Proceedings of the Joint Workshop of the 9th Financial Technology and Natural Language Processing (FinNLP), the 6th Financial Narrative Processing (FNP), and the 1st Workshop on Large Language Models for Finance and Legal (LLMFinLegal)",
    month = jan,
    year = "2025",
    address = "Abu Dhabi, UAE",
    publisher = "Association for Computational Linguistics",
    url = "https://aclanthology.org/2025.finnlp-1.7/",
    pages = "76--95"
}

@inproceedings{lee-etal-2024-finale,
    title = "{FINALE} : Finance Domain Instruction-Tuning Dataset with High-Quality Rationales via Chain-of-Thought Prompting",
    author = "Lee, Sangmin  and
      Oh, Suzie  and
      Park, Saeran  and
      Son, Guijin  and
      Kang, Pilsung",
    editor = "Chen, Chung-Chi  and
      Ishigaki, Tatsuya  and
      Takamura, Hiroya  and
      Murai, Akihiko  and
      Nishino, Suzuko  and
      Huang, Hen-Hsen  and
      Chen, Hsin-Hsi",
    booktitle = "Proceedings of the Eighth Financial Technology and Natural Language Processing and the 1st Agent AI for Scenario Planning",
    month = "3 " # aug,
    year = "2024",
    address = "Jeju, South Korea",
    publisher = "-",
    url = "https://aclanthology.org/2024.finnlp-2.9/",
    pages = "89--106"
}

@misc{yao2023treethoughtsdeliberateproblem,
      title={Tree of Thoughts: Deliberate Problem Solving with Large Language Models}, 
      author={Shunyu Yao and Dian Yu and Jeffrey Zhao and Izhak Shafran and Thomas L. Griffiths and Yuan Cao and Karthik Narasimhan},
      year={2023},
      eprint={2305.10601},
      archivePrefix={arXiv},
      primaryClass={cs.CL},
      url={https://arxiv.org/abs/2305.10601}, 
}

@inproceedings{DeBari2024UML,
  author    = {Daniele De Bari and Giacomo Garaccione and Riccardo Coppola and Marco Torchiano and Luca Ardito},
  title     = {Evaluating Large Language Models in Exercises of UML Class Diagram Modeling},
  booktitle = {Proceedings of the 18th ACM/IEEE International Symposium on Empirical Software Engineering and Measurement (ESEM '24)},
  pages     = {393--399},
  year      = {2024},
  publisher = {ACM},
  doi       = {10.1145/3674805.3690741},
  url       = {https://doi.org/10.1145/3674805.3690741}
}

@misc{chen2022finqadatasetnumericalreasoning,
      title={FinQA: A Dataset of Numerical Reasoning over Financial Data}, 
      author={Zhiyu Chen and Wenhu Chen and Charese Smiley and Sameena Shah and Iana Borova and Dylan Langdon and Reema Moussa and Matt Beane and Ting-Hao Huang and Bryan Routledge and William Yang Wang},
      year={2022},
      eprint={2109.00122},
      archivePrefix={arXiv},
      primaryClass={cs.CL},
      url={https://arxiv.org/abs/2109.00122}, 
}

@misc{yang2020finbertpretrainedlanguagemodel,
      title={FinBERT: A Pretrained Language Model for Financial Communications}, 
      author={Yi Yang and Mark Christopher Siy UY and Allen Huang},
      year={2020},
      eprint={2006.08097},
      archivePrefix={arXiv},
      primaryClass={cs.CL},
      url={https://arxiv.org/abs/2006.08097}, 
}

@misc{wei2022finetunedlanguagemodelszeroshot,
      title={Finetuned Language Models Are Zero-Shot Learners}, 
      author={Jason Wei and Maarten Bosma and Vincent Y. Zhao and Kelvin Guu and Adams Wei Yu and Brian Lester and Nan Du and Andrew M. Dai and Quoc V. Le},
      year={2022},
      eprint={2109.01652},
      archivePrefix={arXiv},
      primaryClass={cs.CL},
      url={https://arxiv.org/abs/2109.01652}, 
}

@misc{300hours2025,
	title        = {{CFA Level 1 Economics Cheat Sheet}},
	author       = {{300Hours}},
	year         = {2025},
	url          = {https://300hours.com/cfa-level-1-economics-cheat-sheet/?utm_source=chatgpt.com}
}

@misc{300hoursEquity2025,
	title        = {{CFA Level 1 Equity Investments: Our Cheat Sheet}},
	author       = {{300Hours}},
	year         = {2025},
	url          = {https://300hours.com/cfa-level-1-equity-investments-cheat-sheet}
}

@book{efron1994introduction,
  title={An Introduction to the Bootstrap},
  author={Efron, Bradley and Tibshirani, Robert J},
  year={1994},
  publisher={CRC press}
}

@misc{analystprep2025,
	title        = {{Economics CFA Level 1 Essential Review Summary}},
	author       = {{AnalystPrep}},
	year         = {2025},
	url          = {https://analystprep.com/blog/economics-cfa-level-1-essential-review-summary/?utm_source=chatgpt.com}
}

@misc{arya2024llms,
	title        = {{5 Best Large Language Models (LLMs) for Financial Analysis}},
	author       = {Arya.ai},
	year         = {2024},
	url          = {https://arya.ai/blog/5-best-large-language-models-llms-for-financial-analysis}
}

@article{avellaneda2008hft,
	title        = {{High-Frequency Trading in a Limit Order Book}},
	author       = {Avellaneda, Marco and Stoikov, Sasha},
	year         = {2008},
	journal      = {Quantitative Finance},
	volume       = {8},
	number       = {3},
	pages        = {217--224},
	doi          = {10.1080/14697680701381228},
	url          = {https://doi.org/10.1080/14697680701381228}
}

@techreport{baselOTC2025,
	title        = {{OTC Derivatives Market Reforms}},
	author       = {Financial Stability Board},
	year         = {2025},
	url          = {https://www.fsb.org/uploads/P251120.pdf},
	institution  = {Financial Stability Board}
}

@article{black1973pricing,
	title        = {{The Pricing of Options and Corporate Liabilities}},
	author       = {Black, Fischer and Scholes, Myron},
	year         = {1973},
	journal      = {Journal of Political Economy},
	volume       = {81},
	number       = {3},
	pages        = {637--654},
	doi          = {10.1086/260062}
}

@misc{bloombergAggregate,
	title        = {{Bloomberg US Aggregate Bond Index}},
	author       = {Bloomberg},
	year         = {2025},
	url          = {https://www.bloomberg.com/quote/LBUSTRUU:IND}
}

@book{bodie2017investments,
	title        = {{Investments}},
	author       = {Bodie, Zvi and Kane, Alex and Marcus, Alan J.},
	year         = {2017},
	publisher    = {McGraw-Hill Education},
	address      = {New York, NY},
	isbn         = {978-1259297840},
	edition      = {11th}
}

@book{brown2007financialReporting,
	title        = {{Financial Reporting, Financial Statement Analysis, and Valuation: A Strategic Perspective}},
	author       = {Stickney, Clyde P. and Brown, Paul R. and Wahlen, James M.},
	year         = {2007},
	publisher    = {Prentice Hall},
	isbn         = {978-0324302950}
}

@article{brown2020language,
	title        = {{Language Models are Few-Shot Learners}},
	author       = {Brown, Tom B. and Mann, Benjamin and Ryder, Nick and Subbiah, Melanie and Kaplan, Jared and Dhariwal, Prafulla and Neelakantan, Arvind and Shyam, Pranav and Sastry, Girish and Askell, Amanda and others},
	year         = {2020},
	journal      = {arXiv preprint arXiv:2005.14165},
	url          = {https://arxiv.org/abs/2005.14165}
}

@misc{caia2025alternative,
	title        = {{Alternative Investment Management}},
	author       = {{CAIA Association}},
	year         = {2025},
	url          = {https://caia.org/curriculum}
}

@inproceedings{callanan-etal-2024-gpt,
	title        = {{Can {GPT} models be Financial Analysts? An Evaluation of {C}hat{GPT} and {GPT}-4 on mock {CFA} Exams}},
	author       = {Callanan, Ethan  and Mbakwe, Amarachi  and Papadimitriou, Antony  and Pei, Yulong  and Sibue, Mathieu  and Zhu, Xiaodan  and Ma, Zhiqiang  and Liu, Xiaomo  and Shah, Sameena},
	year         = {2024},
	month        = {3 } # {aug},
	booktitle    = {Proceedings of the Eighth Financial Technology and Natural Language Processing and the 1st Agent AI for Scenario Planning},
	publisher    = {-},
	address      = {Jeju, South Korea},
	pages        = {23--32},
	url          = {https://aclanthology.org/2024.finnlp-2.2/},
	editor       = {Chen, Chung-Chi  and Ishigaki, Tatsuya  and Takamura, Hiroya  and Murai, Akihiko  and Nishino, Suzuko  and Huang, Hen-Hsen  and Chen, Hsin-Hsi}
}

@misc{cfaAlternativeInvestments2025,
	title        = {{CFA Program Curriculum: Alternative Investments}},
	author       = {{CFA Institute}},
	year         = {2025},
	url          = {https://www.cfainstitute.org/en/programs/cfa/curriculum}
}

@misc{cfaCorporateIssuers2025,
	title        = {{CFA Program Curriculum: Equity Investments \& Fixed Income}},
	author       = {{CFA Institute}},
	year         = {2025},
	url          = {https://www.cfainstitute.org/en/programs/cfa/curriculum}
}

@misc{cfaDerivatives2025,
	title        = {{CFA Program Curriculum: Level II – Derivatives}},
	author       = {{CFA Institute}},
	year         = {2025},
	url          = {https://www.cfainstitute.org/en/programs/cfa/curriculum}
}

@book{cfaEconomics2025,
	title        = {{CFA Program Curriculum 2025: Level II, Volume 2 – Economics}},
	author       = {{CFA Institute}},
	year         = {2024},
	publisher    = {Wiley},
	url          = {https://books.google.com/books/about/2025_CFA_Program_Curriculum_Level_I_Box.html?id=Zl0rEQAAQBAJ},
	note         = {Includes coverage of classical, neoclassical, and endogenous growth models}
}

@misc{cfaEquityInvestments2025,
	title        = {{CFA Program Curriculum: Equity Investments}},
	author       = {{CFA Institute}},
	year         = {2025},
	publisher    = {CFA Institute},
	address      = {Charlottesville, VA},
	url          = {https://www.cfainstitute.org/en/programs/cfa/curriculum}
}

@misc{cfaFinancialReporting2025,
	title        = {{CFA Program Curriculum: Financial Reporting and Analysis}},
	author       = {{CFA Institute}},
	year         = {2025},
	url          = {https://www.cfainstitute.org/programs/cfa-program/curriculum}
}

@misc{cfaFixedIncome2025,
	title        = {{CFA Program Curriculum: Fixed Income (Levels I \& II)}},
	author       = {{CFA Institute}},
	year         = {2025},
	url          = {https://www.cfainstitute.org/en/programs/cfa/curriculum}
}

@misc{cfaPortfolio2025,
	title        = {{CFA Program Curriculum: Portfolio Management}},
	author       = {{CFA Institute}},
	year         = {2025},
	url          = {https://www.cfainstitute.org/en/programs/cfa/curriculum}
}

@book{cfaQuantMethods2025,
	title        = {{CFA Program Curriculum: Quantitative Methods}},
	author       = {{CFA Institute}},
	year         = {2025},
	publisher    = {CFA Institute},
	address      = {Charlottesville, VA},
	note         = {Levels I \& II; covers TVM, probability, hypothesis testing, regression, portfolio stats}
}

@book{damodaran2012investment,
	title        = {{Investment Valuation: Tools and Techniques for Determining the Value of Any Asset}},
	author       = {Damodaran, Aswath},
	year         = {2012},
	publisher    = {Wiley},
	address      = {Hoboken, NJ},
	edition      = {3rd}
}

@article{dichev2008balance,
	title        = {{On the Balance Sheet-Based Model of Financial Reporting}},
	author       = {Dichev, Ilia D.},
	year         = {2008},
	journal      = {Accounting Horizons},
	volume       = {22},
	number       = {4},
	pages        = {453--470},
	doi          = {10.2308/acch.2008.22.4.453},
	url          = {https://publications.aaahq.org/accounting-horizons/article-abstract/22/4/453/1918/On-the-Balance-Sheet-Based-Model-of-Financial}
}

@misc{du2025optimizingtemperaturelanguagemodels,
	title        = {{Optimizing Temperature for Language Models with Multi-Sample Inference}},
	author       = {Weihua Du and Yiming Yang and Sean Welleck},
	year         = {2025},
	url          = {https://arxiv.org/abs/2502.05234},
	eprint       = {2502.05234},
	archiveprefix = {arXiv},
	primaryclass = {cs.LG}
}

@misc{economicsPhillipsCurve2025,
	title        = {{Phillips Curve: Trade‐Off Between Inflation and Unemployment}},
	author       = {{Investopedia}},
	year         = {2025},
	url          = {https://www.investopedia.com/terms/p/phillipscurve.asp?utm_source=chatgpt.com}
}

@misc{efficientlearning2025,
	title        = {{Program Overview: CFA Economics}},
	author       = {{Efficient Learning}},
	year         = {2025},
	url          = {https://www.efficientlearning.com/cfa/resources/program-overview/economics/?utm_source=chatgpt.com}
}

@book{fabozzi2012bond,
	title        = {{Bond Markets, Analysis, and Strategies}},
	author       = {Fabozzi, Frank J.},
	year         = {2012},
	publisher    = {Pearson/Prentice Hall},
	isbn         = {978-0132743070},
	edition      = {8th}
}

@misc{fixedIncomeRebalance,
	title        = {{When to Rebalance a Bond Portfolio}},
	author       = {Investopedia},
	year         = {2025},
	url          = {https://www.investopedia.com/how-to-rebalance-your-portfolio-7973806}
}

@misc{gordon2020cfa,
	title        = {{CFA Exam Level 1 Economics Lecture}},
	author       = {Brian Gordon},
	year         = {2020},
	url          = {https://www.youtube.com/watch?v=SvqKJnN4Tbo},
	howpublished = {\url{https://www.youtube.com/watch?v=SvqKJnN4Tbo}}
}

@misc{prepnuggets2020equityPreview,
	title        = {{CFA Level I: Equity Investments Preview}},
	author       = {{PrepNuggets}},
	year         = {2020},
	url          = {https://www.youtube.com/watch?v=GpBo4eluk38},
	howpublished = {\url{https://www.youtube.com/watch?v=GpBo4eluk38}}
}

@misc{cfanigeria2024equityRevision,
	title        = {{CFA Level 1 Revision Bootcamp - Equity Investment \& Derivatives}},
	author       = {{CFA Society Nigeria}},
	year         = {2024},
	url          = {https://www.youtube.com/live/t8EsywVnGss},
	howpublished = {\url{https://www.youtube.com/live/t8EsywVnGss}}
}

@book{grinold2000active,
	title        = {{Active Portfolio Management}},
	author       = {Grinold, Richard C. and Kahn, Ronald N.},
	year         = {2000},
	publisher    = {McGraw-Hill},
	isbn         = {978-0070248823}
}

@misc{hu2023codepromptingneuralsymbolic,
	title        = {{Code Prompting: a Neural Symbolic Method for Complex Reasoning in Large Language Models}},
	author       = {Yi Hu and Haotong Yang and Zhouchen Lin and Muhan Zhang},
	year         = {2023},
	url          = {https://arxiv.org/abs/2305.18507},
	eprint       = {2305.18507},
	archiveprefix = {arXiv},
	primaryclass = {cs.CL}
}

@book{hull2017options,
	title        = {{Options, Futures, and Other Derivatives}},
	author       = {Hull, John C.},
	year         = {2017},
	publisher    = {Pearson},
	isbn         = {978-0134472089},
	edition      = {10th}
}

@misc{investopediaABS,
	title        = {{Asset-Backed Security (ABS)}},
	author       = {Investopedia},
	year         = {2025},
	url          = {https://www.investopedia.com/terms/a/asset-backedsecurity.asp}
}

@misc{investopediaADAS,
	title        = {{Aggregate Demand and Supply (AD–AS) Model}},
	author       = {{Investopedia}},
	year         = {2025},
	url          = {https://www.investopedia.com/terms/a/aggregatedemand.asp}
}

@misc{investopediaBalanceOfPayments,
	title        = {{Balance of Payments}},
	author       = {{Investopedia}},
	year         = {2025},
	url          = {https://www.investopedia.com/insights/what-is-the-balance-of-payments/}
}

@misc{investopediaBusinessCycle,
	title        = {{Business Cycle: Definition \& 4 Phases}},
	author       = {{Investopedia}},
	year         = {2025},
	url          = {https://www.investopedia.com/terms/b/businesscycle.asp?utm_source=chatgpt.com}
}

@misc{investopediaConvexity,
	title        = {{Convexity in Bonds: Definition, Meaning, and Examples}},
	author       = {Investopedia},
	year         = {2025},
	url          = {https://www.investopedia.com/terms/c/convexity.asp}
}

@misc{investopediaCoupon,
	title        = {{Coupon Rate Definition}},
	author       = {Investopedia},
	year         = {2025},
	url          = {https://www.investopedia.com/terms/c/couponrate.asp}
}

@misc{investopediaCreditSpread,
	title        = {{Credit Spread Definition}},
	author       = {Investopedia},
	year         = {2025},
	url          = {https://www.investopedia.com/terms/c/creditspread.asp}
}

@misc{investopediaDuration,
	title        = {{Duration Definition and Its Use in Fixed Income Investing}},
	author       = {Investopedia},
	year         = {2025},
	url          = {https://www.investopedia.com/terms/d/duration.asp}
}

@misc{investopediaElasticity,
	title        = {{Price Elasticity of Demand}},
	author       = {{Investopedia}},
	year         = {2025},
	url          = {https://www.investopedia.com/terms/e/elasticity.asp?utm_source=chatgpt.com}
}

@misc{investopediaExchangeRate,
	title        = {{Exchange Rate Definition}},
	author       = {{Investopedia}},
	year         = {2025},
	url          = {https://www.investopedia.com/terms/e/exchangerate.asp?utm_source=chatgpt.com}
}

@misc{investopediaFinancialStatements2025,
	title        = {{Financial Statements: List of Types and How to Read Them}},
	author       = {Investopedia},
	year         = {2025},
	url          = {https://www.investopedia.com/terms/f/financial-statements.asp}
}

@misc{investopediaFixedIncome,
	title        = {{Fixed-Income Security Definition, Types, and Examples}},
	author       = {Investopedia},
	year         = {2025},
	url          = {https://www.investopedia.com/terms/f/fixed-income-security.asp}
}

@misc{investopediaFundamentalAnalysis,
	title        = {{Fundamental Analysis: Principles, Types, and How to Use It}},
	author       = {{Investopedia}},
	year         = {2025},
	url          = {https://www.investopedia.com/fundamental-analysis-4689743}
}

@misc{investopediaGDP,
	title        = {{Gross Domestic Product (GDP) Formula}},
	author       = {{Investopedia}},
	year         = {2025},
	url          = {https://www.investopedia.com/terms/g/gdp.asp?utm_source=chatgpt.com}
}

@misc{investopediaInflation,
	title        = {{Consumer Price Index (CPI)}},
	author       = {Investopedia},
	year         = {2025},
	url          = {https://www.investopedia.com/terms/c/consumerpriceindex.asp}
}

@misc{investopediaInterestRates,
	title        = {{Interest Rate Definition \& Impact}},
	author       = {Investopedia},
	year         = {2025},
	url          = {https://www.investopedia.com/terms/i/interestrate.asp}
}

@misc{investopediaMonetaryPolicy,
	title        = {{Monetary Policy Definition}},
	author       = {{Investopedia}},
	year         = {2025},
	url          = {https://www.investopedia.com/terms/m/monetarypolicy.asp?utm_source=chatgpt.com}
}

@misc{investopediaPortersFiveForces,
	title        = {{Porter’s Five Forces Definition}},
	author       = {{Investopedia}},
	year         = {2025},
	url          = {https://www.investopedia.com/terms/p/porter.asp}
}

@misc{investopediaReadReport,
	title        = {{How to Read a Financial Analysis Report}},
	author       = {{Investopedia}},
	year         = {2025},
	url          = {https://www.investopedia.com/articles/investing/032113/basics-financial-analysis-report.asp}
}

@misc{investopediaRegulation2025,
	title        = {{Government Regulations: Do They Help Businesses?}},
	author       = {{Investopedia}},
	year         = {2025},
	url          = {https://www.investopedia.com/articles/economics/11/government-regulations.asp?utm_source=chatgpt.com}
}

@misc{investopediaRule72,
	title        = {{The Rule of 72: What It Is and How to Use It in Investing}},
	author       = {{Investopedia}},
	year         = {2025},
	url          = {https://www.investopedia.com/terms/r/rule-of-72.asp},
	note         = {Provides quick doubling-time approximation}
}

@misc{investopediaSimpleCompoundInterest,
	title        = {{Simple vs. Compound Interest: Definition and Formulas}},
	author       = {{Investopedia}},
	year         = {2025},
	url          = {https://www.investopedia.com/terms/s/simple-interest.asp},
	note         = {Explains TVM basics and formula examples}
}

@misc{investopediaSupplyDemand,
	title        = {{Law of Supply and Demand}},
	author       = {{Investopedia}},
	year         = {2025},
	url          = {https://www.investopedia.com/terms/l/law-of-supply-demand.asp?utm_source=chatgpt.com}
}

@misc{investopediaYieldCurve,
	title        = {{Yield Curve Definition \& Types}},
	author       = {Investopedia},
	year         = {2025},
	url          = {https://www.investopedia.com/terms/y/yieldcurve.asp}
}

@misc{ipassfinance2025,
	title        = {{CFA Economics Study Tips}},
	author       = {{iPassFinanceExams}},
	year         = {2025},
	url          = {https://ipassfinanceexams.com/cfa-economics-study-tips/?utm_source=chatgpt.com}
}

@article{jarrow1995pricing,
	title        = {{Pricing Derivatives on Financial Securities Subject to Credit Risk}},
	author       = {Jarrow, Robert A. and Turnbull, Stuart M.},
	year         = {1995},
	journal      = {Journal of Finance},
	volume       = {50},
	number       = {1},
	pages        = {53--85},
	doi          = {10.1111/j.1540-6261.1995.tb05167.x}
}

@book{jarrow1996derivative,
	title        = {{Derivative Securities}},
	author       = {Jarrow, Robert A. and Turnbull, Stuart M.},
	year         = {1996},
	publisher    = {South-Western College Publishing},
	isbn         = {978-0538860633}
}

@misc{ke2025demystifyingdomainadaptiveposttrainingfinancial,
	title        = {{Demystifying Domain-adaptive Post-training for Financial LLMs}},
	author       = {Zixuan Ke and Yifei Ming and Xuan-Phi Nguyen and Caiming Xiong and Shafiq Joty},
	year         = {2025},
	url          = {https://arxiv.org/abs/2501.04961},
	eprint       = {2501.04961},
	archiveprefix = {arXiv},
	primaryclass = {cs.CL}
}

@misc{liu2025finr1largelanguagemodel,
	title        = {{Fin-R1: A Large Language Model for Financial Reasoning through Reinforcement Learning}},
	author       = {Zhaowei Liu and Xin Guo and Fangqi Lou and Lingfeng Zeng and Jinyi Niu and Zixuan Wang and Jiajie Xu and Weige Cai and Ziwei Yang and Xueqian Zhao and Chao Li and Sheng Xu and Dezhi Chen and Yun Chen and Zuo Bai and Liwen Zhang},
	year         = {2025},
	url          = {https://arxiv.org/abs/2503.16252},
	eprint       = {2503.16252},
	archiveprefix = {arXiv},
	primaryclass = {cs.CL}
}

@article{markowitz1952portfolio,
	title        = {{Portfolio Selection}},
	author       = {Markowitz, Harry},
	year         = {1952},
	journal      = {The Journal of Finance},
	publisher    = {Wiley},
	address      = {New York},
	series       = {Journal of Finance Classics},
	volume       = {7},
	pages        = {77--91},
	doi          = {10.2307/2975974}
}

@book{metrick2010private,
	title        = {{Private Equity and Venture Capital}},
	author       = {Metrick, Andrew and Yasuda, Ayako},
	year         = {2010},
	publisher    = {Wiley},
	isbn         = {978-0470405398}
}

@article{nie2024survey,
	title        = {{A Survey of Large Language Models for Financial Applications: Progress, Prospects and Challenges}},
	author       = {Nie, Yuqi and Kong, Yaxuan and Dong, Xiaowen and Mulvey, John M. and Poor, H. Vincent and Wen, Qingsong and Zohren, Stefan},
	year         = {2024},
	journal      = {arXiv preprint arXiv:2406.11903}
}

@book{palepuHealy2013business,
	title        = {{Business Analysis and Valuation: Using Financial Statements}},
	author       = {Palepu, Krishna G. and Healy, Paul M. and Wright, Sue and Bradbury, Michael and Coulton, Jeff},
	year         = {2013},
	publisher    = {Cengage},
	isbn         = {978-0324302869}
}

@book{penman2012financial,
	title        = {{Financial Statement Analysis and Security Valuation}},
	author       = {Penman, Stephen H.},
	year         = {2012},
	publisher    = {McGraw-Hill Education},
	isbn         = {978-0073530820},
	edition      = {5th}
}

@book{pinto2015equity,
	title        = {{Equity Asset Valuation}},
	author       = {Pinto, Jerald E. and Henry, Elaine and Robinson, Thomas R. and Stowe, John D.},
	year         = {2015},
	publisher    = {Wiley},
	address      = {Hoboken, NJ},
	series       = {CFA Institute Investment Series},
	isbn         = {978-1119320713}
}

@misc{raymondJamesFI,
	title        = {{Fixed Income Strategies}},
	author       = {Raymond James},
	year         = {2025},
	url          = {https://www.raymondjames.com/wealth-management/advice-products-and-services/investment-solutions/fixed-income/fixed-income-strategies}
}

@inproceedings{Renze_2024,
	title        = {{The Benefits of a Concise Chain of Thought on Problem-Solving in Large Language Models}},
	author       = {Renze, Matthew and Guven, Erhan},
	year         = {2024},
	month        = nov,
	booktitle    = {2024 2nd International Conference on Foundation and Large Language Models (FLLM)},
	publisher    = {IEEE},
	pages        = {476–483},
	doi          = {10.1109/fllm63129.2024.10852493},
	url          = {http://dx.doi.org/10.1109/FLLM63129.2024.10852493}
}

@techreport{rockafellar2000cvar,
	title        = {{Optimization of Conditional Value-at-Risk}},
	author       = {Rockafellar, R. Tyrrell and Uryasev, Stanislav},
	year         = {2000},
	number       = {IFOR Technical Report TR00-27},
	url          = {https://doi.org/10.21314/JOR.2000.038},
	institution  = {University of Florida}
}

@misc{schweser2025,
	title        = {{Level I CFA Economics Study Tips}},
	author       = {{Kaplan Schweser}},
	year         = {2025},
	url          = {https://www.schweser.com/cfa/blog/how-to-pass-the-cfa-exam/cfa-level-1-curriculum-topic-economics-tips}
}

@book{tavella2000pricing,
	title        = {{Pricing Financial Instruments: The Finite Difference Method}},
	author       = {Tavella, Dominique and Randall, Curt},
	year         = {2000},
	publisher    = {Wiley},
	isbn         = {978-0471185385}
}

@book{tuckman2011fixed,
	title        = {{Fixed Income Securities: Tools for Today’s Markets}},
	author       = {Tuckman, Bruce and Serrat, Angel},
	year         = {2011},
	publisher    = {Wiley},
	isbn         = {978-0470481685},
	edition      = {3rd}
}

@misc{uworldEconomics2025,
	title        = {{CFA® Economics: Syllabus \& Sample Questions}},
	author       = {{UWorld Finance}},
	year         = {2025},
	url          = {https://finance.uworld.com/cfa/economics/?utm_source=chatgpt.com}
}

@misc{uworldFinance2025,
	title        = {{CFA® Finance Study Resources}},
	author       = {{UWorld Finance}},
	year         = {2025},
	url          = {https://finance.uworld.com/?utm_source=chatgpt.com}
}

@inproceedings{wang-etal-2023-plan,
	title        = {{Plan-and-Solve Prompting: Improving Zero-Shot Chain-of-Thought Reasoning by Large Language Models}},
	author       = {Wang, Lei  and Xu, Wanyu  and Lan, Yihuai  and Hu, Zhiqiang  and Lan, Yunshi  and Lee, Roy Ka-Wei  and Lim, Ee-Peng},
	year         = {2023},
	month        = jul,
	booktitle    = {Proceedings of the 61st Annual Meeting of the Association for Computational Linguistics (Volume 1: Long Papers)},
	publisher    = {Association for Computational Linguistics},
	address      = {Toronto, Canada},
	pages        = {2609--2634},
	doi          = {10.18653/v1/2023.acl-long.147},
	url          = {https://aclanthology.org/2023.acl-long.147/},
	editor       = {Rogers, Anna  and Boyd-Graber, Jordan  and Okazaki, Naoaki}
}

@misc{wei2023chainofthoughtpromptingelicitsreasoning,
	title        = {{Chain-of-Thought Prompting Elicits Reasoning in Large Language Models}},
	author       = {Jason Wei and Xuezhi Wang and Dale Schuurmans and Maarten Bosma and Brian Ichter and Fei Xia and Ed Chi and Quoc Le and Denny Zhou},
	year         = {2023},
	url          = {https://arxiv.org/abs/2201.11903},
	eprint       = {2201.11903},
	archiveprefix = {arXiv},
	primaryclass = {cs.CL}
}

@misc{wikipediaTimeValueMoney,
	title        = {{Time Value of Money}},
	author       = {{Wikipedia contributors}},
	year         = {2025},
	url          = {https://en.wikipedia.org/wiki/Time_value_of_money},
	note         = {Overview of present and future value concepts}
}

@book{wooldridge2013introductory,
	title        = {{Introductory Econometrics: A Modern Approach}},
	author       = {Wooldridge, Jeffrey M.},
	year         = {2013},
	publisher    = {Cengage Learning},
	address      = {Boston, MA},
	edition      = {5th}
}
\clearpage
\onecolumn
\appendix

\section{Expert Reasoning Blueprints}\label{appendix:blueprint_sources}
\begin{tcolorbox}[colback=lightyellow!80!white,
                  colframe=black,
                  title=Economics,
                  fonttitle=\bfseries,
                  boxrule=0.5pt,
                  arc=2mm,
                  sharp corners=southwest]

\scriptsize
\begin{verbatim}
***Economics:*** 
```mermaid
graph TD;
  A[Step 1: Question Breakdown] -->|Extract key terms| A1{Identify Topic}
  A1 -->|Micro: Supply & Demand, Market Structures| A2
  A1 -->|Macro: GDP, Growth, Policy, Trade| A3
  A1 -->|Currency & Regulation| A4

  A2 --> B1[Identify model: Elasticity, Cost Curves, Shutdown Points]
  A3 --> B2[Map to AD-AS, Business Cycles, Growth Theories]
  A4 --> B3[Assess Exchange Rates, Trade, Capital Flows, Regulation]

  B1 -->|Check for formula or concept?| C{Numerical or Conceptual}
  B2 --> C
  B3 --> C

  C -->|Numerical| D1[Extract data, apply formulas, check assumptions]
  C -->|Conceptual| D2[Analyze cause-effect, policy impact]

  D1 --> E[Step 4: Solution Development]
  D2 --> E
  E -->|Construct structured response| E1(Core insight + economic rationale)
  E -->|Consider alternative scenarios| E2(Assess different possibilities)

  E1 --> F[Step 5: Answer Validation]
  E2 --> F
  F -->|Check logic, principles, and assumptions| F1(Verify consistency)
  F1 -->|Ensure completeness & clarity| F2(Confirm answer structure)
```
\end{verbatim}
\end{tcolorbox}
\paragraph{Explanation:}Step-1: Question Breakdown (A) --  
   Extract key terms by parsing the question to see whether it focuses on microeconomics, macroeconomics, or currency/regulation topics~\citep{300hours2025,uworldEconomics2025}.

Step-2: Identify Topic (A1)  
   – Microeconomics (A2): Focus on supply \& demand mechanisms and market structures such as perfect competition, monopoly, oligopoly, and monopolistic competition~\citep{investopediaSupplyDemand,300hours2025}.  
   – Macroeconomics (A3): Consider aggregate demand–aggregate supply analysis, phases of the business cycle (expansion, peak, contraction, trough), and growth models (Solow, endogenous growth)~\citep{investopediaBusinessCycle,cfaEconomics2025}.  
   – Currency \& Regulation (A4): Examine exchange-rate regimes (floating vs.\ pegged), trade balances, capital-flow impacts, and relevant government policies~\citep{investopediaExchangeRate,investopediaRegulation2025}.

Step-3: Model Selection or Strategy Mapping (B1–B3)
   – Micro Models (B1): Choose elasticity calculations and cost‐curve analysis (marginal/average cost, shutdown point) for supply–demand or firm‐behavior questions~\citep{investopediaElasticity}.  
   – Macro Frameworks (B2): Apply AD–AS curves, Phillips‐curve trade‐offs, or business‐cycle indicators to frame policy or growth analysis~\citep{investopediaADAS,economicsPhillipsCurve2025}.  
   – FX \& Regulation (B3): Use exchange‐rate determination models, balance‐of‐payments analysis, or regulatory impact frameworks for currency/trade questions~\citep{investopediaBalanceOfPayments}.

Step-4: Determine Numerical vs.\ Conceptual Approach (C) 
   – Numerical (D1): Gather the relevant data (prices, quantities, rates), apply formulae (e.g., \(\text{Elasticity} = \frac{\%\Delta Q_n}{\%\Delta P}\), \(\text{GDP} = C + I + G + (X - M)\)
), and verify assumptions~\citep{investopediaGDP}.  
   – Conceptual (D2): Construct a narrative explaining cause–effect relationships (e.g., how a monetary‐policy change shifts AD or how trade barriers affect capital flows)~\citep{investopediaMonetaryPolicy}.

Step-5: Solution Development (E) 
   – Structured Response (E1): State the core economic insight first, then provide the step‐by‐step rationale linking theory to the question context~\citep{schweser2025}.  
   – Alternative Scenarios (E2): Where relevant, outline best‐case, base‐case, and worst‐case scenarios or show how supply–demand curves shift under different assumptions~\citep{ipassfinance2025}.

Step-6: Answer Validation (F)
   – Verify Consistency (F1): Check that numerical answers satisfy boundary conditions (e.g., correct sign on elasticity, GDP component sums).  
   – Confirm Clarity (F2): Ensure your explanation is complete, logically ordered, and clearly communicates both the result and its limitations~\citep{uworldFinance2025}.

\paragraph{Source:}300Hours CFA Level 1 Economics Cheat Sheet~\citep{300hours2025}, UWorld Finance's CFA® Economics: Syllabus \& Sample Questions~\citep{uworldEconomics2025}, Kaplan Schweser's Level I Economics tips~\citep{schweser2025}, Efficient Learning's CFA Economics overview~\citep{efficientlearning2025}, iPass Finance Exams' study guide~\citep{ipassfinance2025}, AnalystPrep's essential review summary~\citep{analystprep2025}, and key Investopedia articles on supply and demand~\citep{investopediaSupplyDemand}, GDP~\citep{investopediaGDP}, and business cycles~\citep{investopediaBusinessCycle} as well as Prof. Brian Gordon's CFA Exam Level 1 Economics video~\citep{gordon2020cfa}.

\begin{tcolorbox}[colback=lightyellow!80!white,
                  colframe=black,
                  title=Fixed Income,
                  fonttitle=\bfseries,
                  boxrule=0.5pt,
                  arc=2mm,
                  sharp corners=southwest]

\scriptsize
\scriptsize
\begin{verbatim}
***Fixed Income:***
```mermaid
graph TD
    A[Purpose and Scope] --> B3[Analyze Macro Conditions]
    B --> C[Assess Bond Features]
    C --> D[Risk and Yield Analysis]
    D --> E[Develop Recommendations]
    E --> F[Review Performance]

    %% Notes and detailed steps
    A --> |Set objectives| B
    B --> |Review interest rates and inflation| C
    C --> |Focus on duration, spread| D
    D --> |Assess scenarios| E
``` 
\end{verbatim}
\end{tcolorbox}
\paragraph{Explanation:}Step-1: Purpose and Scope -- Define the investment objective--income generation, capital preservation, hedging, or total return and establish portfolio constraints and benchmarks such as target yield, duration limits, credit quality floors, or sector allocation guidelines~\citep{investopediaFixedIncome}.

Step-2: Analyze Macro Conditions -- Examine current and forecast interest rate paths, since rising rates erode bond prices and falling rates support them~\citep{investopediaInterestRates}; monitor inflation indicators (CPI, PPI) to gauge real yield trends~\citep{investopediaInflation}; and assess yield-curve shapes (normal, inverted, flat) for economic turning points and yield-curve trade opportunities~\citep{investopediaYieldCurve}.

Step-3: Assess Bond Features -- Identify bond type government, corporate, municipal, structured products (ABS/MBS) and note any embedded options (callable, putable, convertible)~\citep{investopediaABS}; review coupon structure (fixed vs.\ floating), payment frequency, and maturity to understand cash-flow timing and reinvestment risk~\citep{investopediaCoupon}.

Step-4. Risk and Yield Analysis -- Calculate duration to estimate price sensitivity to yield changes~\citep{investopediaDuration} and convexity for non-linear price effects~\citep{investopediaConvexity}; analyze credit spreads over benchmarks to gauge default and liquidity risk~\citep{investopediaCreditSpread}; and stress-test the portfolio under parallel shifts, steepeners, and flatteners to assess P\&L impacts.

Step-5: Develop Recommendations -- Formulate strategies such as adjusting overall duration (shorten if rates are likely to rise), implementing barbell or laddered maturity structures, or choosing bullet portfolios to manage reinvestment and rate risk~\citep{raymondJamesFI}.

Step-6: Review Performance -- Track total returns (price changes plus coupon income) against benchmarks like the Bloomberg US Aggregate Bond Index~\citep{bloombergAggregate}; perform attribution analysis to decompose yield carry, curve roll-down, and spread effects; and revisit assumptions and rebalance when market conditions or issuer fundamentals change~\citep{fixedIncomeRebalance}.

\paragraph{Source:}CFA Program Curriculum for Fixed Income~\citep{cfaFixedIncome2025}, Fabozzi's Bond Markets, Analysis, and Strategies~\citep{fabozzi2012bond}, Tuckman \& Serrat's Fixed Income Securities~\citep{tuckman2011fixed}, Jarrow \& Turnbull's credit‐risk derivatives pricing~\citep{jarrow1995pricing}, Basel Committee papers on credit risk, Investopedia articles on fixed‐income concepts~\citep{investopediaFixedIncome,investopediaConvexity,investopediaDuration}, Reuters coverage of convexity risk, and Dichev's balance sheet model for distress prediction~\citep{dichev2008balance}.

\begin{tcolorbox}[colback=lightyellow!80!white,
                  colframe=black,
                  title=Quantitative Methods,
                  fonttitle=\bfseries,
                  boxrule=0.5pt,
                  arc=2mm,
                  sharp corners=southwest]
\scriptsize
\begin{verbatim}
***Quantitative Methods:*** 
```mermaid
graph TD
    A["Articulating Purpose and Context"] --> B["Collecting Input Data"]
    B --> C["Processing and Cleaning Data"]
    C --> D["Selecting Quantitative Models and Tools"]
    D --> E["Estimating Parameters and Testing Hypotheses"]
    E --> F["Interpreting Results and Communicating Findings"]
    F --> G["Monitoring and Model Reassessment"]
```
\end{verbatim}
\end{tcolorbox}
\paragraph{Explanation:}Step-1: Articulating Purpose and Context (A) Define the research question time value of money calculations, probability distributions, hypothesis testing, regression analysis, or portfolio statistics--and establish the CFA application context (market efficiency, risk estimation, cash‐flow forecasting)~\citep{cfaQuantMethods2025}.

Step-2: Collecting Input Data (B) 
   Gather historical returns, economic indicators, financial statements, and market data from reputable sources; ensure relevance by matching data to the chosen objective (e.g., interest rates for TVM, volatility for risk models)~\citep{investopediaSimpleCompoundInterest}.

Step-3: Processing and Cleaning Data (C)
   Perform data quality checks and remove outliers, handle missing values, confirm consistency--and apply transformations (normalization, log-transforms) before analysis~\citep{wooldridge2013introductory}.

Step-4:  Selecting Quantitative Models and Tools (D) 
   Choose appropriate models--ARIMA for time series, linear/multivariate regression, probability distributions, or Monte Carlo simulation--and leverage CFA-recommended software or spreadsheet tools~\citep{damodaran2012investment, investopediaRule72}.

Step-5: Estimating Parameters and Testing Hypotheses (E) 
   Estimate model parameters via regression or maximum likelihood; conduct t-tests, F-tests, or chi-square tests to validate assumptions and results, with Level II emphasis on multivariate regression and sensitivity analysis~\citep{wooldridge2013introductory}.

Step-6: Interpreting Results and Communicating Findings (F)
   Translate coefficients, p-values, and confidence intervals into actionable investment insights; prepare clear visual aids (charts, tables) to support recommendations~\citep{bodie2017investments}.

Step-7: Monitoring and Model Reassessment (G) 
   Track out-of-sample performance against benchmarks; update models as new data arrive, reassess assumptions, and recalibrate parameters to maintain relevance~\citep{wikipediaTimeValueMoney}.

\paragraph{Source:}CFA Program Curriculum: Quantitative Methods~\citep{cfaQuantMethods2025}, Wooldridge's Introductory Econometrics~\citep{wooldridge2013introductory}, Damodaran's Investment Valuation~\citep{damodaran2012investment}, and Bodie, Kane, \& Marcus's Investments~\citep{bodie2017investments}.

\begin{tcolorbox}[colback=lightyellow!80!white,
                  colframe=black,
                  title=Equity Investments,
                  fonttitle=\bfseries,
                  boxrule=0.5pt,
                  arc=2mm,
                  sharp corners=southwest]
\scriptsize
\begin{verbatim}
***Equity Investing:*** 
```mermaid
graph TD
    A[Objective Setting] --> B[Market and Sector Insights]
    B --> C[Industry Competitive Analysis]
    C --> D[Company Review]
    D --> E[Valuation and Risks]
    E --> F[Investment Decision]

    %% Step-specific highlights
    B --> |Look at growth patterns| C
    C --> |Evaluate competitors' positions| D
    D --> |Check financial health| E
    E --> |Combine insights into strategy| F
```
\end{verbatim}
\end{tcolorbox}
\paragraph{Explanation}

Step-1: Objective Setting (A)  
   Define your investment objectives--capital appreciation, dividend income, or total return in line with your risk tolerance and investment horizon; consider constraints such as liquidity needs, tax implications, regulatory requirements, and any specific mandates~\citep{cfaEquityInvestments2025}.

Step-2: Market and Sector Insights (B)  
   Assess macro indicators (GDP growth, interest rates, inflation) to gauge the overall market environment and identify sectors poised for growth or decline based on economic trends, technological shifts, and consumer behavior~\citep{investopediaFundamentalAnalysis}.

Step-3: Industry Competitive Analysis (C)  
   Apply Porter's Five Forces to evaluate industry attractiveness--competitive rivalry, threat of new entrants, bargaining power of suppliers and buyers, and substitute threats--and assess each firm's market share and competitive moat~\citep{investopediaPortersFiveForces}.

Step-4: Company Review (D)  
   Examine financial statements (income statement, balance sheet, cash flows) to measure profitability, liquidity, and stability; evaluate management's track record and strategic vision; and review corporate governance structures to ensure alignment with shareholder interests~\citep{investopediaReadReport,bodie2017investments}.

Step-5: Valuation and Risks (E)  
   Use valuation methods--Discounted Cash Flow (DCF), Price-to-Earnings (P/E) ratios, Dividend Discount Models (DDM)--to estimate intrinsic value; identify key risks such as market volatility, operational challenges, regulatory changes, and competitive threats~\citep{pinto2015equity}.

Step-6: Investment Decision (F)  
   Formulate your Buy, Hold, or Sell recommendation based on the above analyses and determine how the position fits within the broader portfolio--considering diversification, correlation, and overall risk–return objectives~\citep{300hoursEquity2025}.

\paragraph{Source:}CFA Program Curriculum's Equity Investments module~\citep{cfaEquityInvestments2025}, Investopedia's guides on fundamental analysis~\citep{investopediaFundamentalAnalysis}, Porter's Five Forces stock analysis~\citep{investopediaPortersFiveForces}, and reading financial reports~\citep{investopediaReadReport}, 300Hours' CFA Level 1 Equity Cheat Sheet~\citep{300hoursEquity2025}, Bodie, Kane \& Marcus's Investments~\citep{bodie2017investments}, Pinto et al.'s Equity Asset Valuation~\citep{pinto2015equity}, and CFA Level I Equity video lectures by Prof. Brian Gordon~\citep{prepnuggets2020equityPreview,cfanigeria2024equityRevision}.

\begin{tcolorbox}[colback=lightyellow!80!white,
                  colframe=black,
                  title=Portfolio Management,
                  fonttitle=\bfseries,
                  boxrule=0.5pt,
                  arc=2mm,
                  sharp corners=southwest]
\scriptsize
\begin{verbatim}
***Portfolio Management:*** 
```mermaid
graph TD
    A["Define Investment Objectives"] --> B["Establish Investment Constraints"]
    B --> C["Develop Strategic Asset Allocation"]
    C --> D["Incorporate Tactical Adjustments"]
    D --> E["Select and Optimize Securities"]
    E --> F["Execute Implementation and Trading"]
    F --> G["Measure Performance and Attribution"]
    G --> H["Monitor Risk and Compliance"]
    H --> I["Rebalance and Adjust Portfolio"]
```
\end{verbatim}
\end{tcolorbox}
\paragraph{Explanation:}

Step-1: Define Investment Objectives -- 
Clarify whether the portfolio is aimed at capital growth, income generation, or a balanced mix. Specify expected returns, risk tolerance, and liquidity needs. This step forms the foundation for aligning investment strategy with client mandates~\citep{cfaPortfolio2025}.

Step-2: Establish Investment Constraints --
Define legal, regulatory, tax, and unique client considerations such as ESG preferences or geographic limits. These constraints ensure feasibility and compliance of portfolio design~\citep{cfaPortfolio2025}.

Step-3: {Develop Strategic Asset Allocation --
Allocate across major asset classes (equities, fixed income, alternatives, cash) based on expected returns and risk tolerance. Use models from Modern Portfolio Theory and CAPM to inform allocation~\citep{markowitz1952portfolio, bodie2017investments}.

Step-4: Incorporate Tactical Adjustments --
Introduce short-term adjustments to the strategic allocation based on market outlook or economic indicators. These shifts aim to enhance returns through asset or sector rotation~\citep{grinold2000active}.

Step-5: Select and Optimize Securities --
Apply quantitative screens and qualitative research to choose securities. Use optimization techniques such as mean-variance optimization or the Black-Litterman model to maximize risk-adjusted returns~\citep{bodie2017investments, grinold2000active}.

Step-6: Execute Implementation and Trading --
Implement trade strategies that minimize costs and slippage, considering market impact and liquidity. Align execution with strategic intentions~\citep{cfaPortfolio2025}.

Step-7: Measure Performance and Attribution --
Track performance using return metrics, Sharpe ratio, alpha, and beta. Perform attribution to evaluate decisions across asset allocation, sector, and security selection~\citep{grinold2000active}.

Step-8: Monitor Risk and Compliance -- 
Use tools like Value-at-Risk (VaR), stress testing, and tracking error to monitor portfolio risk. Ensure compliance with constraints and regulations~\citep{cfaPortfolio2025}.

Step-9: Rebalance and Adjust Portfolio -- 
Periodically adjust the portfolio to maintain alignment with the strategic asset allocation as market conditions evolve.
\paragraph{Source:}CFA Program Curriculum's Portfolio Management module~\citep{cfaPortfolio2025}, Bodie, Kane \& Marcus's Investments for portfolio theory and risk-return optimization~\citep{bodie2017investments}, Grinold \& Kahn's Active Portfolio Management for advanced attribution and optimization techniques~\citep{grinold2000active}, and Markowitz's seminal Portfolio Selection on diversification and risk-adjusted returns~\citep{markowitz1952portfolio}.

\begin{tcolorbox}[colback=lightyellow!80!white,
                  colframe=black,
                  title=Derivatives,
                  fonttitle=\bfseries,
                  boxrule=0.5pt,
                  arc=2mm,
                  sharp corners=southwest]
\scriptsize
\begin{verbatim}
***Derivatives:*** 
```mermaid
graph TD
    A[Define Objective and Context] --> B[Identify Derivative Instrument]
    B --> C[Understand Contract Specifications]
    C --> D[Gather Market Data]
    D --> E[Apply Valuation Models]
    E --> F[Assess Risks: Market, Counterparty, etc.]
    F --> G[Construct Payoff Diagrams or Strategies]
    G --> H[Interpret Results and Make Recommendations]
    H --> I[Review, Monitor, and Adjust Strategies]

    %% Example labels or notes (optional)
    A --> |Hedging, speculation, arbitrage| B
    C --> |Features like notional amount, expiration| D
    D --> |Market prices, volatility, risk-free rates| E
    F --> |Sensitivity to Greeks: Delta, Gamma, Vega, etc.| G
    H --> |Adjust based on changing market conditions| I
````
\end{verbatim}
\end{tcolorbox}
\paragraph{Explanation:}

Step-1: Define Objective and Context --
Clarify the purpose of using derivatives: hedging, speculation, or arbitrage. Identify relevant constraints, such as regulatory limitations or portfolio mandates~\citep{cfaDerivatives2025,hull2017options}.

Step-2: Identify Derivative Instrument --
Choose the appropriate derivative: options, futures, forwards, swaps, or structured/exotic products~\citep{jarrow1996derivative}.

Step-3: Understand Contract Specifications --
Review contract parameters, including the underlying asset, strike price, expiration, settlement method (physical or cash), and style (European, American)~\citep{cfaDerivatives2025}.

Step-4: Gather Market Data --
Collect input variables such as spot price, volatility, risk-free rate, dividends, and term structure of interest rates~\citep{hull2017options}.

Step-5: Apply Valuation Models --
Apply pricing frameworks suited to the derivative:
\begin{itemize}
    \item Black-Scholes model for European options~\citep{black1973pricing}.
    \item Binomial Tree for path-dependent or American-style options~\citep{hull2017options}.
    \item Cost-of-carry model for futures and forwards~\citep{jarrow1996derivative}.
    \item Finite-difference methods for complex derivatives~\citep{tavella2000pricing}.
\end{itemize}

Step-6: Assess Risks --
Use Greeks (Delta, Gamma, Vega, Theta, Rho) to evaluate sensitivity to market factors. Consider counterparty and credit risk in OTC markets~\citep{hull2017options, baselOTC2025}.

Step-7: Construct Payoff Diagrams or Strategies --
Visualize outcomes using payoff graphs. Design strategies such as straddles, collars, or protective puts based on desired exposure~\citep{hull2017options}.

Step-8: Interpret Results and Make Recommendations --
Translate model output into actionable insights: confirm hedge effectiveness, profit potential, or risk exposure.

Step-9: Review, Monitor, and Adjust Strategies --
Continuously monitor derivative positions in light of market conditions, risk metrics, and investment objectives~\citep{baselOTC2025}.
\paragraph{Source:} Based on Hull's comprehensive treatment of markets and pricing models~\citep{hull2017options}, the CFA Institute Level II Derivatives readings~\citep{cfaDerivatives2025}, Black \& Scholes's seminal option pricing model~\citep{black1973pricing}, Jarrow \& Turnbull's practical engineering perspective~\citep{jarrow1996derivative}, Tavella \& Randall's numerical finite‐difference techniques~\citep{tavella2000pricing}, and the Basel Committee's OTC derivatives reforms for regulatory context~\citep{baselOTC2025}.

\begin{tcolorbox}[colback=lightyellow!80!white,
                  colframe=black,
                  title=Financial Reporting,
                  fonttitle=\bfseries,
                  boxrule=0.5pt,
                  arc=2mm,
                  sharp corners=southwest]
\scriptsize
\begin{verbatim}
***Financial Reporting:**
```mermaid
graph TD
A[Articulating Purpose and Context] --> B[Collecting Input Data]
    B --> C[Processing Data]
    C --> D[Analyzing and Interpreting Processed Data]
    D --> E[Developing and Communicating Conclusions]
    E --> F[Doing Follow-Up]

    A --> |Defines goals, tools, and audience| B
    B --> |Gather data on economy and industry| C
    C --> |Use tools like ratios and charts| D
    D --> |Interpret data for conclusions| E
    F --> |Periodic review and iteration| A
```
\end{verbatim}
\end{tcolorbox}
\paragraph{Explanation:}

Step-1: Articulating Purpose and Context \\
Define the objectives of the analysis--such as assessing profitability, liquidity, or solvency. Identify stakeholders (e.g., investors, creditors, management) and tailor the analysis to their needs. Set the framework, including accounting standards (IFRS or US GAAP) and the time horizon~\citep{cfaFinancialReporting2025}.

Step-2: Collecting Input Data \\
Gather primary financial statements: income statement, balance sheet, and cash flow statement. Supplement this with industry benchmarks and macroeconomic data. Ensure the quality, accuracy, and completeness of all collected data~\citep{investopediaFinancialStatements2025}.

Step-3: Processing Data \\
Standardize data for comparability by adjusting for non-recurring items or differences in accounting policies. Compute financial ratios such as ROE, current ratio, and debt-to-equity. Use visualizations (e.g., charts, graphs) to uncover trends and patterns~\citep{brown2007financialReporting}.

Step-4: Analyzing and Interpreting Processed Data \\
Assess financial health by interpreting computed ratios. Benchmark against peer companies and industry averages. Identify strengths and weaknesses to determine strategic implications~\citep{palepuHealy2013business}.

Step-5: Developing and Communicating Conclusions \\
Summarize findings in a clear, concise report. Offer actionable recommendations--e.g., restructuring debt or improving efficiency. Tailor communication style and depth to fit the audience, whether board members, analysts, or external investors.

Step-6: Doing Follow-Up \\
Monitor outcomes of implemented actions and assess whether financial targets are met. Update the analysis regularly with new data and refine recommendations. Incorporate feedback to improve future analysis cycles.
\paragraph{Source:}CFA Program Curriculum's Financial Reporting and Analysis readings covering ratio analysis, cash flow analysis, and IFRS/GAAP standards~\citep{cfaFinancialReporting2025} alongside Investopedia's overview of financial statement components~\citep{investopediaFinancialStatements2025}, Paul R. Brown's strategic perspective on statement analysis and valuation~\citep{brown2007financialReporting}, and Palepu \& Healy's MBA‐level treatment of business analysis and valuation using financial statements~\citep{palepuHealy2013business}. 

\begin{tcolorbox}[colback=lightyellow!80!white,
                  colframe=black,
                  title=Alternative Investments,
                  fonttitle=\bfseries,
                  boxrule=0.5pt,
                  arc=2mm,
                  sharp corners=southwest]
\scriptsize
\begin{verbatim}
***Alternative Investments:*** 
```mermaid
graph TD
    A["Define Investment Objectives and Mandate"] --> B["Identify Alternative Asset Classes"]
    B --> C["Conduct Manager and Strategy Due Diligence"]
    C --> D["Perform Valuation and Pricing Analysis"]
    D --> E["Assess Risk and Liquidity"]
    E --> F["Allocate Alternatives in Portfolio"]
    F --> G["Monitor Performance and Rebalance"]
```
\end{verbatim}
\end{tcolorbox}
\paragraph{Explanation:}

Step-1: Define Investment Objectives and Mandate --
Clarify the purpose of including alternative investments--whether for diversification, higher return potential, or hedging against market volatility. Define constraints such as time horizon, liquidity needs, regulatory frameworks, and risk tolerance~\citep{cfaAlternativeInvestments2025}.

Step-2: Identify Alternative Asset Classes --
Explore the universe of alternatives, including hedge funds, private equity, real estate, infrastructure, commodities, and venture capital. Assess how each class contributes to portfolio diversification via low correlation to traditional assets~\citep{bodie2017investments, caia2025alternative}.

Step-3: Conduct Manager and Strategy Due Diligence --
Evaluate managers based on their track record, investment philosophy, risk management, and operational quality. Understand the specific strategies (e.g., long/short, event-driven, global macro) and their alignment with investment mandates~\citep{caia2025alternative, metrick2010private}.

Step-4: Perform Valuation and Pricing Analysis --
Address the unique valuation challenges of illiquid assets. Use models like discounted cash flow (DCF) or mark-to-model, and apply appropriate liquidity or opacity discounts. Compare performance with custom or market benchmarks~\citep{metrick2010private}.

Step-5: Assess Risk and Liquidity --
Identify key risks including market, manager, and operational risks. Analyze downside risk and tail event exposure. Evaluate liquidity risks, such as lock-up periods and redemption windows, that may affect rebalancing ability~\citep{cfaAlternativeInvestments2025}.

Step-6: Allocate Alternatives in Portfolio --
Determine appropriate weighting of alternative assets, guided by expected return, volatility, and correlation with traditional investments. Make strategic allocation decisions with room for tactical adjustments based on market conditions~\citep{bodie2017investments}.

Step-7: Monitor Performance and Rebalance --
Track returns over time, evaluate them against relevant benchmarks, and assess if performance remains consistent with expectations. Rebalance periodically to ensure alignment with objectives, risk profile, and current market landscape~\citep{caia2025alternative}.

\paragraph{Source:}CFA Program Curriculum's Alternative Investments readings covering hedge funds, private equity, real assets, and due diligence frameworks~\citep{cfaAlternativeInvestments2025}--together with Metrick \& Yasuda's deep dive into private equity and venture capital~\citep{metrick2010private}, CAIA Association's comprehensive CAIA-level materials on hedge funds, real estate, commodities, and other alternatives~\citep{caia2025alternative}, and Bodie, Kane \& Marcus's chapters on alternative asset classes and portfolio integration in \emph{Investments}~\citep{bodie2017investments}.

\begin{tcolorbox}[colback=lightyellow!80!white,
                  colframe=black,
                  title=Corporate Issuers,
                  fonttitle=\bfseries,
                  boxrule=0.5pt,
                  arc=2mm,
                  sharp corners=southwest]
\scriptsize
\begin{verbatim}
***Corporate Issuer Analysis:*** 
```mermaid
graph TD
    A["Corporate Issuer Overview"] --> B["Industry Classification"]
    B --> C["Sector Trends and Competitive Landscape"]
    A --> D["Financial Statement Analysis"]
    D --> E["Profitability, Liquidity, Leverage"]
    A --> F["Credit Risk Assessment"]
    F --> G["Rating Agencies and Default Probabilities"]
    A --> H["Capital Structure and Issuance History"]
    H --> I["Bond Issuances and Debt Maturities"]
    A --> J["Corporate Governance and Management"]
    J --> K["Board Quality and Managerial Competence"]
    A --> L["Valuation and Investment Analysis"]
    L --> M["DCF, Relative Valuation, Multiples"]
```
\end{verbatim}
\end{tcolorbox}
\paragraph{Explanation:}

Step-1: Corporate Issuer Overview --
Begin with a high-level understanding of the firm's business model, market positioning, and strategic objectives. This foundational context is essential for both equity and fixed income analysis~\citep{cfaCorporateIssuers2025}.

Step-2: Industry Classification and Sector Trends --
Classify the firm by sector or sub-sector (e.g., financials, consumer discretionary) and evaluate the competitive landscape. Analyze market trends, industry growth prospects, and systemic risks. This industry context shapes performance expectations and relative valuation~\citep{penman2012financial}.

Step-3: Financial Statement Analysis and Key Metrics --
Analyze income statement, balance sheet, and cash flow data. Focus on metrics like revenue growth, operating margin, return on equity, and leverage. This step reveals the firm's financial health and operational efficiency~\citep{penman2012financial, cfaCorporateIssuers2025}.

Step-4: Credit Risk Assessment and Rating Measures --
Evaluate creditworthiness through agency ratings (e.g., S\&P, Moody's), credit spreads, and financial ratios. Analyze the probability of default and credit cycle indicators. This step is vital for bondholders and fixed income portfolio managers~\citep{fabozzi2012bond}.

Step-5: Capital Structure, Issuance History, and Debt Profile --
Examine the firm's financing structure, including the mix of debt vs. equity, historical issuance patterns, and maturity schedules. This informs views on solvency and refinancing risks~\citep{fabozzi2012bond}.

Step-6: Corporate Governance and Leadership Quality --
Assess governance practices such as board independence, shareholder rights, and disclosure quality. Evaluate the management team's execution track record and alignment with shareholder interests~\citep{cfaCorporateIssuers2025}.

Step-7: Valuation and Investment Analysis --
Use valuation models like DCF, P/E, or EV/EBITDA to derive intrinsic value. Develop an investment thesis based on fundamental insights. These valuation techniques are central to both equity and credit investing~\citep{penman2012financial}.
\paragraph{Source:}CFA Program Curriculum's Equity Investments and Fixed Income readings--which cover firm analysis, industry evaluation, and credit assessment frameworks~\citep{cfaCorporateIssuers2025}--along with Penman's Financial Statement Analysis and Security Valuation for accounting-to-valuation linkages~\citep{penman2012financial}, and Fabozzi's Bond Markets, Analysis, and Strategies for credit risk and corporate debt issuance insights~\citep{fabozzi2012bond}.

\clearpage
\section{Prompt Template}
\subsection{Structured Chain-of-Thought (ST-CoT)}
\begin{tcolorbox}[colback=lightyellow!80!white,
                  colframe=black,
                  title=ST-CoT for CFA Exam,
                  fonttitle=\bfseries,
                  boxrule=0.5pt,
                  arc=2mm,
                  sharp corners=southwest]
\scriptsize
\begin{verbatim}
You are a CFA (chartered financial analyst) taking a test to evaluate your knowledge of finance. You think step-by-step approach 
to answer queries.  

Follow these steps:
1. Think through the problem step by step within the <thinking> tags.
2. Provide your final, concise answer within the <output> tags.

The <thinking> sections are for your internal reasoning process only. 
Do not include any part of the final answer in these sections.
The actual response to the query must be entirely contained within the <output> tags.

### Response Format:
<thinking>
[Reasoning through options A, B, and C to understand and solve the problem.]
</thinking>
<output>
"answer": [Final your answer (A , B , or C )]
</output>
\end{verbatim}
\end{tcolorbox}
\subsection{FinCoT}\label{appendix:fincot_prompt_template}
\begin{tcolorbox}[colback=lightyellow!80!white,
                  colframe=black,
                  title=FinCoT for CFA Exam,
                  fonttitle=\bfseries,
                  boxrule=0.5pt,
                  arc=2mm,
                  sharp corners=southwest]
\scriptsize
\begin{verbatim}
You are taking a test for the Chartered Financial Analyst (CFA) program designed to evaluate your knowledge of different topics in 
finance. You think step-by-step approach with reflection to answer queries. 

Follow these steps:
1. Think through the problem step by step reflect and verify while reasoning within the <thinking> tags.
2. Please and put the answer your final, concise answer within the <output> tags.

The <thinking> sections are for your internal reasoning process only. 
Do not include any part of the final answer in these sections.
The actual response to the query must be entirely contained within the <output> tags.

Hint:{THOUGHT.get("embedding_expert_blueprints_[i]")} 

### Response Format:
<thinking>
[Think step by step and respond with your thinking and the correct answer (A, B, or C ), considering the specific sector.]
</thinking>

<output>
"sector": [The sector being addressed],
"question": [The financial question],
"answer": [Reflect and verify the final answer (A, B, or C)]
</output>
\end{verbatim}
\end{tcolorbox}
\subsection{Classify Domain}\label{appendix:classify_domain_prompt}
\begin{tcolorbox}[colback=lightyellow!80!white,
                  colframe=black,
                  title=Classify Domain CFA Exam,
                  fonttitle=\bfseries,
                  boxrule=0.5pt,
                  arc=2mm,
                  sharp corners=southwest]

\begin{Verbatim}[fontsize=\small]
SYSTEM_INSTRUCTION = """You are a CFA expert. Categorize the given CFA question into exactly one 
of these categories:

    Ethical and Professional Standards
    - Code of Ethics, Standards of Professional Conduct, professional integrity
    - Professional responsibilities, ethical decision-making, client interests
    Category code: Ethics

    Quantitative Methods
    - Statistical analysis, probability theory, hypothesis testing
    - Time value of money, financial mathematics, regression analysis
    Category code: Quant.Meth.

    Economic Analysis and Market Forces
    - Microeconomics: supply, demand, market structures
    - Macroeconomics: GDP, inflation, monetary policy, economic cycles
    Category code: Economics

    Financial Reporting and Analysis
    - Financial statements, accounting standards, ratio analysis
    - Balance sheets, income statements, cash flow analysis
    Category code: Fin.Reporting

    Corporate Finance and Issuers
    - Capital structure, dividend policy, corporate governance
    - Mergers & acquisitions, capital budgeting, risk management
    Category code: Corp.Issuers

    Equity Investments
    - Stock valuation, equity markets, company analysis
    - Market efficiency, equity portfolio management
    Category code: EquityInvest.

    Fixed Income Investments
    - Bond markets, yield curves, duration analysis
    - Credit analysis, fixed income portfolio management
    Category code: FixedIncome

    Derivative Instruments
    - Options, futures, forwards, swaps
    - Hedging strategies, derivative pricing, risk management
    Category code: Derivatives

    Alternative Investments
    - Real estate, private equity, hedge funds
    - Commodities, structured products, crypto assets
    Category code: Alter.Invest.

    Portfolio Management
    - Asset allocation, portfolio construction, rebalancing
    - Risk management, performance measurement, client objectives
    Category code: Port.Manage.

Respond with only the single most appropriate category code, nothing else. For example: Ethics, 
Port.Manage., etc.
"""
\end{Verbatim}
\end{tcolorbox}

\clearpage
\section{Domain Distribution}\label{appendix:fig_domain_distribution}
\begin{figure}[!ht]
    \centerline{
    \includegraphics[width=\linewidth,keepaspectratio]{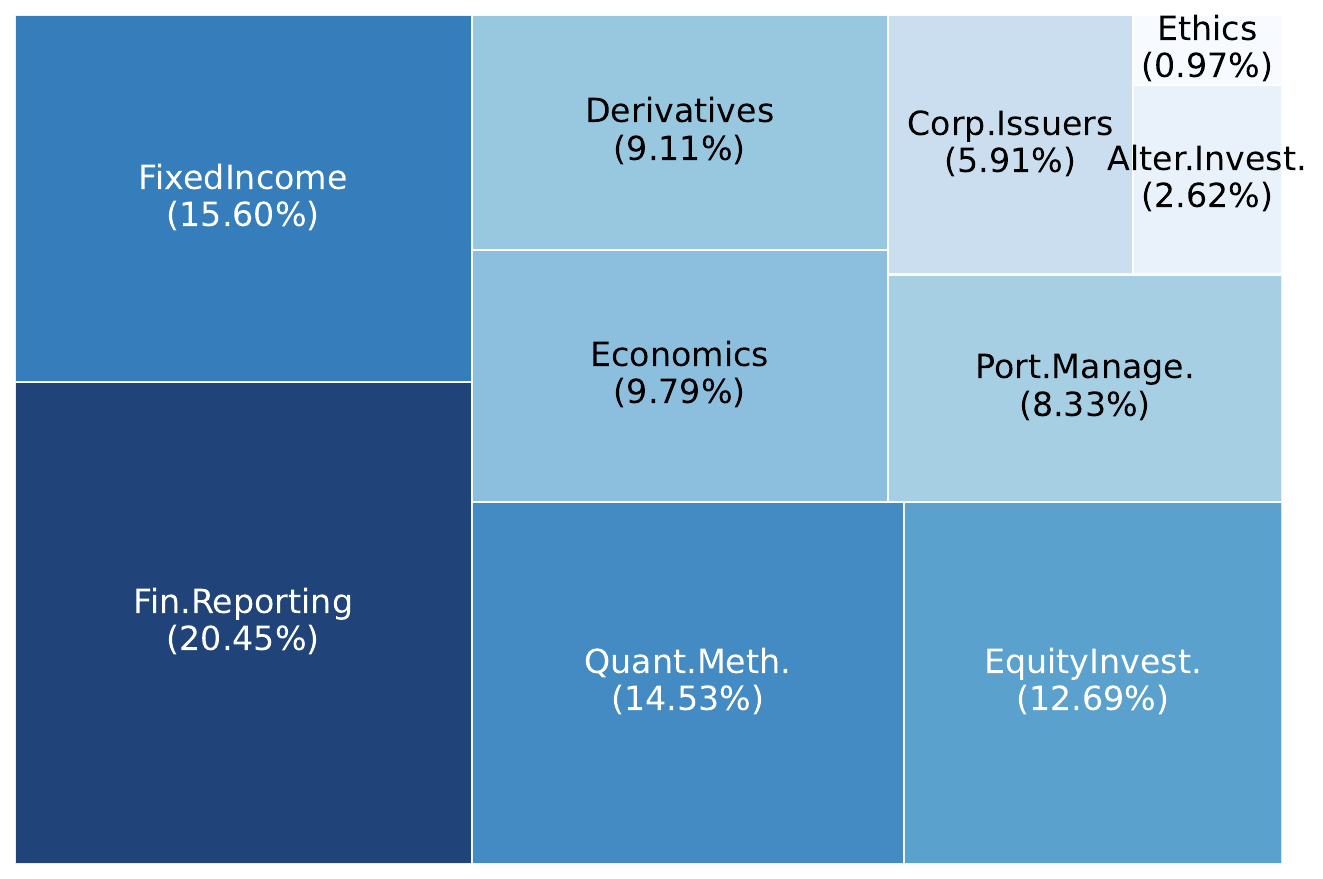}}
    \caption{GPT-4o classified the benchmark domain distribution of CFA. A random sample of 100 items was manually audited by a financial expert to validate domain labels.}
    \label{fig:domain_distribution}
\end{figure}
\clearpage
\section{Average Input and Output Tokens}\label{appendix:tab_average_input_output_tokens}
\subsection{Average Input Tokens}\label{appendix:tab_average_input_tokens}
\begin{table*}[htbp]
  \centering
  \scriptsize
  \setlength\tabcolsep{1.0pt}
  \renewcommand\arraystretch{1.0}
  \resizebox{\textwidth}{!}{%
    \begin{tabular}{lccccccccc}
      \toprule
      \multirow{2}{*}{\textbf{Prompt}} &
      \multicolumn{9}{c}{\textbf{Average Input Tokens (k)}} \\
      \cmidrule(lr){2-10}
      & \makecell{Qwen2.5-7B}
      & \makecell{Qwen2.5-7B\\Instruct}
      & \makecell{Qwen3-8B\\Base}
      & Qwen3-8B
      & \makecell{Gemma-3-12B\\IT}
      & \makecell{Qwen3-8B\\(Thinker)}
      & Fin-R1
      & \makecell{DianJin-R1\\7B}
      & \makecell{Fin-o1-8B} \\
      \midrule
      SP                      & \textbf{0.07}* & \textbf{0.07}* & \textbf{0.07}* & \textbf{0.07}* & \textbf{0.07}* & \textbf{0.07}* & \textbf{0.07}* & \textbf{0.07}* & \textbf{0.07}* \\
      UST-CoT                 & 0.09 & 0.09 & 0.09 & 0.09 & 0.09 & 0.09 & 0.09 & 0.09 & 0.09 \\
      ST-CoT                  & 0.18 & 0.18 & 0.18 & 0.18 & 0.19 & 0.18 & 0.18 & 0.18 & 0.18 \\
      FinCoT (All Blueprints) & 1.75 & 1.75 & 1.75 & 1.75 & 1.78 & 1.75 & 1.75 & 1.75 & 1.75 \\
      \midrule
      \rowcolor{gray!16}
      \multicolumn{10}{c}{\textbf{Domain-wise performance of FinCoT}} \\
      \midrule
      FinCoT (Economics)      & 0.55 & 0.55 & 0.55 & 0.55 & 0.56 & 0.55 & 0.55 & 0.55 & 0.55 \\
      FinCoT (FixedIncome)    & 0.34 & 0.34 & 0.34 & 0.34 & 0.36 & 0.34 & 0.34 & 0.34 & 0.34 \\
      FinCoT (Quant.Meth.)    & 0.33 & 0.33 & 0.33 & 0.33 & 0.34 & 0.33 & 0.33 & 0.33 & 0.33 \\
      FinCoT (EquityInvest.)  & 0.34 & 0.34 & 0.34 & 0.34 & 0.36 & 0.34 & 0.34 & 0.34 & 0.34 \\
      FinCoT (Port.Manage.)   & 0.33 & 0.33 & 0.33 & 0.33 & 0.35 & 0.33 & 0.33 & 0.33 & 0.33 \\
      FinCoT (Derivatives)    & 0.39 & 0.39 & 0.39 & 0.39 & 0.44 & 0.39 & 0.39 & 0.39 & 0.39 \\
      FinCoT (Fin. Reporting) & 0.37 & 0.37 & 0.37 & 0.37 & 0.38 & 0.37 & 0.37 & 0.37 & 0.37 \\
      FinCoT (Alter.Invest.)  & 0.32 & 0.32 & 0.32 & 0.32 & 0.34 & 0.32 & 0.32 & 0.32 & 0.32 \\
      FinCoT (Corp.Issuers)   & 0.39 & 0.39 & 0.39 & 0.39 & 0.41 & 0.39 & 0.39 & 0.39 & 0.39 \\
      \bottomrule
    \end{tabular}%
  }
  \caption{Comparison of prompting techniques: average input token length (k) across models. \textbf{Bold} values highlight the prompt variant that uses the least tokens for each model.}
  \label{tab:input_tokens}
\end{table*}
\subsection{Average Output Tokens}\label{appendix:tab_average_output_tokens}
\begin{table*}[htbp]
  \centering
  \scriptsize
  \setlength\tabcolsep{1.0pt}
  \renewcommand\arraystretch{1.0}
  \resizebox{\textwidth}{!}{%
    \begin{tabular}{lccccccccc}
      \toprule
      \multirow{2}{*}{\textbf{Prompt}} &
      \multicolumn{9}{c}{\textbf{Average Output Tokens (k)}} \\
      \cmidrule(lr){2-10}
      & \makecell{Qwen2.5-7B}
      & \makecell{Qwen2.5-7B\\Instruct}
      & \makecell{Qwen3-8B\\Base}
      & Qwen3-8B
      & \makecell{Gemma-3-12B\\IT}
      & \makecell{Qwen3-8B\\(Thinker)}
      & Fin-R1
      & \makecell{DianJin-R1\\7B}
      & \makecell{Fin-o1-8B} \\
      \midrule
      SP                      & 0.45  & \textbf{0.05}* & 0.89  & 0.32  & \textbf{0.27}*  & 1.52  & 0.88  & 2.18  & \textbf{0.46}* \\
      UST‑CoT                 & 0.48  & 0.28           & \textbf{0.31}* & 0.46  & 0.39  & 1.50  & \textbf{0.58}* & 2.28  & 0.53 \\
      ST‑CoT                  & \textbf{0.39}*  & 0.22           & 3.42  & \textbf{0.25}* & 0.31  & 1.35  & 2.22  & 7.20  & 0.58 \\
      FinCoT (All Blueprints) & 2.22  & 0.29           & 0.38  & 0.36  & 0.32  & \textbf{1.23}* & 1.92  & \textbf{1.60}*  & 0.79 \\
      \midrule
      \rowcolor{gray!16}
      \multicolumn{10}{c}{\textbf{Domain‑wise performance of FinCoT}} \\
      \midrule
      FinCoT (Economics)      & 0.36  & 0.38 & 0.99  & 0.39  & 0.38  & 1.25  & \textbf{2.01} & 12.65 & 0.76 \\
      FinCoT (FixedIncome)    & 0.42  & 0.27 & 4.55  & 0.30  & \textbf{0.32}  & 1.24  & 2.31         & \textbf{8.31}  & 0.81 \\
      FinCoT (Quant.Meth.)    & 0.48  & 0.27 & 3.07  & \textbf{0.31} & 0.35  & 1.22  & 2.17         & 8.60  & 0.80 \\
      FinCoT (EquityInvest.)  & \textbf{0.32}  & 0.31 & 7.18  & 0.37  & 0.34  & 1.19  & 2.16         & 10.07 & 0.78 \\
      FinCoT (Port.Manage.)   & 0.38  & \textbf{0.26} & 0.56  & 0.30  & 0.33  & 1.20  & 2.14         & 9.46  & 0.79 \\
      FinCoT (Derivatives)    & 0.36  & 0.30 & \textbf{0.42} & 0.39  & 0.34  & 1.24  & 2.05         & 5.54  & 0.81 \\
      FinCoT (Fin. Reporting) & 0.46  & 0.28 & 0.93  & 0.33  & 0.34  & 1.19  & 2.13         & 8.76  & \textbf{0.73} \\
      FinCoT (Alter.Invest.)  & 0.47  & \textbf{0.26} & 0.50  & 0.38  & 0.34  & 1.23  & 2.16         & 11.53 & 0.77 \\
      FinCoT (Corp.Issuers)   & 0.52  & \textbf{0.26} & 1.18  & 0.32  & 0.33  & \textbf{1.16} & 2.08     & 11.37 & 0.82 \\
      \bottomrule
    \end{tabular}%
  }
  \caption{Comparison of prompting techniques: average output token length (k) across models. \textbf{Bold} values highlight the prompt variant that uses the least tokens for each model. (*) Indicates that the change in average output token count among the model‐level prompt variants is statistically significant ($p<0.05$) based on paired bootstrap testing; domain‐specific rows are not tested for significance.}
  \label{tab:output_tokens}
\end{table*}

\Needspace{0.95\textheight}
\subsubsection{Efficiency of Input and Output Cost in Simulation}
\label{sec:efficiency-simulation}

\begingroup
\setlength{\abovedisplayskip}{4pt}
\setlength{\belowdisplayskip}{4pt}
\setlength{\abovecaptionskip}{4pt}
\setlength{\belowcaptionskip}{0pt}

\noindent\small
This appendix reports a cost--efficiency analysis under realistic output--input price ratios.
Let $I$ and $O$ denote the average input and output tokens for a (prompt, model) pair.
For a price ratio $r$,
\[
  \mathrm{Cost}(r)= I + r\,O,\qquad
  \mathrm{Efficiency}(r)=
  \frac{\mathrm{Cost}_{\mathrm{baseline}}(r)}{\mathrm{Cost}_{\mathrm{prompt}}(r)}
  = \frac{I_{\mathrm{base}} + r\,O_{\mathrm{base}}}{I_{\mathrm{prompt}} + r\,O_{\mathrm{prompt}}}.
\]

\noindent\textbf{Units and normalization.}
We measure cost in ``input-token dollars'': the effective input price is $1$, and
$r=\text{price}_{\text{out}}/\text{price}_{\text{in,eff}}$ carries the output premium.
This rescaling makes $\mathrm{Efficiency}$ dimensionless and invariant to any common price factor.

\noindent\textbf{Break-even and sensitivity.}
For a candidate prompt $p$ vs.\ baseline $b$, the break-even ratio solving
$\mathrm{Cost}_p(r)=\mathrm{Cost}_b(r)$ is
\[
  r^\star=\frac{I_p-I_b}{\,O_b-O_p\,}\qquad (O_p\neq O_b).
\]
If $O_p<O_b$, then $\dfrac{d}{dr}\mathrm{Efficiency}(r)=\dfrac{O_b I_p - O_p I_b}{(I_p+rO_p)^2}>0$:
the candidate improves as $r$ increases; if $O_p>O_b$, the trend reverses.
When $O_p=O_b$, ranking depends only on inputs ($I_b$ vs.\ $I_p$) and is independent of $r$.

\noindent\textbf{Caching and effective input price.}
With prompt caching,
\[
  p_{\mathrm{in,eff}}(K)=p_{\mathrm{read}}+\frac{p_{\mathrm{write}}}{K},\qquad
  r(K)=\frac{\text{price}_{\mathrm{out}}}{p_{\mathrm{in,eff}}(K)},
\]
where $K$ is the number of reuses. Hence $r(K)$ increases monotonically in $K$ and approaches
$\text{price}_{\mathrm{out}}/p_{\mathrm{read}}$ as $K\!\to\!\infty$.

\noindent\textbf{Price instantiation (grid for plots).}
From public price points we use $r\in\{5,\,6.9,\,8,\,14.29,\,22.22,\,40,\,44.44,\,50,\,80\}$:
\emph{GPT-5}\footnote{OpenAI API pricing: \url{https://openai.com/api/pricing/}.}
input \$1.25/MTok (cached \$0.125/MTok), output \$10/MTok
$\Rightarrow r{=}10/1.25{=}8$, $r_{\mathrm{cached}}{=}10/0.125{=}80$;
\emph{Claude Opus 4.1}\footnote{Anthropic pricing: \url{https://www.anthropic.com/pricing\#api}.}
input \$15/MTok, output \$75/MTok; caching write \$18.75/MTok, read \$1.50/MTok
$\Rightarrow r{=}5$ (no cache), $6.9$ ($K{=}2$), $14.29$ ($K{=}5$), $22.22$ ($K{=}10$),
$40$ ($K{=}50$), $44.44$ ($K{=}100$), and the read-only limit $50$ ($K{\to}\infty$).
We display $r$ on a log scale because the grid spans an order of magnitude (5--80).

\noindent\textbf{Worked example (illustrative).}
Baseline $(I_b,O_b)=(100,300)$; candidate $(I_p,O_p)=(250,150)$.
At $r=8$ (GPT-5 no cache): $\mathrm{Cost}_b=100+8\!\cdot\!300=2500$,
$\mathrm{Cost}_p=250+8\!\cdot\!150=1450$, so $\mathrm{Efficiency}\approx 1.72$.
Break-even $r^\star=(250-100)/(300-150)=1$; the candidate dominates for $r>1$.

\noindent\textbf{Note (scope).}
All models evaluated in this appendix are \emph{open-source}. The curves
simulate dollar costs by pairing the measured $(I,O)$ token counts from these models with
\emph{provider API prices} (GPT-5 and Claude Opus 4.1)---prices are used for \emph{simulation only};
no paid API runs were executed for these experiments.

\vspace{6pt}

\begin{minipage}{\linewidth}\centering
  \includegraphics[width=\linewidth,height=0.45\textheight,keepaspectratio]{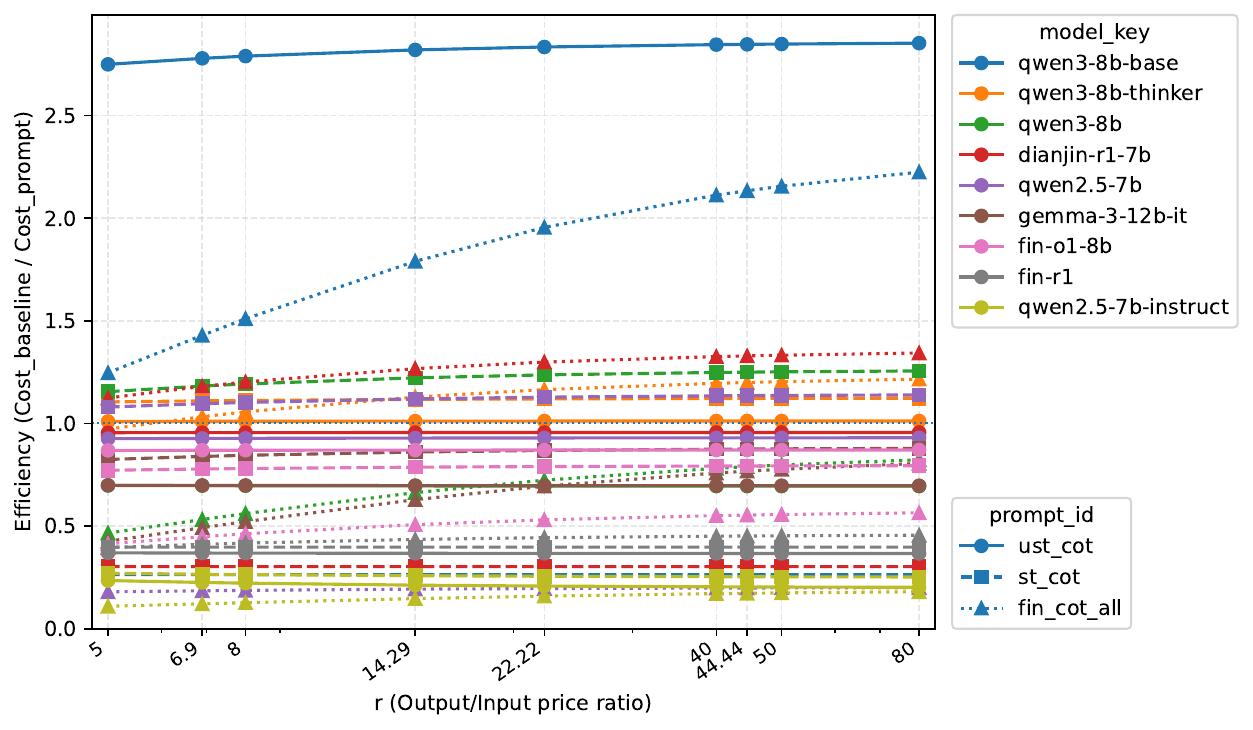}
  \captionof{figure}{\textbf{Cost--efficiency vs.\ price ratio $r$} for UST-CoT, ST-CoT, and
  FinCoT-All across models. Efficiency is
  $\mathrm{Cost}_{\mathrm{baseline}}/\mathrm{Cost}_{\mathrm{prompt}}$ with $\mathrm{Cost}=I+rO$;
  values $>1$ indicate lower cost than the baseline. $r$ values use provider prices for
  GPT-5 and Claude Opus~4.1 as described above. \emph{Notation:} MTok = million tokens; USD per MTok.}
  \label{fig:efficiency-input-output}
\end{minipage}
\endgroup

\clearpage
\section{Significance Testing}
\subsection{Accuracy}
\begin{table*}[!htbp]
  \centering
  \scriptsize
  \setlength{\tabcolsep}{3pt}
  \renewcommand\arraystretch{1.1}
  \begin{adjustbox}{max width=\textwidth, max totalheight=\textheight, keepaspectratio}
    \begin{tabular}{lllcclc}
      \toprule
      \textbf{Model} & \textbf{Baseline} & \textbf{Comparison} & $\Delta$ (pp) & 95\% CI (pp)       & $p$-value & Significant \\
      \midrule
      \multirow{6}{*}{Qwen2.5-7B}
          & SP      & UST-CoT        & -13.76  & [-15.89, -11.63] & 0.0000 & \checkmark \\
          & SP      & ST-CoT         & -16.27  & [-18.60, -14.05] & 0.0000 & \checkmark \\
          & SP      & FinCoT (All)   &  -7.94  & [ -9.59,  -6.30] & 0.0000 & \checkmark \\
          & UST-CoT & ST-CoT         &  -2.52  & [ -3.49,  -1.65] & 0.0000 & \checkmark \\
          & UST-CoT & FinCoT (All)   &   5.81  & [  4.46,   7.27] & 0.0000 & \checkmark \\
          & ST-CoT  & FinCoT (All)   &   8.33  & [  6.69,   9.98] & 0.0000 & \checkmark \\
      \midrule
      \multirow{6}{*}{Qwen2.5-7B-Instruct}
          & SP      & UST-CoT        &  6.00   & [ 4.65,   7.56] & 0.0000 & \checkmark \\
          & SP      & ST-CoT         &  4.84   & [ 3.59,   6.20] & 0.0000 & \checkmark \\
          & SP      & FinCoT (All)   &  4.55   & [ 3.29,   5.91] & 0.0000 & \checkmark \\
          & UST-CoT & ST-CoT         & -1.16   & [-1.84,  -0.58] & 0.0000 & \checkmark \\
          & UST-CoT & FinCoT (All)   & -1.45   & [-2.23,  -0.78] & 0.0000 & \checkmark \\
          & ST-CoT  & FinCoT (All)   & -0.29   & [-0.68,   0.00] & 0.1024 & --        \\
      \midrule
      \multirow{6}{*}{Qwen3-8B-Base}
          & SP      & UST-CoT        & -9.40   & [-11.24, -7.66] & 0.0000 & \checkmark \\
          & SP      & ST-CoT         & -15.31  & [-17.54,-13.18] & 0.0000 & \checkmark \\
          & SP      & FinCoT (All)   & -17.35  & [-19.77,-15.02] & 0.0000 & \checkmark \\
          & UST-CoT & ST-CoT         & -5.91   & [-7.36,  -4.55] & 0.0000 & \checkmark \\
          & UST-CoT & FinCoT (All)   & -7.96   & [-9.59,  -6.30] & 0.0000 & \checkmark \\
          & ST-CoT  & FinCoT (All)   & -2.04   & [-3.00,  -1.26] & 0.0000 & \checkmark \\
      \midrule
      \multirow{6}{*}{Qwen3-8B}
          & SP      & UST-CoT        &  7.95   & [ 6.30,   9.69] & 0.0000 & \checkmark \\
          & SP      & ST-CoT         &  6.60   & [ 5.14,   8.14] & 0.0000 & \checkmark \\
          & SP      & FinCoT (All)   &  6.70   & [ 5.23,   8.24] & 0.0000 & \checkmark \\
          & UST-CoT & ST-CoT         & -1.35   & [-2.13,  -0.68] & 0.0000 & \checkmark \\
          & UST-CoT & FinCoT (All)   & -1.26   & [-1.94,  -0.68] & 0.0000 & \checkmark \\
          & ST-CoT  & FinCoT (All)   &  0.10   & [ 0.00,   0.29] & 0.7370 & --        \\
      \midrule
      \multirow{6}{*}{Qwen3-8B (Thinker)}
          & SP      & UST-CoT        & -0.87   & [-1.45,  -0.39] & 0.0002 & \checkmark \\
          & SP      & ST-CoT         &  0.00   & [ 0.00,   0.00] & 2.0000 & --        \\
          & SP      & FinCoT (All)   &  0.96   & [ 0.39,   1.55] & 0.0002 & \checkmark \\
          & UST-CoT & ST-CoT         &  0.87   & [ 0.39,   1.45] & 0.0002 & \checkmark \\
          & UST-CoT & FinCoT (All)   &  1.83   & [ 1.07,   2.71] & 0.0000 & \checkmark \\
          & ST-CoT  & FinCoT (All)   &  0.96   & [ 0.39,   1.55] & 0.0002 & \checkmark \\
      \midrule
      \multirow{6}{*}{Gemma-3-12B-IT}
          & SP      & UST-CoT        & -24.999 & [-27.71, -22.38] & 0.0000 & \checkmark \\
          & SP      & ST-CoT         & -23.934 & [-26.55, -21.32] & 0.0000 & \checkmark \\
          & SP      & FinCoT (All)   & -22.765 & [-25.39, -20.16] & 0.0000 & \checkmark \\
          & UST-CoT & ST-CoT         &   1.065 & [  0.48,   1.74] & 0.0002 & \checkmark \\
          & UST-CoT & FinCoT (All)   &   2.234 & [  1.36,   3.20] & 0.0000 & \checkmark \\
          & ST-CoT  & FinCoT (All)   &   1.169 & [  0.58,   1.84] & 0.0000 & \checkmark \\
      \midrule
      \multirow{6}{*}{Fin-R1}
          & SP      & UST-CoT        & -9.49   & [-11.34, -7.75] & 0.0000 & \checkmark \\
          & SP      & ST-CoT         & -8.62   & [-10.37, -6.88] & 0.0000 & \checkmark \\
          & SP      & FinCoT (All)   & -10.07  & [-11.92,-8.24]  & 0.0000 & \checkmark \\
          & UST-CoT & ST-CoT         &  0.87   & [ 0.39,   1.45] & 0.0002 & \checkmark \\
          & UST-CoT & FinCoT (All)   & -0.58   & [-1.07,  -0.19] & 0.0042 & \checkmark \\
          & ST-CoT  & FinCoT (All)   & -1.45   & [-2.23,  -0.78] & 0.0000 & \checkmark \\
      \midrule
      \multirow{6}{*}{Dianjin-R1-7B}
          & SP      & UST-CoT        &  10.662 & [  8.82,  12.60] & 0.0000 & \checkmark \\
          & SP      & ST-CoT         &   9.594 & [  7.75,  11.43] & 0.0000 & \checkmark \\
          & SP      & FinCoT (All)   &  -1.361 & [ -2.13,  -0.68] & 0.0000 & \checkmark \\
          & UST-CoT & ST-CoT         &  -1.068 & [ -1.74,  -0.48] & 0.0000 & \checkmark \\
          & UST-CoT & FinCoT (All)   & -12.023 & [-14.05, -10.08] & 0.0000 & \checkmark \\
          & ST-CoT  & FinCoT (All)   & -10.956 & [-12.89,  -9.01] & 0.0000 & \checkmark \\
      \midrule
      \multirow{6}{*}{Fino1-8B}
          & SP      & UST-CoT        &   0.294 & [  0.00,   0.68] & 0.0984 & --         \\
          & SP      & ST-CoT         &   1.264 & [  0.58,   1.94] & 0.0000 & \checkmark \\
          & SP      & FinCoT (All)   &   2.429 & [  1.55,   3.39] & 0.0000 & \checkmark \\
          & UST-CoT & ST-CoT         &   0.970 & [  0.39,   1.55] & 0.0000 & \checkmark \\
          & UST-CoT & FinCoT (All)   &   2.134 & [  1.26,   3.00] & 0.0000 & \checkmark \\
          & ST-CoT  & FinCoT (All)   &   1.165 & [  0.58,   1.84] & 0.0000 & \checkmark \\
      \bottomrule
    \end{tabular}
  \end{adjustbox}
  \caption{Paired bootstrap significance testing ($B = 10{,}000$ samples) for accuracy differences across prompt strategies. $\Delta$ indicates average accuracy difference (in percentage points), with 95\% confidence intervals (CI) and $p$-values. A result is considered statistically significant if $p < 0.05$.}
  \label{tab:significance_accuracy}
\end{table*}

\clearpage
\subsection{Average Output Tokens}
\begin{table*}[!htbp]
  \centering
  \scriptsize
  \setlength{\tabcolsep}{3pt}
  \renewcommand\arraystretch{1.1}
  \begin{adjustbox}{max width=\textwidth, max totalheight=\textheight, keepaspectratio}
    \begin{tabular}{lllcclc}
      \toprule
      \textbf{Model} & \textbf{Baseline} & \textbf{Comparison} & $\Delta$ (k) & 95\% CI (k)       & $p$-value & Significant \\
      \midrule
      \multirow{6}{*}{Qwen2.5-7B}
          & SP      & UST-CoT        & -0.09867 & [-0.09867, -0.09867] & 0.0000 & \checkmark \\
          & SP      & ST-CoT         &  0.02273 & [ 0.02273,  0.02273] & 0.0000 & \checkmark \\
          & SP      & FinCoT (All)   & -0.11555 & [-0.11555, -0.11555] & 0.0000 & \checkmark \\
          & UST-CoT & ST-CoT         &  0.12140 & [ 0.12140,  0.12140] & 0.0000 & \checkmark \\
          & UST-CoT & FinCoT (All)   & -0.01688 & [-0.01688, -0.01688] & 0.0000 & \checkmark \\
          & ST-CoT  & FinCoT (All)   & -0.13828 & [-0.13828, -0.13828] & 0.0000 & \checkmark \\
      \midrule
      \multirow{6}{*}{Qwen2.5-7B-Instruct}
          & SP      & UST-CoT        & -0.227   & [-0.227, -0.227] & 0.0000 & \checkmark \\
          & SP      & ST-CoT         & -0.169   & [-0.169, -0.169] & 0.0000 & \checkmark \\
          & SP      & FinCoT (All)   & -0.241   & [-0.241, -0.241] & 0.0000 & \checkmark \\
          & UST-CoT & ST-CoT         &  0.058   & [ 0.058,  0.058] & 0.0000 & \checkmark \\
          & UST-CoT & FinCoT (All)   & -0.014   & [-0.014, -0.014] & 0.0000 & \checkmark \\
          & ST-CoT  & FinCoT (All)   & -0.072   & [-0.072, -0.072] & 0.0000 & \checkmark \\
      \midrule
      \multirow{6}{*}{Qwen3-8B-Base}
          & SP      & UST-CoT        & -0.584   & [-0.584, -0.584] & 0.0000 & \checkmark \\
          & SP      & ST-CoT         &  2.522   & [ 2.522,  2.522] & 0.0000 & \checkmark \\
          & SP      & FinCoT (All)   & -0.516   & [-0.516, -0.516] & 0.0000 & \checkmark \\
          & UST-CoT & ST-CoT         &  3.106   & [ 3.106,  3.106] & 0.0000 & \checkmark \\
          & UST-CoT & FinCoT (All)   &  0.068   & [ 0.068,  0.068] & 0.0000 & \checkmark \\
          & ST-CoT  & FinCoT (All)   & -3.038   & [-3.038, -3.038] & 0.0000 & \checkmark \\
      \midrule
      \multirow{6}{*}{Qwen3-8B}
          & SP      & UST-CoT        &  0.141   & [ 0.141,  0.141] & 0.0000 & \checkmark \\
          & SP      & ST-CoT         & -0.067   & [-0.067, -0.067] & 0.0000 & \checkmark \\
          & SP      & FinCoT (All)   &  0.048   & [ 0.048,  0.048] & 0.0000 & \checkmark \\
          & UST-CoT & ST-CoT         & -0.208   & [-0.208, -0.208] & 0.0000 & \checkmark \\
          & UST-CoT & FinCoT (All)   & -0.093   & [-0.093, -0.093] & 0.0000 & \checkmark \\
          & ST-CoT  & FinCoT (All)   &  0.115   & [ 0.115,  0.115] & 0.0000 & \checkmark \\
      \midrule
      \multirow{6}{*}{Qwen3-8B (Thinker)}
          & SP      & UST-CoT        & -0.018   & [-0.018, -0.018] & 0.0000 & \checkmark \\
          & SP      & ST-CoT         & -1.271   & [-1.271, -1.271] & 0.0000 & \checkmark \\
          & SP      & FinCoT (All)   & -0.168   & [-0.168, -0.168] & 0.0000 & \checkmark \\
          & UST-CoT & ST-CoT         & -1.253   & [-1.253, -1.253] & 0.0000 & \checkmark \\
          & UST-CoT & FinCoT (All)   & -0.150   & [-0.150, -0.150] & 0.0000 & \checkmark \\
          & ST-CoT  & FinCoT (All)   &  1.103   & [ 1.103,  1.103] & 0.0000 & \checkmark \\
      \midrule
      \multirow{6}{*}{Gemma-3-12B-IT}
          & SP      & UST-CoT        &  0.11985 & [ 0.11985,  0.11985] & 0.0000 & \checkmark \\
          & SP      & ST-CoT         &  0.03661 & [ 0.03661,  0.03661] & 0.0000 & \checkmark \\
          & SP      & FinCoT (All)   &  0.04523 & [ 0.04523,  0.04523] & 0.0000 & \checkmark \\
          & UST-CoT & ST-CoT         & -0.08324 & [-0.08324, -0.08324] & 0.0000 & \checkmark \\
          & UST-CoT & FinCoT (All)   & -0.07462 & [-0.07462, -0.07462] & 0.0000 & \checkmark \\
          & ST-CoT  & FinCoT (All)   &  0.00862 & [ 0.00862,  0.00862] & 0.0000 & \checkmark \\
      \midrule
      \multirow{6}{*}{Fin-R1}
          & SP      & UST-CoT        &  1.526   & [ 1.526,  1.526] & 0.0000 & \checkmark \\
          & SP      & ST-CoT         &  1.338   & [ 1.338,  1.338] & 0.0000 & \checkmark \\
          & SP      & FinCoT (All)   &  1.035   & [ 1.035,  1.035] & 0.0000 & \checkmark \\
          & UST-CoT & ST-CoT         & -0.188   & [-0.188, -0.188] & 0.0000 & \checkmark \\
          & UST-CoT & FinCoT (All)   & -0.491   & [-0.491, -0.491] & 0.0000 & \checkmark \\
          & ST-CoT  & FinCoT (All)   & -0.303   & [-0.303, -0.303] & 0.0000 & \checkmark \\
      \midrule
      \multirow{6}{*}{Dianjin-R1-7B}
          & SP      & UST-CoT        &  0.10023 & [ 0.10023,  0.10023] & 0.0000 & \checkmark \\
          & SP      & ST-CoT         &  5.02159 & [ 5.02159,  5.02159] & 0.0000 & \checkmark \\
          & SP      & FinCoT (All)   & -0.57669 & [-0.57669, -0.57669] & 0.0000 & \checkmark \\
          & UST-CoT & ST-CoT         &  4.92136 & [ 4.92136,  4.92136] & 0.0000 & \checkmark \\
          & UST-CoT & FinCoT (All)   & -0.67692 & [-0.67692, -0.67692] & 0.0000 & \checkmark \\
          & ST-CoT  & FinCoT (All)   & -5.59828 & [-5.59828, -5.59828] & 0.0000 & \checkmark \\
      \midrule
      \multirow{6}{*}{Fino1-8B}
          & SP      & UST-CoT        &  0.06788 & [ 0.06788,  0.06788] & 0.0000 & \checkmark \\
          & SP      & ST-CoT         &  0.11765 & [ 0.11765,  0.11765] & 0.0000 & \checkmark \\
          & SP      & FinCoT (All)   &  0.33273 & [ 0.33273,  0.33273] & 0.0000 & \checkmark \\
          & UST-CoT & ST-CoT         &  0.04977 & [ 0.04977,  0.04977] & 0.0000 & \checkmark \\
          & UST-CoT & FinCoT (All)   &  0.26485 & [ 0.26485,  0.26485] & 0.0000 & \checkmark \\
          & ST-CoT  & FinCoT (All)   &  0.21508 & [ 0.21508,  0.21508] & 0.0000 & \checkmark \\
      \bottomrule
    \end{tabular}
  \end{adjustbox}
  \caption{Paired bootstrap significance testing ($B = 10{,}000$ samples) for average output token differences across prompt strategies. $\Delta$ indicates mean difference in output length (in thousands of tokens), with 95\% confidence intervals and $p$-values. A result is significant if $p < 0.05$.}
  \label{tab:significance_tokens}
\end{table*}
\clearpage
\section{Radar Behavior Accuracy}\label{appendix:radar_behavior_accuracy}
\begin{figure}[!ht]
  \centering
  \includegraphics[width=\linewidth,keepaspectratio]{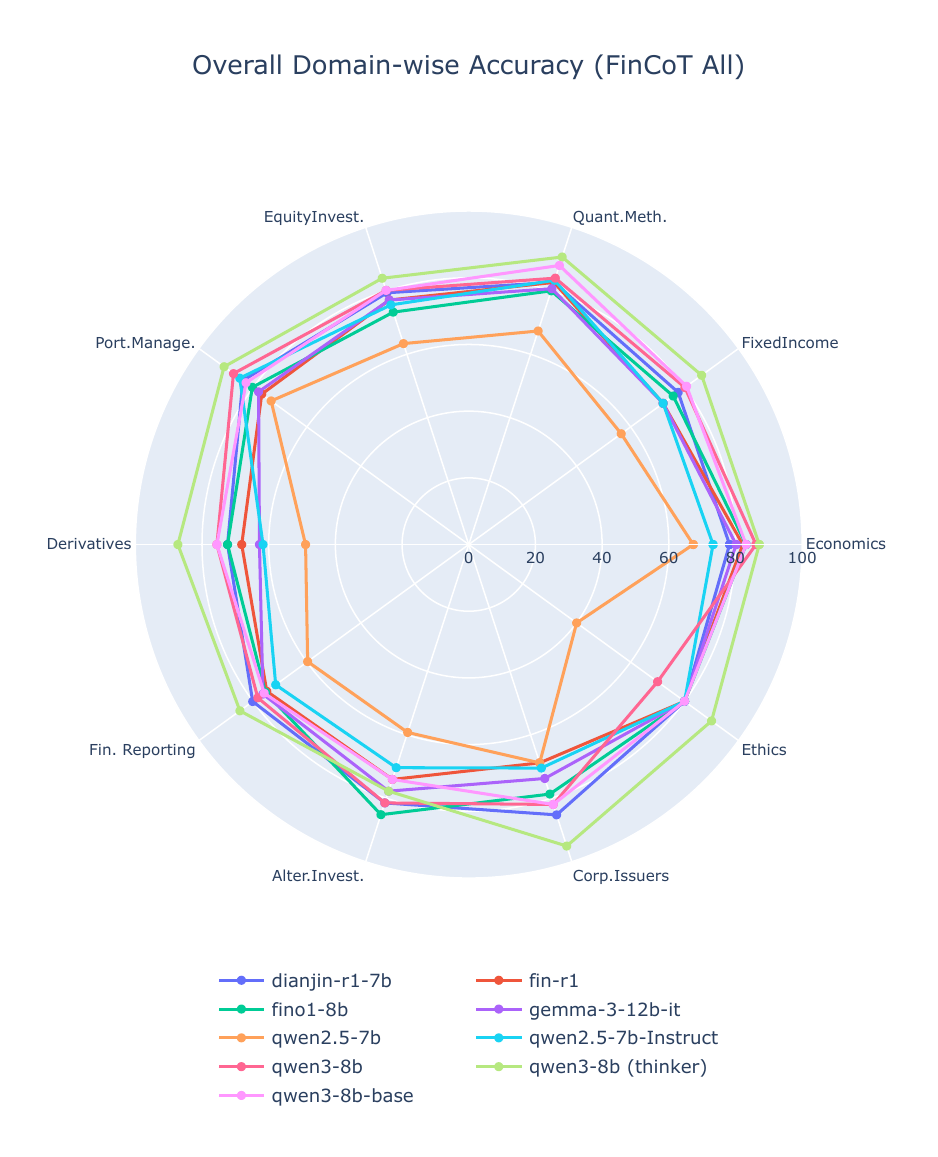}
  \caption{Overall FinCoT behaviour accuracy.}
  \label{fig:overall_FinCoT_All}
\end{figure}

\clearpage
\begin{figure*}[!ht]
  \centering
  \begin{subfigure}{0.48\linewidth}
    \centering
    \includegraphics[width=\linewidth,keepaspectratio]{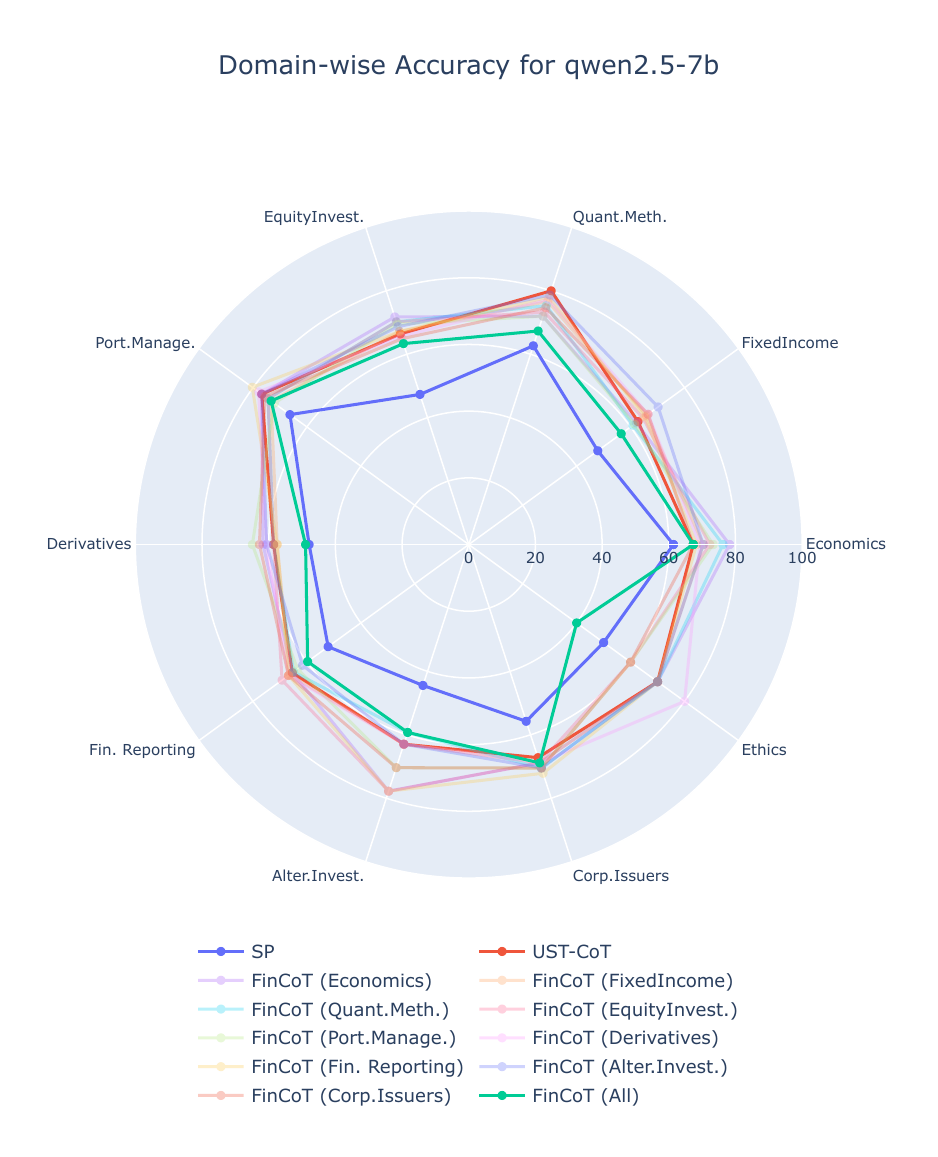}
    \caption{Qwen2.5-7B}
  \end{subfigure}
  \hfill
  \begin{subfigure}{0.48\linewidth}
    \centering
    \includegraphics[width=\linewidth,keepaspectratio]{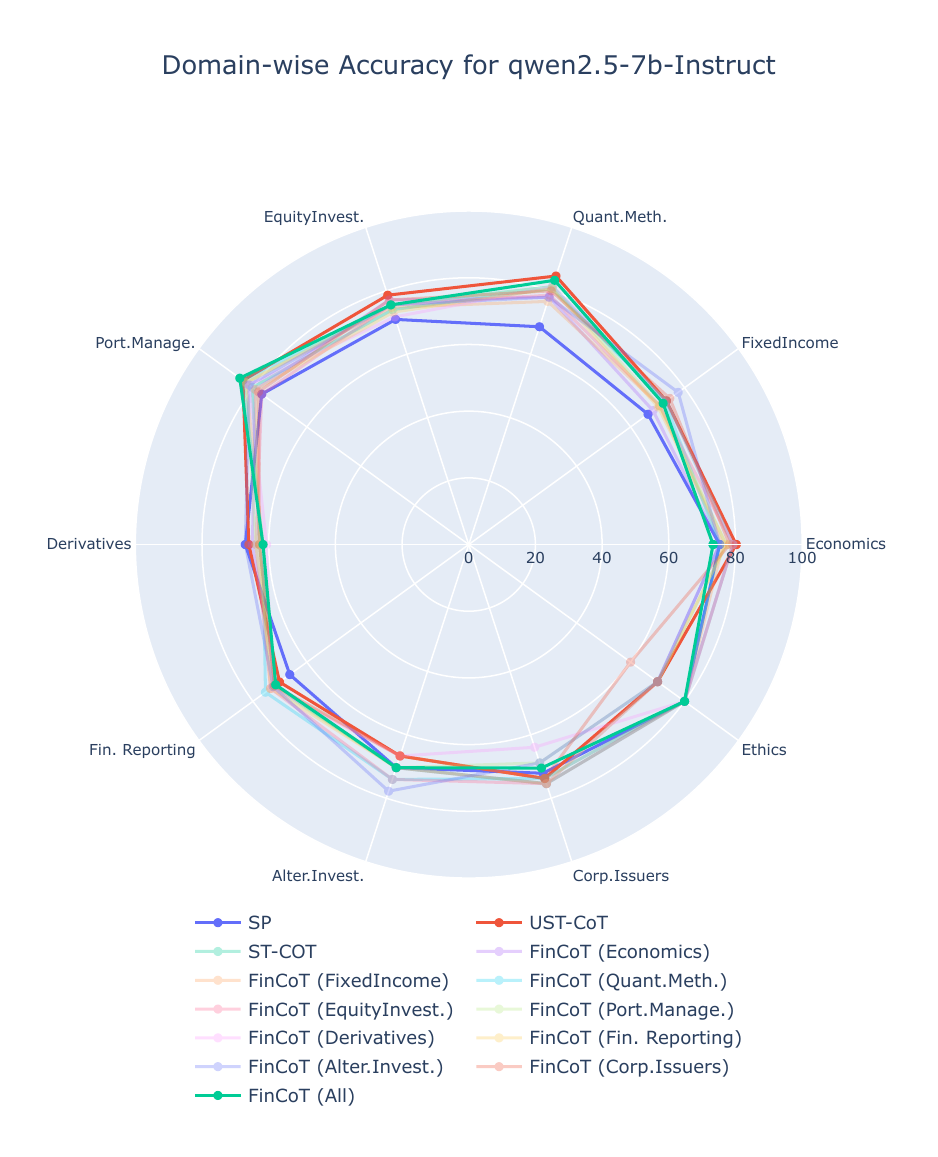}
    \caption{Qwen2.5-7B-Instruct}
  \end{subfigure}

  \vspace{1em} 

  \begin{subfigure}{0.48\linewidth}
    \centering
    \includegraphics[width=\linewidth,keepaspectratio]{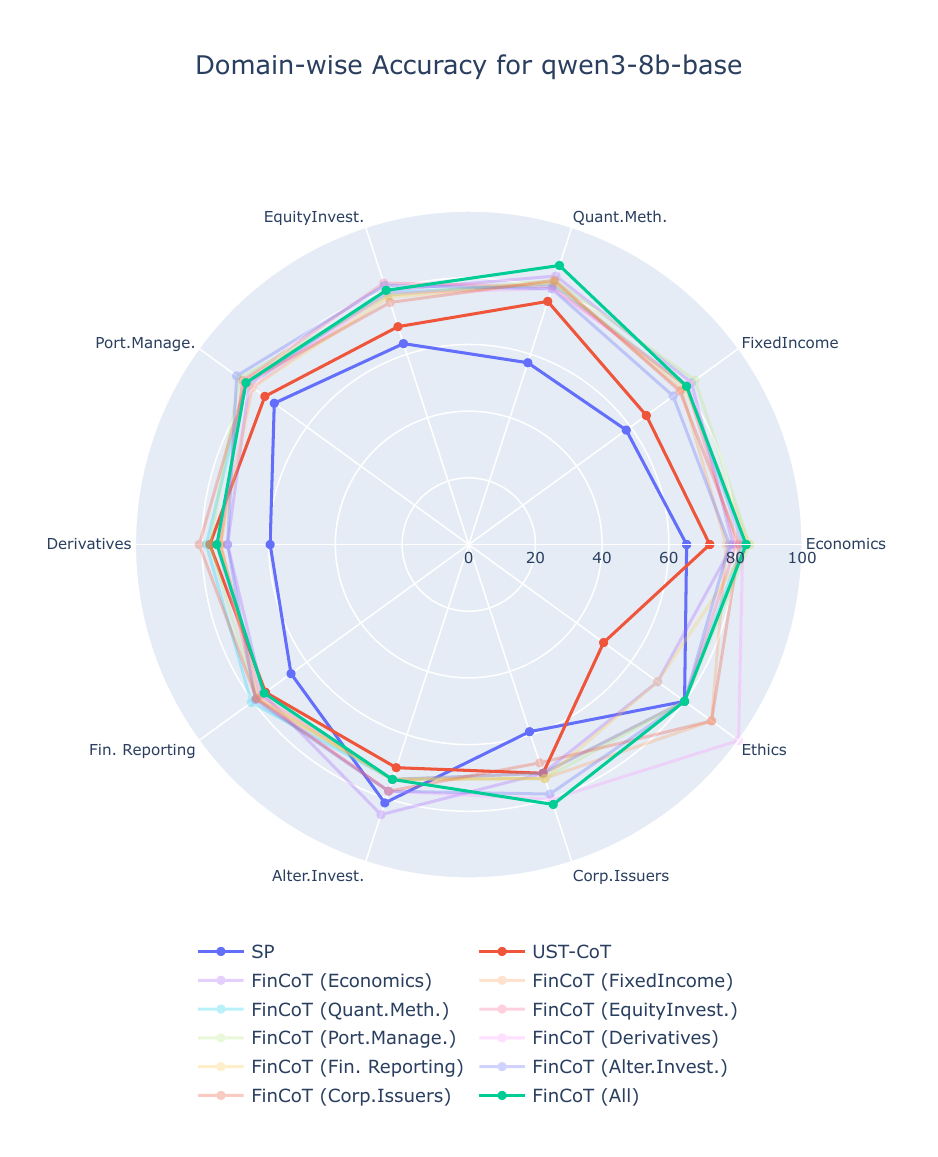}
    \caption{Qwen3-8B-Base}
  \end{subfigure}
  \hfill
  \begin{subfigure}{0.48\linewidth}
    \centering
    \includegraphics[width=\linewidth,keepaspectratio]{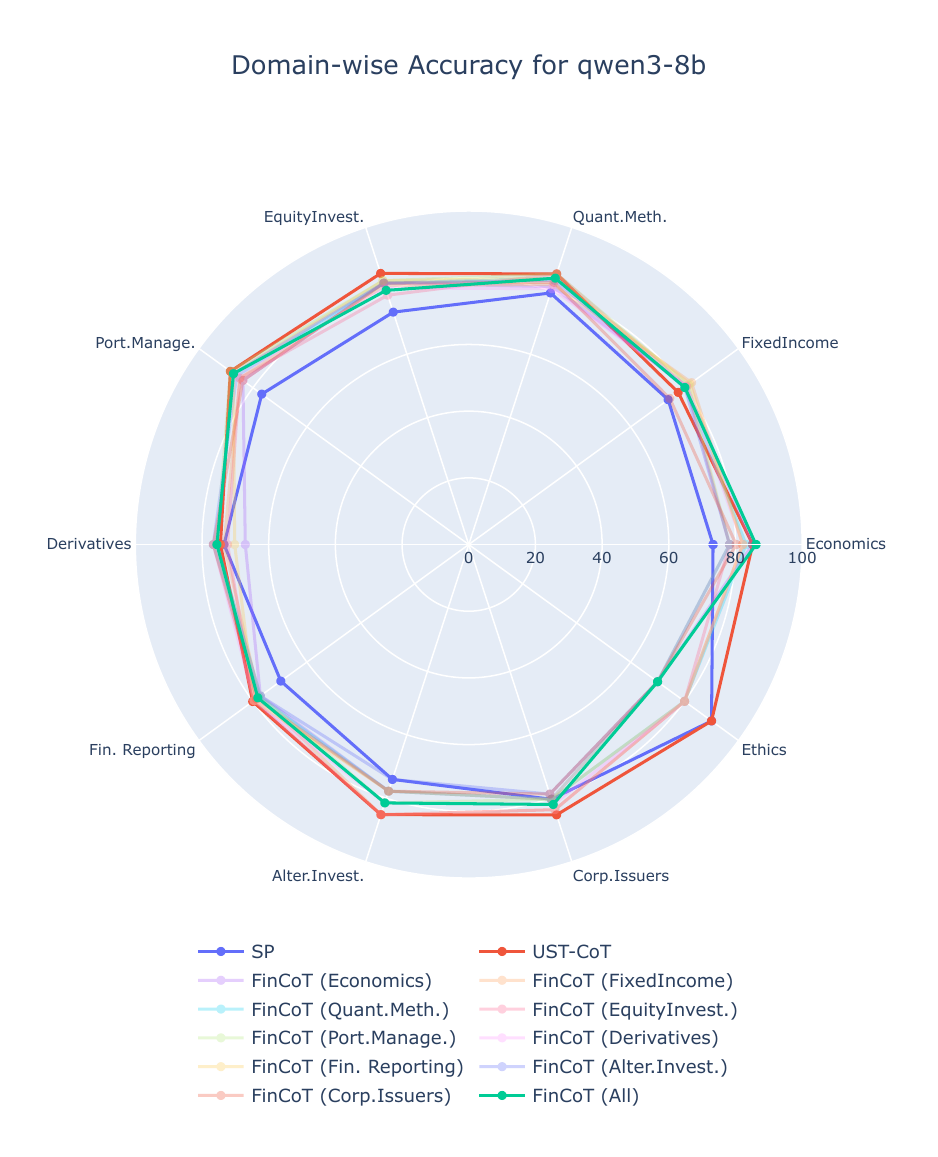}
    \caption{Qwen3-8B}
  \end{subfigure}

  \caption{Radar charts for each model variant.}
  \label{fig:radar_variants_1_4}
\end{figure*}

\clearpage
\begin{figure*}[!ht]
  \centering
  \begin{subfigure}{0.48\linewidth}
    \centering
    \includegraphics[width=\linewidth,keepaspectratio]{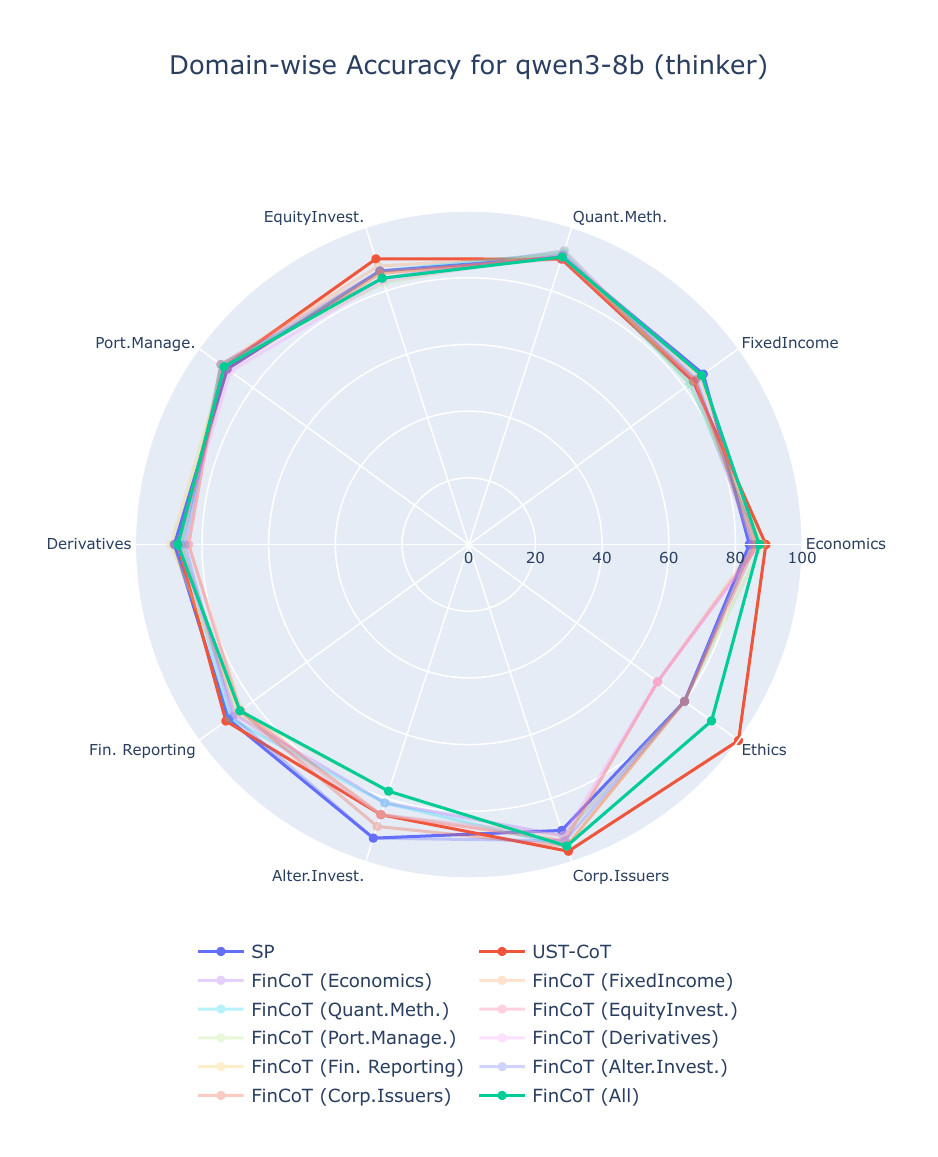}
    \caption{Qwen3-8B (Thinker)}
  \end{subfigure}
  \hfill
  \begin{subfigure}{0.48\linewidth}
    \centering
    \includegraphics[width=\linewidth,keepaspectratio]{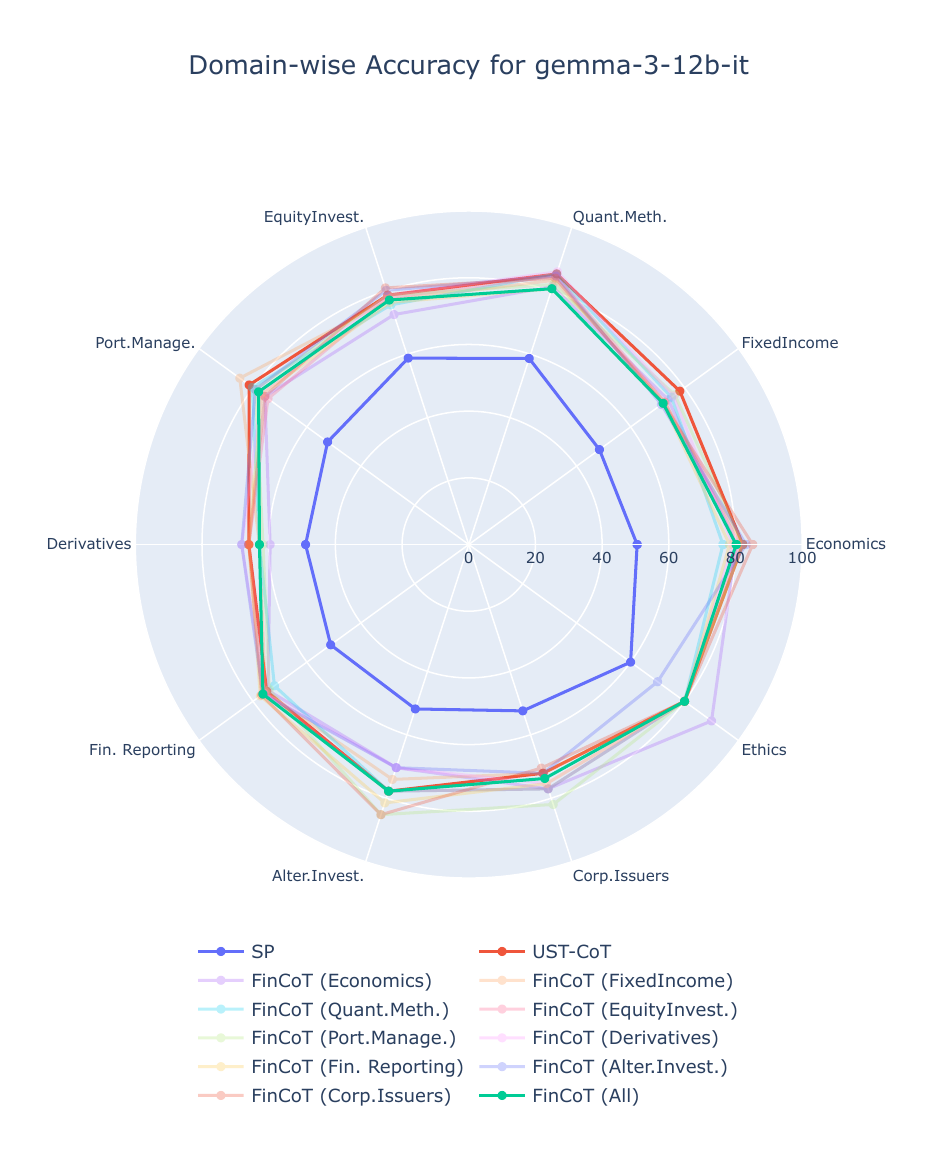}
    \caption{Gemma-3-12B-IT}
  \end{subfigure}

  \vspace{1em}

  \begin{subfigure}{0.48\linewidth}
    \centering
    \includegraphics[width=\linewidth,keepaspectratio]{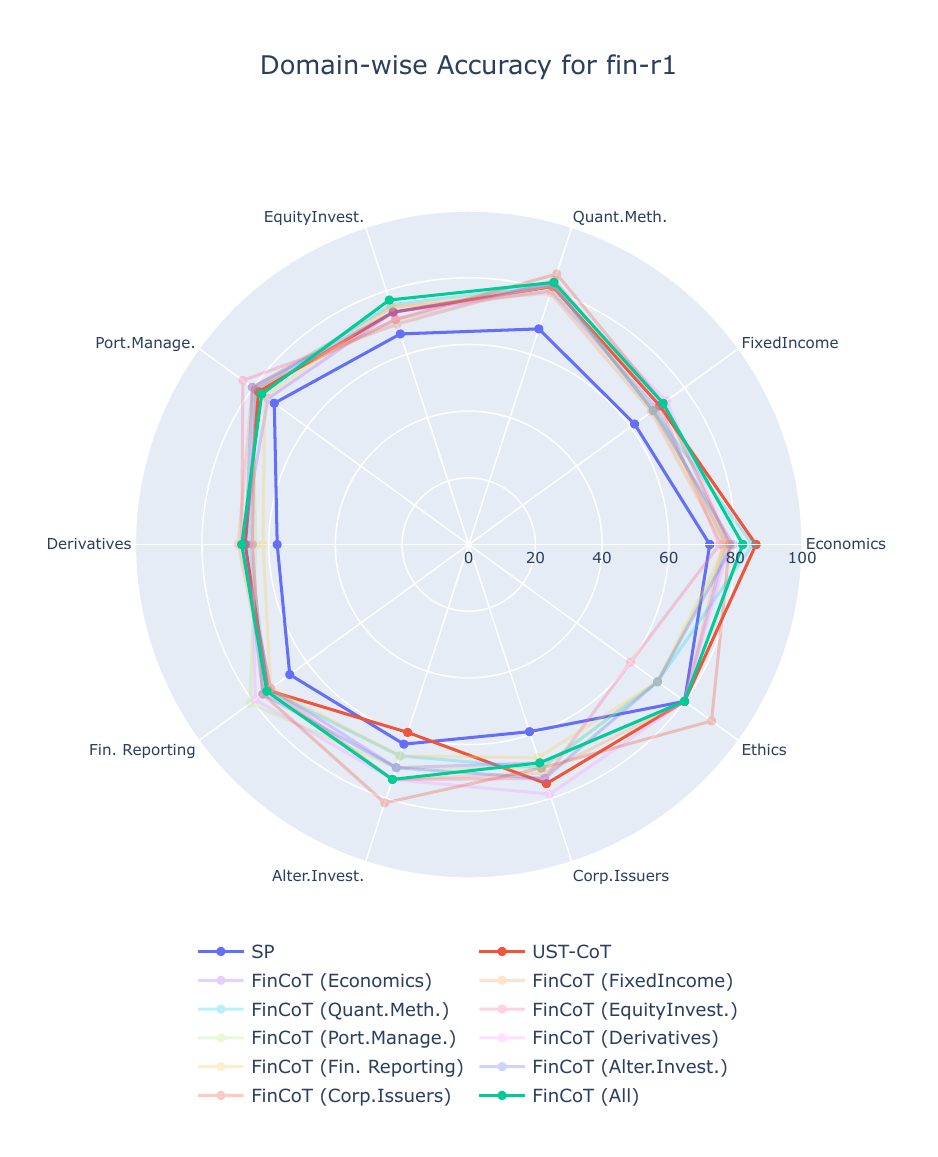}
    \caption{Fin-R1}
  \end{subfigure}
  \hfill
  \begin{subfigure}{0.48\linewidth}
    \centering
    \includegraphics[width=\linewidth,keepaspectratio]{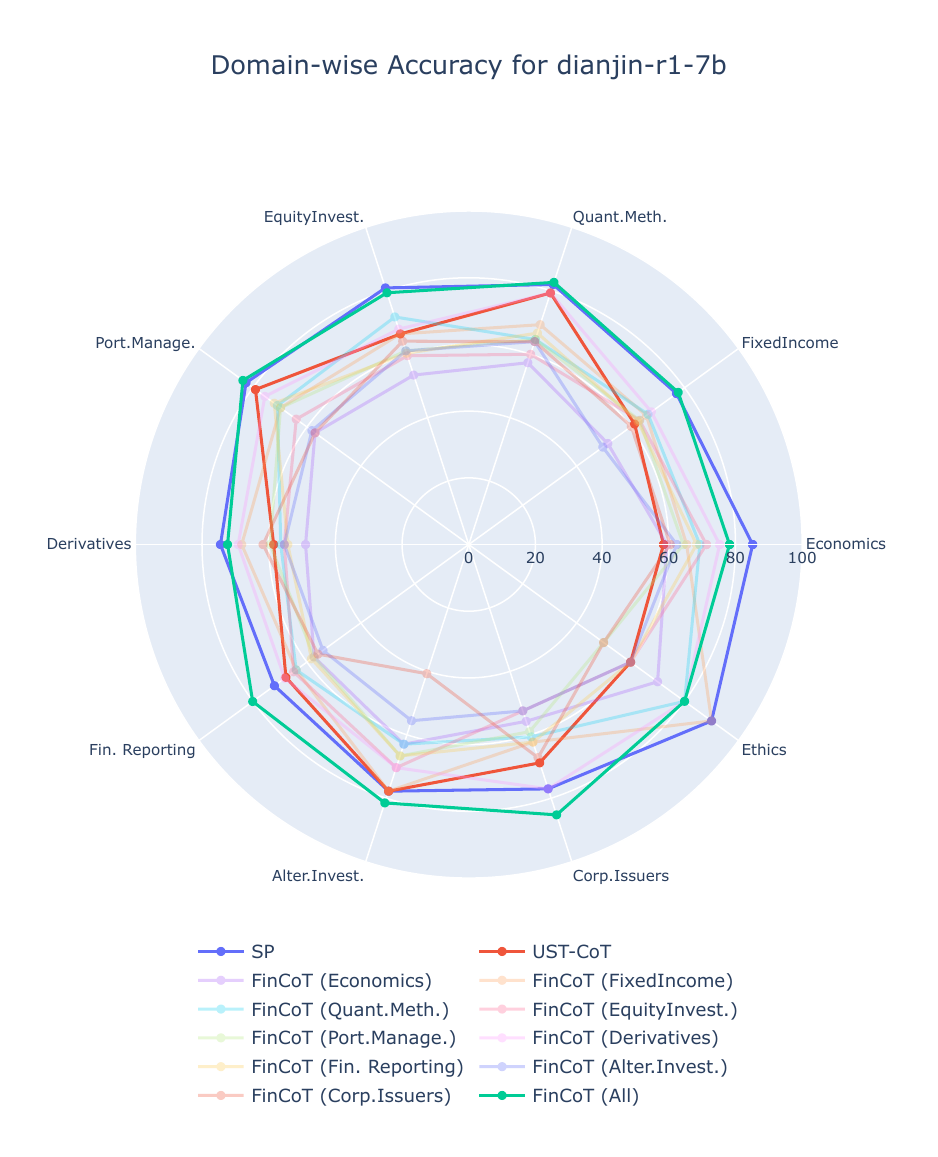}
    \caption{Dianjin-R1-7B}
  \end{subfigure}

  \caption{Radar charts for each model variant (charts 5–8).}
  \label{fig:radar_variants_5_8}
\end{figure*}

\clearpage
\begin{figure*}[!ht]
  \centering
  \includegraphics[width=0.48\linewidth,keepaspectratio]{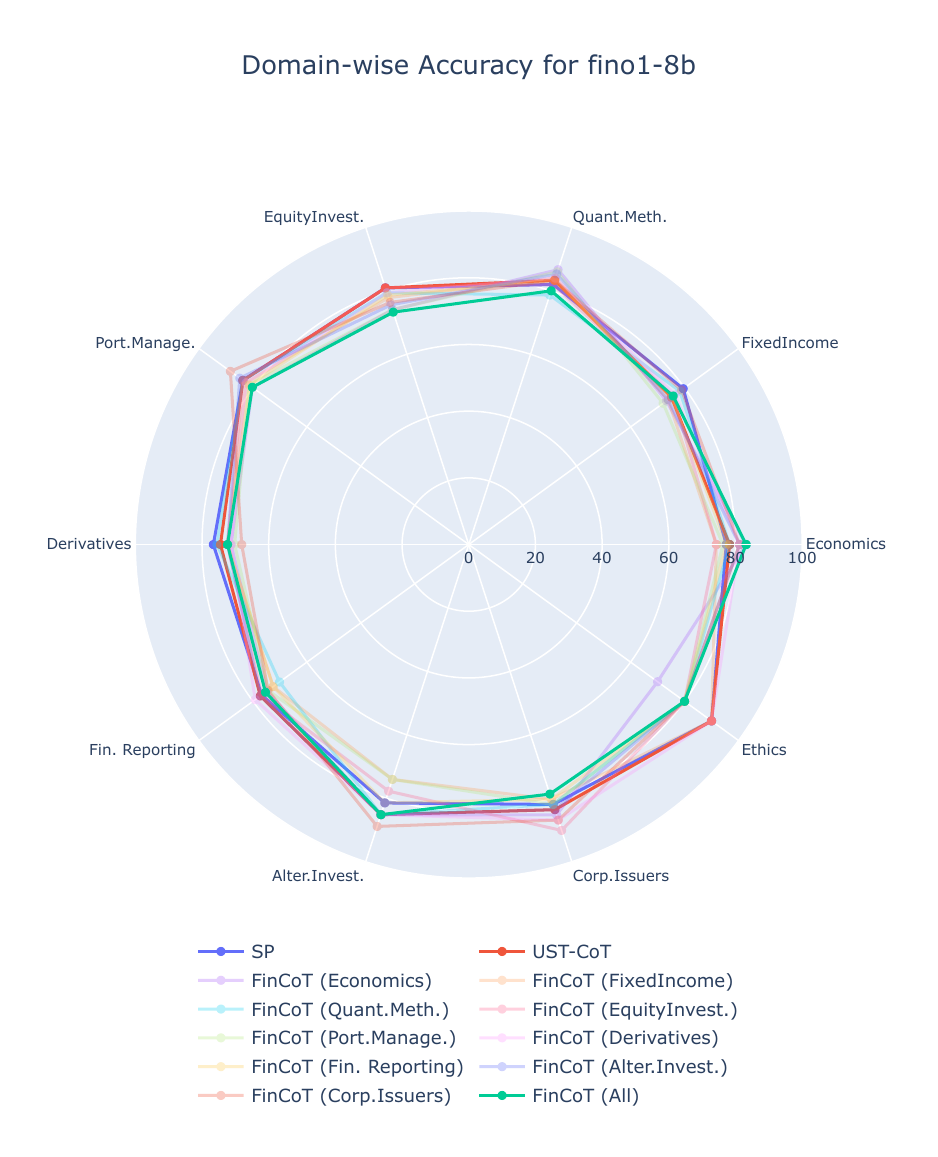}\hfill  
  \caption{Radar charts for each model variant (charts 9).}
  \label{fig:radar_variants_9}
\end{figure*}
\end{document}